\documentclass[final,3p,times]{elsarticle}

\usepackage{amssymb}
\usepackage{amsmath}
\usepackage{indentfirst}
\usepackage{graphicx}
\usepackage{color}
\usepackage{url}
\usepackage{bigstrut}
\usepackage{multirow,booktabs}
\usepackage[section]{algorithm}
\usepackage[noend]{algorithmic}
\usepackage{subfigure}

\usepackage{bm}
\usepackage{ragged2e}
\usepackage{setspace}
\usepackage{amsfonts}
\renewcommand{\raggedright}{\leftskip=0pt \rightskip=0pt plus 0cm}

\usepackage{array}
\newcommand{\PreserveBackslash}[1]{\let\temp=\\#1\let\\=\temp}
\newcolumntype{C}[1]{>{\PreserveBackslash\centering}p{#1}}
\newcolumntype{R}[1]{>{\PreserveBackslash\raggedleft}p{#1}}
\newcolumntype{L}[1]{>{\PreserveBackslash\raggedright}p{#1}}

\journal{Information Sciences}

\begin{document}

\begin{frontmatter}
\title{A Periodicity-based Parallel Time Series Prediction Algorithm in Cloud Computing Environments}
\author[Addr1,Addr5]{Jianguo~Chen}
\ead{jianguochen@hnu.edu.cn}
\author[Addr1,Addr2]{Kenli Li\corref{cor1}}
\ead{lkl@hnu.edu.cn}
\author[Addr1]{Huigui~Rong}
\ead{ronghg@hnu.edu.cn}
\author[Addr3]{Kashif~Bilal}
\ead{kashifbilal@ciit.net.pk}
\author[Addr1,Addr4]{Keqin Li\corref{cor1}}
\ead{lik@newpaltz.edu}
\author[Addr5]{Philip S. Yu}
\ead{psyu@uic.edu}

\cortext[cor1]{Corresponding authors.}

\address[Addr1]{College of Computer Science and Electronic Engineering, Hunan University, Changsha, Hunan 410082, China}
\address[Addr2]{National Supercomputing Center in Changsha, Changsha, Hunan 410082, China}
\address[Addr3]{Comsats Institute of Information Technology, Abbottabad 45550, Pakistan}
\address[Addr4]{Department of Computer Science, State University of New York, New Paltz, NY 12561, USA}
\address[Addr5]{Department of Computer Science, University of Illinois, Chicago.}

\begin{abstract}
In the era of big data, practical applications in various domains continually generate large-scale time-series data.
Among them, some data show significant or potential periodicity characteristics, such as meteorological and financial data.
It is critical to efficiently identify the potential periodic patterns from massive time-series data and provide accurate predictions.
In this paper, a Periodicity-based Parallel Time Series Prediction (PPTSP) algorithm for large-scale time-series data is proposed and implemented in the Apache Spark cloud computing environment.
To effectively handle the massive historical datasets, a Time Series Data Compression and Abstraction (TSDCA) algorithm is presented, which can reduce the data scale as well as accurately extracting the characteristics.
Based on this, we propose a Multi-layer Time Series Periodic Pattern Recognition (MTSPPR) algorithm using the Fourier Spectrum Analysis (FSA) method.
In addition, a Periodicity-based Time Series Prediction (PTSP) algorithm is proposed.
Data in the subsequent period are predicted based on all previous period models, in which a time attenuation factor is introduced to control the impact of different periods on the prediction results.
Moreover, to improve the performance of the proposed algorithms, we propose a parallel solution on the Apache Spark platform, using the Streaming real-time computing module.
To efficiently process the large-scale time-series datasets in distributed computing environments, Distributed Streams (DStreams) and Resilient Distributed Datasets (RDDs) are used to store and calculate these datasets.
Logical and data dependencies of RDDs in the P-TSDCA, P-MTSPPR, and P-PTSP processes are considered, and the corresponding parallel execution solutions are conducted.
Extensive experimental results show that our PPTSP algorithm has significant advantages compared with other algorithms in terms of prediction accuracy and performance.
\end{abstract}

\begin{keyword}
Big data \sep Distributed computing \sep  Parallel computing \sep Periodic pattern recognition  \sep Time series prediction.
\end{keyword}

\end{frontmatter}

\section{Introduction}
\subsection{Motivation}
With the rapid development of the Internet, sensor network, Internet of Things (IoT), mobile Internet and other media, a large number of datasets are continuously generated in various fields, such as large commercial, medical, engineering, and social sciences \cite{dataurl01, dataurl02, dataurl03,dataurl04}.
Time-series data are collections of datasets arranged in a time sequence, such as stock price, exchange rate, sales volume, production capacity, weather data, ocean engineering \cite{ec03, ec11,ec01}.
As important and complex data objects, massive time-series data truly record valuable information and knowledge about the applications, playing an important role in different application fields.
Abundant data mining and analysis technologies have been provided to seek the potentially available knowledge from these datasets.
Based on the previously observed time-series data, we can forecast the probable data in the coming periods.
It is interesting to seek high-performance approaches to handle the large-scale and streaming arrivals of time-series data.
In addition, the accuracy and robustness of time-series data processing methods are also hot topics in the academic and industrial fields.

The era of big data has brought both opportunities and challenges to the processing of large-scale time-series datasets.
On the one hand, in the era of big data, data generation and collection are becoming easier and less costly.
Massive datasets are continuously generated through various means, providing rich data sources for big data analysis and mining \cite{e28,e29}.
On the other hand, for time-series prediction, the emergence of the big data era also posed serious problems and challenges besides the obvious benefits.

\begin{itemize}
\item Periodic pattern recognition of time-series data is essential for time series prediction.
The periodic pattern of time-series data in the real world does not always keep a constant length (e.g. one day or one month) and may show dynamic length over time \cite{ec11}.
In addition, many time-series data have the characteristic of multi-layer periods.
Most of the existing periodic pattern recognition work calculate and analyze the single-layer period patterns.
It is necessary to adaptively identify time periodic patterns based on data-driven to discover the potential multi-layer periodic patterns.

\item To achieve accurate prediction, massive historical and real-time datasets are required for combination and analysis, which costs a lot of time to thoroughly excavate the historical data \cite{e30}.
Therefore, it is an important challenge that how to quickly process and analyze the massive historical data in the real-time prediction process.
The volume of massive datasets is usually much larger than the storage capacity of hard disks and memory on a single computer.
Therefore, we need to use distributed computing clusters to store and calculate these datasets.
This raises issues, such as data communication, synchronization waiting, and workload balancing, which need further consideration and resolution.

\item The performance of data analysis and prediction is also essential for large-scale time-series data.
There are increasingly strict time requirements for real-time time series prediction in various application fields, such as stock market, real-time pricing, and online applications \cite{ec04}.
Rapidly developed cloud computing and distributed computing provide high-performance computing capabilities for big data mining.
We need to propose efficient prediction algorithms for time-series data and execute these algorithms in high-performance computing environments.
In such a case, these algorithms can take full advantage of high-performance computing capabilities and increase their performance and scalability, while keeping lower data communication costs.
\end{itemize}

\subsection{Our Contributions}

In this paper, we focus on the periodic pattern recognition and prediction of large-scale time-series data with periodic characteristics, and a Periodicity-based Parallel Time Series Prediction (PPTSP) algorithm for time-series data in cloud computing environments.
A data compression and abstraction method is proposed for time-series data to effectively reduce the scale of massive historical datasets and extract the core information.
Fourier Spectrum Analysis (FSA) method is introduced to detect potential single-layer or multi-layer periodic patterns from the compressed time-series data.
The prediction algorithm is parallelized in the Apache Spark cloud platform, which effectively improves the performance of the algorithm and maintains high scalability and low data communication.
Extensive experimental results show that our PPTSP algorithm has significant advantages compared with other algorithms in terms of accuracy and performance.
Our contributions in this paper are summarized as follows.

\begin{itemize}
\item To effectively handle the massive historical datasets, a Time Series Data Compression and Abstraction (TSDCA) algorithm is presented, which can reduce the data scale as well as accurately extracting the characteristics.
\item We propose a Multi-layer Time Series Periodic Pattern Recognition (MTSPPR) algorithm using the FSA method.
    The first-layer periodic pattern is identified adaptively with the FSA method and morphological similarity measure.
     Then, potential multi-layer periodic patterns are discovered in the same way.
\item Based on the detected periodic patterns, a Periodicity-based Time Series Prediction (PTSP) algorithm is proposed to predict data values in subsequent time periods.
    An exponential attenuation factor is defined to control the impact of each previous periodic model on the prediction results.
\item To improve the performance of the proposed algorithms, we propose a parallel solution on the Apache Spark platform, using the Streaming real-time computing module.
    Distributed-Streams (DStreams) objects and Resilient Distributed Datasets (RDDs) are used to store and calculate these datasets in distributed computing environments.
\end{itemize}

The rest of the paper is organized as follows.
Section \ref{section2} reviews the related work.
Section \ref{section3} gives the multi-layer period prediction algorithm for time-series data, including the data compression and abstraction, FSA-based periodic pattern recognition, and periodicity-based time series prediction methods.
Parallel implementation of the periodic pattern recognition algorithm with Spark Streaming is developed in Section \ref{section4}.
Experimental results and evaluations are shown in Section \ref{section5} from the aspects of prediction accuracy and performance.
Finally, Section \ref{section6} concludes the paper with a discussion of future work and research directions.

\section{Related Work}
\label{section2}
In this section, we review the related work about time-series data mining from the perspectives of data compression and representation, periodic pattern recognition, time-series data prediction, and performance acceleration.

Focusing on large-scale time-series data compression and representation, various effective methods were proposed in \cite{ea02, ea04,ea03, ea05}.
In \cite{ea02}, the Chebyshev polynomials (CHEB) method was used to approximate and index the $d$-dimensional Spatio-Temporal trajectory, and the best extraction solution was obtained by minimizing the maximum deviation from the true value (termed the minimax polynomial).
However, CHEB is a global technique and requires expensive computational overhead for the large eigenvalue and eigenvector matrices.
As an approximation technique, the Piecewise Linear Approximation (PLA) algorithm was proposed in \cite{ea07} to approximate a time-series with line segments.
The representation consists of piecewise linear segments to represent the shape of the original time series.
In addition, an Indexable PLA (IPLA) algorithm was proposed in \cite{ea08} for efficient similarity search on time-series datasets.
Focusing on dimensionality reduction technique, Eamonn \emph{et al}. introduced a Piecewise Aggregate Approximation (PAA)  algorithm \cite{ea03} for high-dimensional time-series datasets.
In \cite{ea09}, a locally adaptive dimensionality reduction technique - Adaptive Piecewise Constant Approximation (APCA) algorithm was explored for indexing large-scale time-series databases.
There are other dimensionality reduction techniques, such as Singular Value Decomposition (SVD) \cite{ea06}, Discrete Fourier transform (DFT) \cite{ea04}, and Discrete Wavelet Transform (DWT) \cite{ea05}.
Detail experiments were performed in \cite{ea01} to compare the above time-series data representation methods and test their effectiveness on various time-series datasets.
However, most of the existing algorithms are implemented by dimensionality reduction or approximation, where DWT, PAA, and APCA are approximation methods with a discontinuous piecewise function.
The TSDCA algorithm proposed in this work falls in the category of approximation technique.
Different from the existing studies, TSDCA can extract the critical characteristics in each dimension to form a data abstraction without reducing the data dimensions.
It can guarantee the invariability of the data structure between the data abstraction and the raw dataset.
Similarity measurements, periodic pattern recognition, and prediction methods can be applied indiscriminately to the compressed dataset without any modification.

In the field of periodic pattern recognition of time series, various methods have been proposed \cite{eb04, eb03, eb05}, such as the complete periodic pattern, partial periodic pattern, period association rule, synchronous periodic pattern, and asynchronous periodic pattern.
In \cite{eb03}, Loh \emph{et al}. proposed an efficient method to mine temporal patterns in the popularity of web items, where the popularity of web items is treated as time series and a gap measurement method was proposed to quantify the difference between the popularity of two web items.
They further proposed a density-based clustering algorithm using the gap measure to find clusters of web items and illustrated the effectiveness of the proposed approach using real-world datasets on the Google Trends website.
In \cite{eb01, eb02}, Elfeky \emph{et al}. defined two types of periodicities: segment periodicity and symbol periodicity, and then proposed the corresponding algorithms (CONV and WARP) to discover the periodic patterns of unknown periods.
However, based on the convolution technique, the CONV algorithm works well on datasets with perfect periodicity, but faces limitations on noisy time series datasets.
The WARP algorithm uses the time warping technique to overcome the problem of noisy time series.
However, both CONV and WARP can only detect segment periodicity rather than symbol or sequence periodicity, and limited in detecting partial periodic patterns.
In \cite{eb07}, Sheng \emph{et al}. developed a ParPer-based algorithm to detect periodic patterns in time series datasets, where the dense periodic areas in the time series are detected using optimization steps.
However, this method requires pre-set expected period values.
In such a case, users should have the specific domain knowledge to generate patterns.
Rasheed \emph{et al}. proposed a Suffix-Tree-based Noise-Resilient (STNR) algorithm to generate patterns and detect periodicity from time series datasets \cite{eb06}.
The STNR algorithm can overcome the problem of finding periodicity without user specification and interaction.
However, the limitation of STNR is that it only works well in detecting fixed-length rigid periodic patterns, and it is poor effectiveness in tapping variable-length flexible patterns.
To overcome this limitation, Chanda \emph{et al}. introduced a Flexible Periodic Pattern Mining (FPPM) algorithm, which uses a suffix tree data structure and Discrete Fourier Transform (DFT) to detect flexible periodic patterns by ignoring unimportant or undesired events and only considering the important events \cite{eb03}.
However, in practical time series mining, the definition of the events of unimportant and important is difficult and infeasible.
In addition, most of the existing studies focused on static time-series database and the periodic pattern recognition in a single layer.
Considering that there are multiple nested periods on some real-world time-series datasets, i.e., the temperature shows periods both daily and seasonally, we focus on the potential multi-layer periodicity pattern recognition in this work.
In addition, to effectively detect flexible periodic patterns without user preparation knowledge, we propose a novel morphological similarity measurement and introduce the Fourier Spectrum Analysis (FSA) method for multi-layer periodicity pattern detection.
The morphological similarity is measured by a five-tuple ($AS_{a,b}$, $TLS_{a,b}$, $MaxS_{a,b}$, $MinS_{a,b}$, $VIS_{a,b}$), which refer to the angular similarity, time-length similarity, maximum similarity, minimum similarity, and the value-interval similarity, respectively.
The combination of the FSA and morphological similarity measurement can efficiently calculate the compressed time series from incremental online time series streams.
Moreover, the morphological similarity measurement can be further applied to various periodic pattern recognition algorithms.

Over the past several decades, various time series prediction algorithms were proposed in existing studies, such as seasonal autoregressive differential sliding average, Holt-Winters index \cite{ec05, ec07, ec11,ec10}.
In \cite{ec05}, a novel high-order weighted fuzzy time series model was proposed and applied in nonlinear time series prediction.
George \emph{et al.} used an online sequential learning algorithm for time-series prediction, where a feed-forward neural network was introduced as an online sequential learning model \cite{ec06}.
Focus on local modeling, Marcin \emph{et al.} proposed a period-aware local modeling and data selection for time series prediction \cite{ec07}, where the period of time series is determined by using autocorrelation function and moving average filter.
Shi \emph{et al}. proposed an offline seasonal adjustment factor plus GARCH model to model the seasonal heteroscedasticity in traffic flow series \cite{ec11}.
However, this model faces limitations in real-world transportation time-series processing.
In \cite{ec08}, Huang \emph{et al}. introduced an online seasonal adjustment factors plus adaptive Kalman filter (OSAF+AKF) algorithm for the prediction of the seasonal heteroscedasticity in traffic flow datasets.
Considering the seasonal patterns in traffic time-series datasets, four types of online seasonal adjustment factors are introduced in the OSAF+AKF algorithm.
In addition, Tan \emph{et al}. defined a time-decaying online convex optimization problem and explored a Time-Decaying Adaptive Prediction (TDAP) algorithm for time series prediction \cite{ec12}.
In the biomedical field, time-series forward prediction algorithms were used for real-time brain oscillation detection and phase-locked stimulation in \cite{ee01}.

With the emergence of big data, the processing performance and real-time response requirements of large-scale time series applications have received increasing attention.
Various acceleration and parallel methods were proposed for massive time-series data processing \cite{e30, ec12, e19}.
In \cite{e30}, a GP-GPU parallelization solution was introduced for fast knowledge discovery from time-series datasets, where a General Programming (GP) framework was presented using the CUDA platform.
Efforts on distributed and parallel time-series data mining based on high-performance computing and cloud computing have achieved abundant favorable achievements \cite{e35, e36}.
Apache Spark \cite{url02} is another good cloud platform that is suitable for data mining.
It allows us to store a data cache in memory and to perform computations and iteration of the same data directly from memory.
The Spark platform saves huge amounts of disk I/O operation time.
Spark Streaming is a real-time computing framework based on the Spark cloud environment.
It provides many rich APIs and high-speed engines based on memory computing.
Users can combine the Spark Streaming with applications such as flowing computing, batch processing, and interactive queries.
In \cite{e29}, the Spark Streaming module was used to implement the nearest neighbor classification algorithm  for high-speed big data streams.
In \cite{ec04}, an effective prediction algorithm was proposed based on the Apache Spark for missing data over multi-variable time series.

\section{Periodicity-based Time Series Prediction Algorithm}
\label{section3}
In this section, we propose a Multi-layer Time Series Periodic Pattern Recognition (MTSPPR) algorithm for time-series data with periodic characteristics.
In Section \ref{section3.1}, to accelerate the periodic pattern recognition process of large-scale time-series datasets, a data compression and abstraction method is proposed, which can effectively extract the characteristics of data while reducing the scale of massive datasets.
In Section \ref{section3.2}, the Fourier Spectrum Analysis (FSA) method is used to identify periodic patterns from the compressed time-series dataset.
On these bases, Section \ref{section3.3} describes the multi-layer periodic pattern recognition algorithm.
Each potential senior-layer period model is constructed successively based on the periods in the previous low-layer models.

\subsection{Time-series Data Compression and Abstraction}
\label{section3.1}
In many actual applications, time-series datasets grow at high speed over time.
Although various storage technologies continue to be improved and storage costs are declining, it is still difficult to cope with the rapid development of large-scale datasets.
To process large-scale and continuous time-series datasets using limited storage and computing resources, we propose a Time-Series Data Compression and Abstraction (TSDCA) algorithm to effectively reduce the data volume and extract key knowledge.

\begin{figure}[!ht]
 \setlength{\abovecaptionskip}{0pt}
 \setlength{\belowcaptionskip}{0pt}
\centering
\includegraphics[width=4.5in]{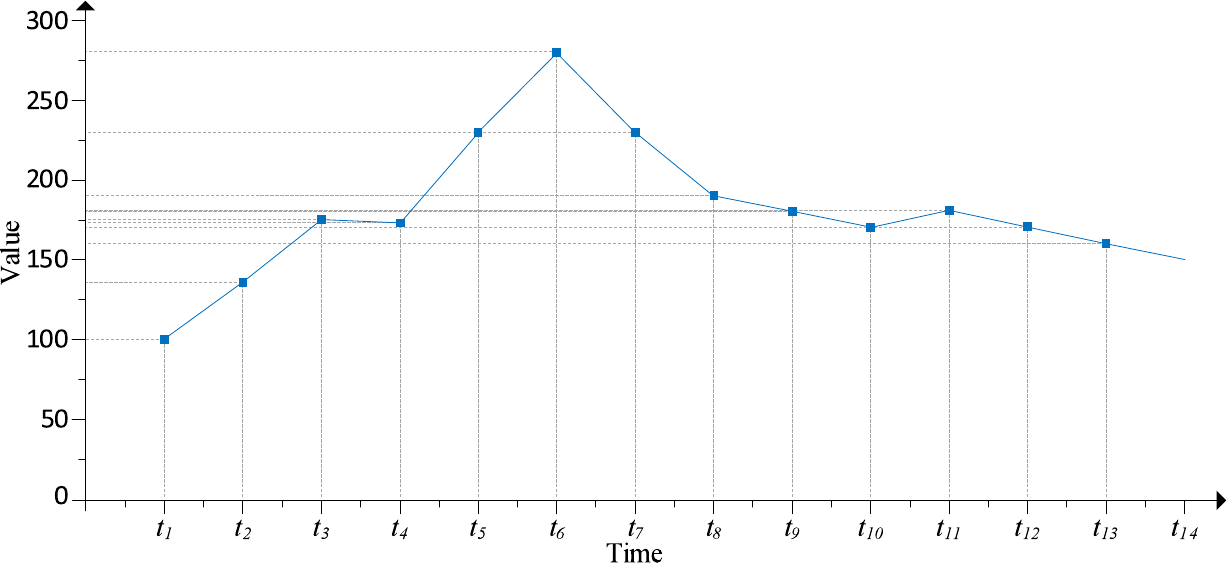}
\caption{Data compression and abstraction of large-scale time-series datasets.}
\label{fig01}
\end{figure}

Given a big data processing application, let $X_{T}=\{(x_{1},~t_{1}),~(x_{2},~t_{2}), ~..., ~(x_{n},~t_{n})\}$ be the raw time-series dataset with temporal and periodic attributes, where $x_{i}$ is the data point at the time stamp $t_{i}$.
In this way, the raw dataset can be compressed by a series of data points and the slopes between these points.
An example of the raw two-dimensional time-series dataset is compressed in Figure \ref{fig01}.

(1) Inclination measurement and inflection points mark.

\begin{figure}[!ht]
 \setlength{\abovecaptionskip}{0pt}
 \setlength{\belowcaptionskip}{0pt}
\centering
\includegraphics[width=3.4in]{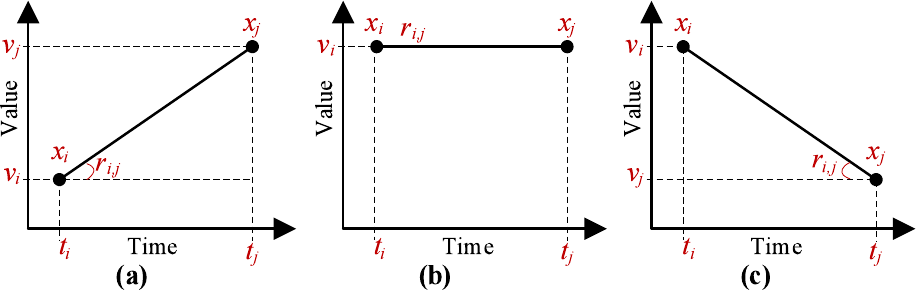}
\caption{Inclination relationships between two data points.}
\label{fig02}
\end{figure}

To extract the characteristics of a large-scale time-series dataset, we calculate the inclination of every two data points and identify the inflection points of the dataset.
The inclination between two data points is the ratio of the value difference and time difference between the two data points, as defined in Equation (\ref{eq01}):

\begin{equation}
\label{eq01}
\begin{aligned}
r_{i,j} = \frac{x_{j} - x_{i}}{t_{j} - t_{i}}   ~~~~~~(i<j),
\end{aligned}
\end{equation}
where $r_{i,j}$ is the inclination between data points $x_{i}$ and $x_{j}$.
There are three conditions for $r_{i,j}$:
(a) $r_{i,j} > 0$ refers to an upward trend;
(b) $r_{i,j} = 0$ shows a steady trend;
and (c) $r_{i,j} < 0$ refers to a downward trend.
Examples of the inclination relationships between two data points are shown in Figure \ref{fig02}.

\begin{figure}[!ht]
 \setlength{\abovecaptionskip}{0pt}
 \setlength{\belowcaptionskip}{0pt}
\centering
\includegraphics[width=4.5in]{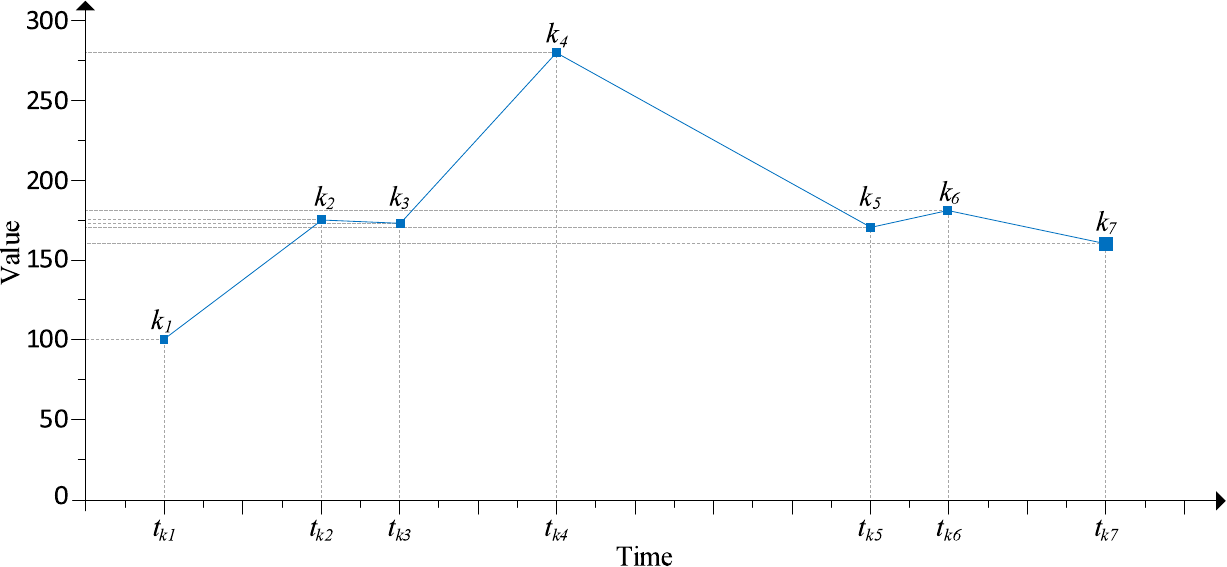}
\caption{Inflection points of the raw time-series dataset.}
\label{fig03}
\end{figure}

The inflection points set for $X_{T}$ is initialized as an empty set ($X_{K} = \{null\}$).
Set the first inflection point $k_{1} = x_{1}$, $t_{k_{1}} = t_{1}$.
We continuously calculate the inclination $r_{1,2}$ between $k_{1}$ and data point $x_{2}$, and $r_{2,3}$ between data points $x_{2}$ and $x_{3}$.
If $r_{1,2} \times r_{2,3} > 0$, the data points $k_{1}$, $x_{2}$, and $x_{3}$ have a congruous trend.
Namely, $x_{2}$ is not an inflection point here.
In this case, we continue to calculate the slopes of the subsequent data points and multiply them with the inclination rate $r_{12}$.
Otherwise, if $r_{1,2} \times r_{2,3} \leq 0$, it indicates an incongruous trend of data points $k_{1}$ to $x_{2}$ and $x_{2}$ to $x_{3}$.
That is, here $x_{2}$ represents an inflection point.
We append $x_{2}$ to the inflection point set $X_{K}$ and set $k_{2} = x_{2}$, $t_{k_{2}} = t_{2}$.
Similarly, the slopes of the remaining data points are computed sequentially by repeating the above steps.
In this way, the large-scale raw time-series dataset $X_{T}$ is compressed and re-expressed as the form of inflection points $X_{K}$, as described as:
\begin{center}
$X_{K}= \left\{(k_{1}, ~t_{k_{1}}), ~(k_{2}, ~t_{k_{2}}), ~..., ~(k_{m}, ~t_{k_{m}})\right\}$,
\end{center}
where $m$ ($1 \leq m \ll n$) is the number of inflection points.
Note that the scale of the compressed dataset is much smaller than the raw dataset.
Depending on these inflection points, the raw time-series dataset is divided into multiple linear segments.
These segments can be connected to form an abstract representation of the raw dataset, as shown in Figure \ref{fig03}.

(2) Pseudo-inflection points deletion.

Considering that the set of inflection points still contains a lot of inflection points that have similar values to the neighbors in the abstract representation dataset.
We need to further identify and remove these pseudo-inflection points to effectively describe the significant outlines of the raw dataset.

\textbf{Definition 1: (Pseudo-inflection point)}.
\textit{Pseudo-inflection points refer to the inflection points of which the values have a negligible difference from their neighbors.
These data points have little impact on the distribution trends and patterns of the neighborhood of the abstract representation.
After removing these pseudo-inflection points, the overall outlines of the abstract representation dataset will be well maintained.}

We respectively calculate the slopes of every three adjacent inflection points to determine whether the middle one is a pseudo-inflection point.
Let $R_{i,i+1}$ be the inclination between inflection points $k_{i}$ and $k_{i+1}$, as calculated in Equation (\ref{eq02}):

\begin{equation}
\label{eq02}
\begin{aligned}
R_{i,i+1} = \frac{k_{i+1} - k_{i}}{t_{k_{i+1}} - t_{k_{i}}}.
\end{aligned}
\end{equation}

According to Equation (\ref{eq02}), we continue to calculate the slopes $R_{i,i+1}$, $R_{i+1,i+2}$, and $R_{i,i+2}$ among $k_{i}$, $k_{i+1}$, and $k_{i+2}$, respectively.
The inclination relationship of inflection points $k_{i}$, $k_{i+1}$, and $k_{i+2}$ is calculated in terms of value differences and time difference, as defined in Equation (\ref{eq03}) and (\ref{eq04}):
\begin{equation}
\label{eq03}
\begin{aligned}
S_{R}(i,i+2) = \frac{|R_{i,i+2}|}{max(|R_{i,i+1}|, ~|R_{i+1,i+2}|)} \geq \delta,
\end{aligned}
\end{equation}
\begin{equation}
\label{eq04}
\begin{aligned}
S_{T}(i,i+2) = \frac{max(t_{k_{i+1}}-t_{k_{i}},~t_{k_{i+2}} - t_{k_{i+1}})}{t_{k_{i+2}} - t_{k_{i}}} \geq \varepsilon,
\end{aligned}
\end{equation}
where $\delta$ is the inclination threshold ($0< \delta \leq 1$ ) and $\varepsilon$ is the threshold of the length of time ($0< \varepsilon \leq 1$ ).
If the inflection points $k_{i}$, $k_{i+1}$, and $k_{i+2}$ satisfy the inclination relationships in Equations (\ref{eq03}) and (\ref{eq04}), then $k_{i+1}$ is identified as a pseudo-inflection point and removed from $X_{K}$.

\begin{figure}[!ht]
 \setlength{\abovecaptionskip}{0pt}
 \setlength{\belowcaptionskip}{0pt}
\centering
\includegraphics[width=4.5in]{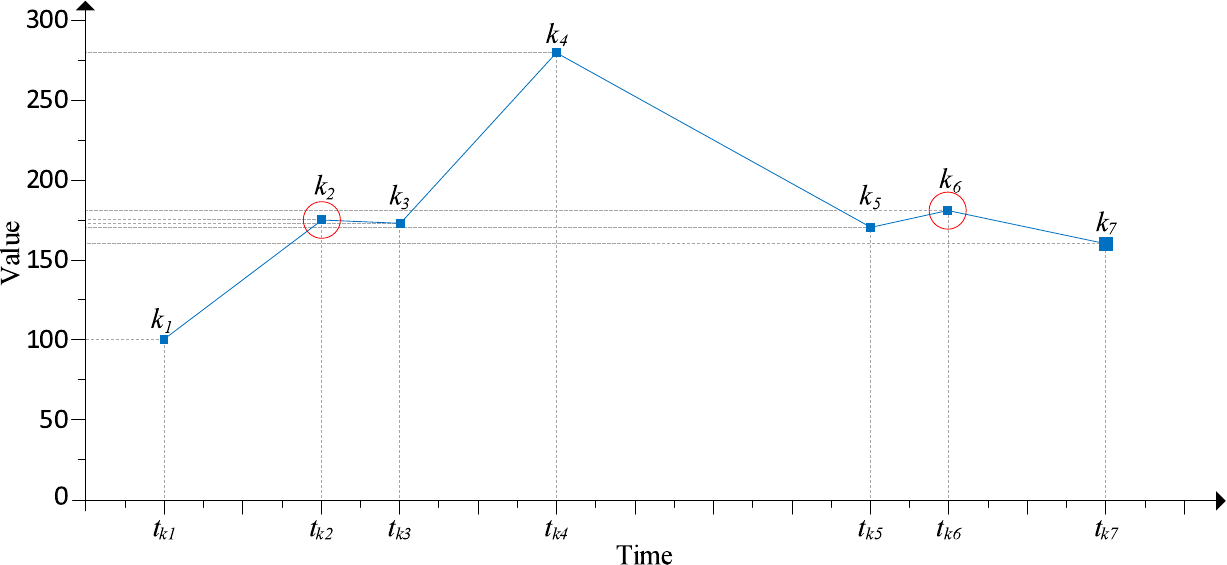}
\caption{Pseudo-inflection points of data abstract representation.}
\label{fig04}
\end{figure}

For example, in Figure \ref{fig04}, $k_{2}$ and $k_{6}$ are identified as pseudo-inflection points.
After removing $k_{2}$ and $k_{6}$, the set of inflection points $X_{K}$ is updated to $X_{K}=\{k_{1},~k_{3},~k_{4},~k_{5},~k_{7}\}$.

(3) Data compression and abstraction of the raw time-series dataset.

\begin{figure}[!ht]
 \setlength{\abovecaptionskip}{0pt}
 \setlength{\belowcaptionskip}{0pt}
\centering
\includegraphics[width=4.5in]{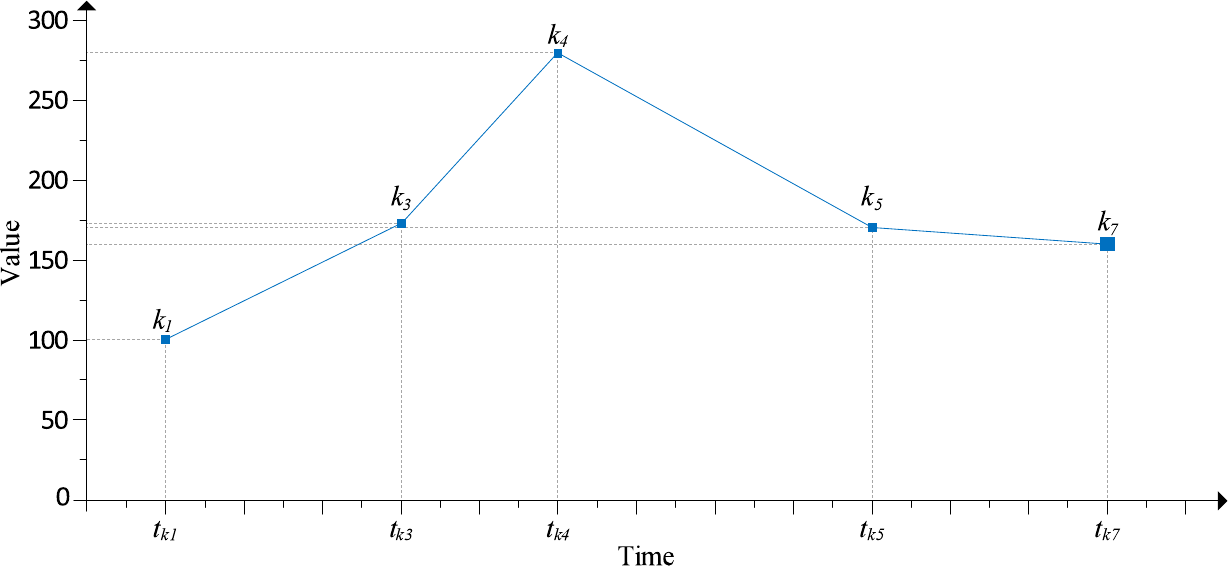}
\caption{Data compression and abstraction of the raw time-series dataset.}
\label{fig05}
\end{figure}

For the raw time-series dataset $X_{T}$, inflection points, excepting the pseudo-inflection points, are collected to form a compressed and abstracted representation $X_{K}$.
In this way, the large-scale dataset can be effectively compressed to reduce the data size while effectively extracting the core information.
For example, the data abstraction of the raw dataset in Figure \ref{fig01} is shown in Figure \ref{fig05}.
Detailed steps for time-series data compression and abstraction algorithm are given in Algorithm \ref{alg1}.

\begin{algorithm}[!ht]
\caption{Time-series data compression and abstraction (TSDCA) algorithm}
\label{alg1}
\begin{algorithmic}[1]
\REQUIRE ~\\
    $X_{T}$: the raw time-series dataset;\\
    $\delta$: the inclination threshold;\\
    $\varepsilon$: the threshold value of the time-window length;\\
\ENSURE ~\\
    $X_{K}$: the data abstraction of $X_{T}$.
\STATE  initialize the inflection point set as empty $X_{K} = \{null\}$;
\STATE  set the first inflection point $k_{1}$ $\leftarrow$ $x_{1}$, and $t_{k_{1}} \leftarrow t_{1}$;
\FOR{each data point $x_{i}$ in $X_{T}$}
\STATE  calculate inclination rates $r_{i,i+1} \leftarrow \frac{x_{i+1} - x_{i}}{t_{i+1} - t_{i}}$, $r_{i+1,i+2} \leftarrow \frac{x_{i+2} - x_{i+1}}{t_{i+2} - t_{i+1}}$;
\IF {($r_{i,i+1} \times r_{i+1,i+2} >0$)}
\STATE \textbf{continue};
\ELSE
\STATE mark inflection point $X_{K} \leftarrow (k_{j} \leftarrow x_{i+1}, ~t_{k_{j}} \leftarrow t_{i+1})$;
\ENDIF
\ENDFOR
\FOR{each inflection point $k_{j}$ in $X_{K}$}
\STATE  calculate inclination rates $R_{j,j+1} \leftarrow \frac{k_{j+1} - k_{i}}{t_{k_{j+1}} - t_{k_{j}}}$, $R_{j+1,j+2} \leftarrow \frac{k_{j+2} - k_{i+1}}{t_{k_{j+2}} - t_{k_{j+1}}}$;
\STATE  calculate inclination relationship of value $S_{R}(i,i+2) \leftarrow \frac{R_{i,i+2}}{max(R_{i,i+1}, ~R_{i+1,i+2})}$;
\STATE  calculate inclination relationship of time $S_{T}(i,i+2) \leftarrow \frac{t_{k_{i+2}} - t_{k_{i}}}{max(t_{k_{i+1}}-t_{k_{i}},~t_{k_{i+2}} - t_{k_{i+1}})}$;
\IF {($S_{R}(i,i+2) \geq \delta$ and $S_{T}(i,i+2)  \geq \varepsilon$)}
\STATE remove pseudo-inflection point $k_{j+1}$ from $X_{K}$;
\ENDIF
\ENDFOR
\RETURN $X_{K}$.
\end{algorithmic}
\end{algorithm}

The TSDCA algorithm consists of the processes of inflection points marking and pseudo-inflection points deletion.
Assuming that the number of data points of the raw dataset $X_{T}$ is equal to $n$ and the data abstraction $X_{K}$ has $m$ inflection points, the time complexity of Algorithm \ref{alg1} is $O(n+m)$.
The data compression ratio between $X_{T}$ and $X_{K}$ is $\frac{n}{m}$.
Benefitting from the data compression and abstraction, the storage requirement and the data processing workload of big data are reduced effectively.

\subsection{Multi-layer Time Series Periodic Pattern Recognition}
\label{section3.2}
In this section, we propose a Multi-layer Time Series Periodic Pattern Recognition (MTSPPR) algorithm.
A morphological similarity measurement is proposed for the continuous arrival time-series datasets.
The Fourier Spectrum Analysis (FSA) method is used to identify the potential periodic models of time-series datasets.

\subsubsection{FSA-based Periodic Pattern Recognition}
Given a data abstraction $X_{K} = \{k_{1}, ~k_{2}, ~..., ~k_{m}\}$ of the raw time-series dataset $X_{T}$, $X_{K}$ can be described as a non-stationary data model, including trend item $H_{t}$, periodic item $P_{t}$, and random item $Y_{t}$, as described as:

\begin{equation}
\label{eq05}
\begin{aligned}
X_{K} = H_{t} + P_{t} + Y_{t}, ~(t = 1,~2,~...,~m).
\end{aligned}
\end{equation}
If the periodic item $P_{t}$ satisfies the expansion conditions of the Fourier series, we can extend $P_{t}$ in $(-\infty, ~+\infty)$ using a periodic length $m$.
Then, we get the Fourier expansion of $P_{t}$ in the interval $[-\frac{m}{2},~+\frac{m}{2}]$.
Namely, $P_{t}$ is represented as the sum of a series of spectrums doubling in frequency in the interval $[-\frac{m}{2},~+\frac{m}{2}]$, as described in Equation (\ref{eq06}):

\begin{equation}
\label{eq06}
\begin{aligned}
P_{t}' = a_{0} + \sum_{i=1}^{k}{\left[a_{i}\cos{(i\omega_{0}t)} + b_{i}\sin(i\omega_{0}t)\right]},
\end{aligned}
\end{equation}
where $P_{t}'$ is an estimate of $P_{t}$, which makeups by $k$ spectrums and the average component $a_{0}$.
$k=\left\lfloor \frac{m}{2} \right\rfloor$ is the highest item in these spectrums.
$a_{i} = \lambda_{i}\cos\theta_{i}$ and $b_{i} = \lambda_{i}\sin\theta_{i}$ are the amplitudes of the cosine and sine components of each spectrum.
$\theta_{i}$ is the initial phase angle of each spectrum.
$\omega_{0}=\frac{2\pi}{m}$ is the basic angular frequency.
The number of these $k$ spectrums does not exceed $\frac{m}{2}$, namely, $P_{t}$ is approximated by a limited number of spectrums.

According to the least square method, to obtain the values of the coefficients in Equation (\ref{eq06}), the quadratic sum $Q$ of the fitting error $e_{t}$ in Equation (\ref{eq07}) should be minimized.

\begin{equation}
\label{eq07}
Q = \sum_{t=1}^{m}|e_{t}|^{2} = \sum_{t=1}^{m}|P_{t} - P_{t}'|^{2}.
\end{equation}

We calculate partial derivatives of Q with $a_{i}$ and $b_{i}$ and make them equal to 0, then
\begin{spacing}{2.0}
\begin{equation}
\label{eq08}
\left\{
\begin{array}{ll}
\sum\limits_{t=1}^{m}{\left[P_{t} - \sum\limits_{i=1}^{k}{\left(a_{i}\cos{(i\omega_{0}t)} + b_{i}\sin(i\omega_{0}t)\right)} \right]} | (\cos{(i\omega_{0}t)} = 0), \\
\sum\limits_{t=1}^{m}{\left[P_{t} - \sum\limits_{i=1}^{k}{\left(a_{i}\cos{(i\omega_{0}t)} + b_{i}\sin(i\omega_{0}t)\right)}\right]} |  (\sin{(i\omega_{0}t)} = 0).\\
\end{array}
\right.
\end{equation}
\end{spacing}

According to the orthogonality of the trigonometric function, we calculate Equation (\ref{eq09}) to get the estimated expression of each Fourier spectrum coefficient:
\begin{spacing}{1.5}
\begin{equation}
\label{eq09}
\left\{
\begin{array}{ll}
a_{0} = \frac{1}{m} \sum\limits_{t=1}^{m}{P_{t}},\\
a_{i} = \frac{2}{m} \sum\limits_{t=1}^{m}{P_{t}\cos{\frac{2\pi i}{m}t}}, \\
b_{i} = \frac{2}{m} \sum\limits_{t=1}^{m}{P_{t}\sin{\frac{2\pi i}{m}t}}. \\
\end{array}
\right.
\end{equation}
\end{spacing}
The overall variance of the periodic item $P_{t}$ of the time-series data abstraction $X_{K}$ is defined in Equation (\ref{eq10}):
\begin{equation}
\label{eq10}
\begin{aligned}
S_{P}^{2} &= \frac{1}{m}\sum_{t=1}^{m}{(P_{t}^{2}-\overline{P_{t}})^{2}}\\
          &=\frac{1}{m}\sum_{t=1}^{m}{(a_{0} + \sum_{i=1}^{k}{\left[a_{i}\cos{(i\omega_{0}t)} + b_{i}\sin(i\omega_{0}t)\right]}-\overline{P_{t}})^{2}}\\
          &=\frac{1}{m}\sum_{t=1}^{m}\left\{\sum_{i=1}^{k}{\left[a_{i}\cos{(i\omega_{0}t)} + b_{i}\sin(i\omega_{0}t)\right]}\right\}^{2}\\
          &=\frac{1}{2}\sum_{i=1}^{k}{(a_{i}^{2} + b_{b}^{2})}.
\end{aligned}
\end{equation}

Let $C_{i}^{2} = (a_{i}^{2} + b_{b}^{2})$ be the spectrum compositions of $X_{K}$, we use a statistic $\vartheta$ to evaluate the significance of the variance of each spectrum, as defined in Equation (\ref{eq11}):

\begin{equation}
\label{eq11}
\vartheta_{i} = \frac{C_{i}^{2}}{S_{P}^{2}-\frac{1}{m}C_{i}^{}2}.
\end{equation}
According to Equation (\ref{eq11}), we get the spectrum with the maximum significance $i = arg ~max(\vartheta_{i})$ and set $L=\frac{m}{i}$ as the period length of $X_{K}$.

\subsubsection{Morphological Similarity Measurement}

From Equation (\ref{eq09}), we can see that the Fourier coefficients of each spectrum depend on the time sequence length $m$.
In practical applications, time-series datasets are generated in an endless flow.
Namely, new arriving time-series datasets continuously append to the original sequence.
In such a case, $m$ is constantly updated with the arriving of new datasets, resulting in the Fourier coefficients need to be recalculated repeatedly.
To effectively improve the performance of the periodic pattern recognition, we propose a morphological similarity measurement and optimize Equation (\ref{eq09}) for the new arriving time-series datasets.
In the morphological similarity measurement, the sequence of the data abstraction $X_{K}$ is partitioned into multiple subsequences.
Then, we calculate the morphological similarities of these subsequences and provide new estimated expression of each Fourier spectrum coefficient.

Given a periodic item $P_{t}$, assuming that there are two subsequences $\ell_{a}$ and $\ell_{b}$ in $P_{t}$, and each subsequence consists of $h$ inflection points.
Namely, there are $(h-1)$ subsequences in $\ell_{a}$ and $\ell_{b}$, respectively.
The morphological similarity between subsequences $\ell_{a}$ and $\ell_{b}$ is measured from five aspects: angular similarity, time length similarity, maximum similarity, minimum similarity, and value-interval similarity.

(1) Angular similarity.

\textbf{Definition 2: (Angular similarity)}.
\textit{The angular similarity $AS_{a,b}$ between two subsequences $\ell_{a}$ and $\ell_{b}$ refers to the average of the angular similarities between the individual linear segments in the two subsequences.
The angular similarity $AS_{ai,bi}$ of each linear-segment part $s_{i}$ in $\ell_{a}$ and $\ell_{b}$ is equal to the ratio of the difference of inclination rates between these two segments to the larger inclination rate.
$AS_{a,b}$ is calculated by Equation (\ref{eq12}):}

\begin{equation}
\label{eq12}
\begin{aligned}
AS_{a,b} =\frac{1}{(h-1)}\sum_{i=1}^{h-1} {AS_{ai,bi}}
         =\frac{1}{(h-1)}\sum_{i=1}^{h-1} {(1 - \frac{|R_{ai} - R_{bi}|}{max(R_{ai}, R_{bi})} ) \times 100\%},
\end{aligned}
\end{equation}
where $R_{ai}$ is the inclination rate of linear segment $s_{i}$ in subsequence $\ell_{a}$ and $R_{bi}$ is that of $s_{i}$ in $\ell_{b}$.
The time-length similarity $TLS_{a,b}$, maximum similarity $MaxS_{a,b}$, and minimum similarity $MinS_{a,b}$ between subsequences $\ell_{a}$ and $\ell_{b}$ are calculated in the same way.

(2) Value-interval similarity.

\textbf{Definition 3: (Value-interval similarity)}.
\textit{
The value interval of a sequence is the difference between the mean value of all peaks of the sequence and the mean value of all valleys of the sequence.
The value-interval similarity $VIS_{a,b}$ of two subsequences $\ell_{a}$ and $\ell_{b}$ refers to the degree of similarity between their value intervals.
The value-interval similarity $VIS_{a,b}$ of $\ell_{a}$ and $\ell_{b}$ is defined in Equation (\ref{eq13}):}

\small{
\begin{equation}
\label{eq13}
VIS_{a,b} =\left( 1 - \frac{\frac{1}{h-1}\sum\limits_{i=1}^{h-1}{(Max_{ai} - Min_{ai})}}{\frac{1}{h-1}\sum\limits_{i=1}^{h-1}{(Max_{bi} - Min_{bi})}} \right) \times 100\%,
\end{equation}}
where $Max_{ai}$ and $Min_{ai}$ are the values of the peaks and valleys of the $i$-th segment in $\ell_{a}$, respectively.

\begin{figure}[!ht]
 \setlength{\abovecaptionskip}{0pt}
 \setlength{\belowcaptionskip}{0pt}
\centering
\includegraphics[width=2.19in]{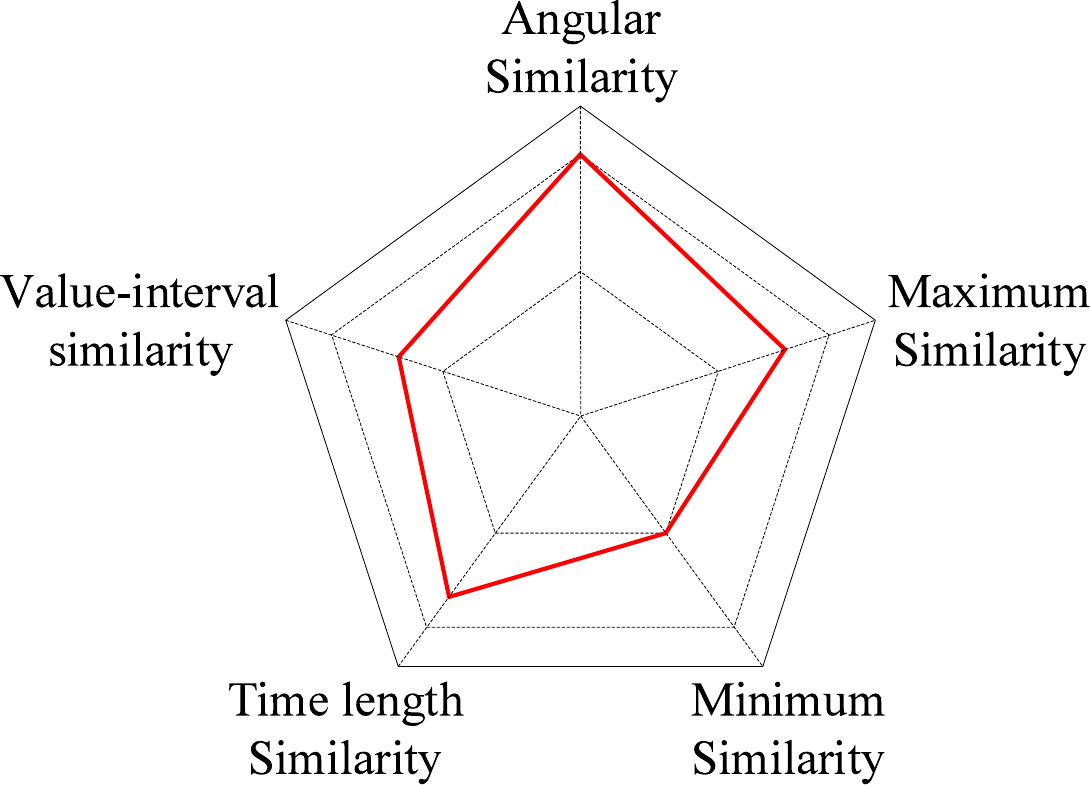}
\caption{Morphological similarity measurement of time-series data.}
\label{fig06}
\end{figure}

Based on the above five similarity indicators, we propose a five-dimension radar chart measurement method to evaluate the morphological similarity of the time-series data abstraction.
The morphological similarity between subsequences $\ell_{a}$ and $\ell_{b}$ is defined as $S_{a,b}$, where each score range of each indicator is (0 $\sim$ 1].
Therefore, as shown in Figure \ref{fig06}, the radar chart of $S_{a,b}$ is plotted as a pentagon, where the distance from the center to each vertex is equal to 1.

According to the radar chart, the value of $S_{a,b}$ is the area composed of the five indicators, as calculated in Equation (\ref{eq14}):

\begin{equation}
\label{eq14}
\begin{aligned}
S_{a,b} = & f(AS_{a,b}, ~TLS_{a,b}, ~MaxS_{a,b}, ~MinS_{a,b}, ~VIS_{a,b})\\
    = & \frac{AS_{a,b} \times TLS_{a,b} + TLS_{a,b} \times MaxS_{a,b}}{2} \times \sin {72^{o}}\\
        & +  \frac{ MaxS_{a,b} \times MinS_{a,b} + MinS_{a,b} \times VIS_{a,b}}{2} \times \sin {72^{o}}\\
    & + \frac{ VIS_{a,b} \times AS_{a,b}}{2} \times \sin {72^{o}}.
\end{aligned}
\end{equation}
It is easy to obtain each side-length of the pentagon is approximately equal to 1.18, and the area of the pentagon is $S_{pentagon} = \frac{(1.18)^{2}}{4} \sqrt{25+10\sqrt{5}} \approx 2.39$.
Hence, the value of $S_{a,b}$ is within the range of ($0 \sim 2.39$).
This novel similarity measure method addresses the problem of inaccurate distance measurement due to the different data shifts and time lengths.

Based on the morphological similarity measurement in Equation (\ref{eq14}), we update the estimated expression of each Fourier spectrum coefficient.
Assuming that $|\ell_{a}|$ is the length of $\ell_{a}$ and $\vartheta$ is the growth step of the comparison subsequences $\ell_{a}$ and $\ell_{b}$, the estimated expression of each Fourier spectrum coefficient is calculated by Equation (\ref{eq15}):
\begin{spacing}{1.7}
\begin{equation}
\label{eq15}
\left\{
\begin{array}{ll}
a_{0} = \frac{1}{m} \sum\limits_{t=1}^{m}{P_{t}},\\
a_{i(a,b)} = \frac{2}{m} \sum\limits_{t=1}^{m}{P_{t}\cos{\left(\frac{2\pi i}{|\ell_{a}+\ell_{b}|}S_{a,b}t + i \vartheta\right)}}, \\
b_{i(a,b)} = \frac{2}{m} \sum\limits_{t=1}^{m}{P_{t}\sin{\left(\frac{2\pi i}{|\ell_{a}+\ell_{b}|}S_{a,b}t + i \vartheta\right)}}. \\
\end{array}
\right.
\end{equation}
\end{spacing}
We calculate the quadratic sum $Q$ of fitting residual sequences $e_{t}$ for each subsequence pair in $P_{t}$ and get the results $\{Q_{1}$, $Q_{2}$, $...$, $Q_{q}\}$, where $q$ is the number of spectrums.
Finally, the optimal period length $L$ of the periodic item $P_{t}$ is the fundamental frequency corresponding to $min{(Q_{1}, ~Q_{2}, ~..., ~Q_{q})}$.

(2) Periodic pattern recognition.

Different from the traditional periodic pattern recognition algorithms, a new method of periodic pattern recognition based on the time-series data abstraction is proposed in this section.
The similarity of the time-series data is calculated by subsequences with the same number of inflection points.
Afterwards, the subsequence with the most similarity is found out as a period of time-series data.

Set $\mu$ as the growth step of the comparison subsequences, namely, there are $\mu$ inflection points increasingly  incorporated into the comparison subsequences each time.
Let $\mu =2$ as an example, that is, 2 inflection points are incorporated into the comparison subsequences each time.
Set $\ell_{a} = \{k_{1}, ~k_{2}\}$ as the first time subsequence and $\ell_{b}=\{k_{2}, ~k_{3}\}$ as a comparison subsequence.
The two subsequences are compared with the morphological similarity measure, which is defined as $S_{P_{1}} = S_{\ell_{1},~\ell_{2}}$. The detail of calculation method has been explained in the previous section.
We continue incorporate the subsequent $\mu$ inflection points into the $\ell_{a}$, namely $\ell_{a} = \{k_{1}, ~k_{2}, ~k_{3}, ~k_{4}\}$.
And then, the same number of inflection points in the data abstraction are collected to compose the comparison subsequence $\ell_{b}$, namely $\ell_{b} = \{k_{4}, ~k_{5}, ~k_{6}, ~k_{7}\}$.

\begin{figure}[!ht]
 \setlength{\abovecaptionskip}{0pt}
 \setlength{\belowcaptionskip}{0pt}
\centering
\includegraphics[width=3.4in]{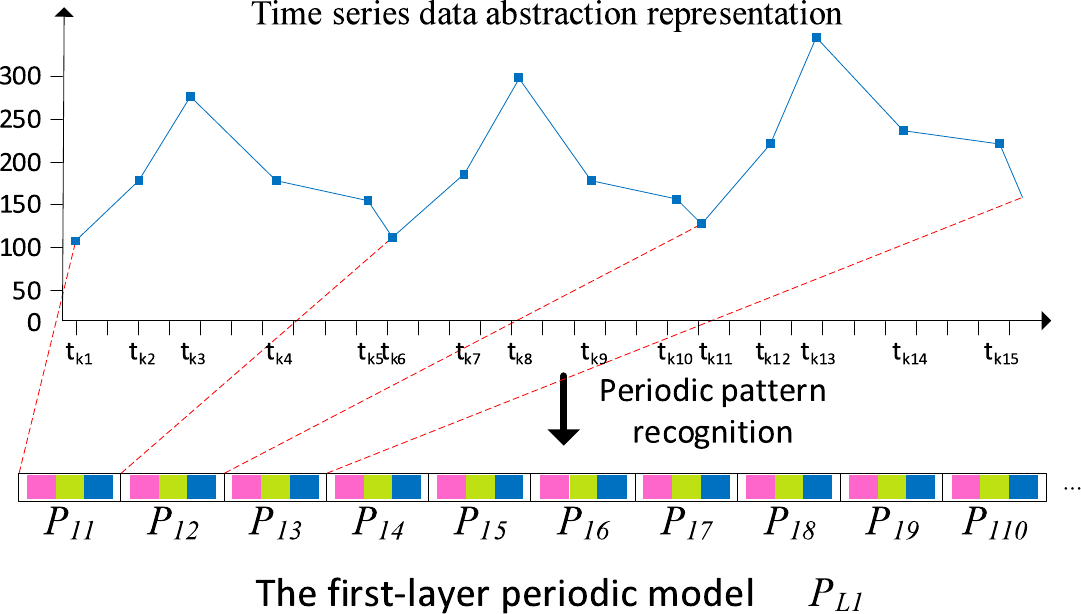}
\caption{The first-layer period model of time-series dataset.}
\label{fig07}
\end{figure}

In addition, the number of inflection points in the comparison subsequences that might exist periodic patterns may be slightly different due to the inflection points marking and the pseudo inflection points deletion operations.
Therefore, we introduce a scaling ratio factor $\varphi$ ($\varphi \in [0,1)$) to control the number of inflection points of the latter comparison subsequence $\ell_{b}$.
In this way, the comparison subsequences are optimized from fixed-length rigid sequences to variable-length flexible sequences.
The length of $\ell_{b}$ is within the range of the left and right extension of the length of the previous comparison subsequence $\ell_{a}$.
Let $n_{a}$ be the number of inflection points of subsequence $\ell_{a}$ and $n_{b}$ be the number of inflection points of subsequences $\ell_{b}$, the scaling ratio factor $\varphi$ is calculated in Equation (\ref{eq16}):
\begin{equation}
\label{eq16}
\lfloor(1 - \varphi) n_{a}\rfloor \leq n_{b} \leq \left\lceil(1 + \varphi) n_{a}\right\rceil.
\end{equation}
For example, assuming that $n_{a} = 6$ and $\varphi = 0.3$, then the value of $n_{b}$ is in the range of ($4 \leq n_{b} \leq  8$).
In other words, for subsequence $\ell_{a}$ with 5 inflection points, $\{4,~5,~6,~7,~8\}$ inflection points closely followed $\ell_{a}$ are taken as the corresponding candidate subsequences $\ell_{b}'$, respectively.
Namely, for subsequence $\ell_{a}$ with 10 inflection points ($\ell_{a}=\{k_{1},~k_{2},~k_{3},~k_{4},~k_{5},~k_{6}\}$), there are 5 different candidate comparison subsequences $\ell_{b}'$ with different numbers of inflection points constructed for similarity measure.
The candidate comparison subsequences $\ell_{b}$ are listed as follows:
$\ell_{b}^{1}=\{k_{6},~k_{7},~k_{8},~k_{9}\}$;
$~~~~$$\ell_{b}^{2}=\{k_{6},~k_{7},~k_{8},~k_{9},~k_{10}\}$;
$~~~~$$\ell_{b}^{3}=\{k_{6},~k_{7},~k_{8},~k_{9},~k_{10},~k_{11}\}$;
$~~~~$$\ell_{b}^{4}=\{k_{6},~k_{7},~k_{8},~k_{9},~k_{10},~k_{11},~k_{12}\}$;
$~~~~$$\ell_{b}^{5}=\{k_{6},~k_{7},~k_{8},~k_{9},~k_{10},~k_{11},~k_{12},~k_{13}\}$.
And each $\ell_{b}'$ is introduced to calculate the similarity with $\ell_{a}$ respectively.
Finally, the candidate subsequence $\ell_{b}'$ with the maximum similarity value is obtained as the comparison subsequence $\ell_{b}$, and the corresponding number of inflection points $n_{b}$ is taken as the length of $\ell_{b}$.
Thus, this pair of comparison subsequences $\ell_{a}$ and $\ell_{b}$ are termed as $S_{P_{i}}$, namely $P_{i} = (\ell_{a} = \{k_{1}, ~k_{2}, ~k_{3}, ~k_{4},~k_{5},~k_{6}\}, ~\ell_{b} = \{k_{6},~k_{7},~k_{8},~k_{9},~k_{10}\})$, where $\ell_{b}$ with 5 inflection points is the comparison subsequence with the highest similarity.
Thus, the first-layer period of time-series dataset $X_{T}$ is recognized using the optimal period length $L$, as described as:
\begin{equation}
\label{eq17}
\centering
P_{L1} = \{P_{11}, P_{12}, ..., P_{1k}\},
\end{equation}
where the length of each period is $L$ ($|P_{li}| = L$).
An example of the first-layer period of time-series dataset $X_{T}$ is shown in Figure \ref{fig07}.
The detailed steps of the FSA-based time series periodic pattern recognition algorithm are presented in Algorithm \ref{alg2}.

\begin{algorithm}[!ht]
\caption{Multi-layer Time series periodic pattern recognition (MTSPPR) algorithm}
\label{alg2}
\begin{algorithmic}[1]
\REQUIRE ~\\
    $X_{K}$: the abstraction of the raw time-series dataset;\\
    $\mu$: the growth step of the comparison subsequences;\\
    $\varphi$: the scaling ratio factor of the comparison subsequence length;\\
\ENSURE ~\\
    $P_{L1}$: the first-layer periodic model of $X_{K}$.
\STATE calculate the non-stationary data model $X_{K} \rightarrow H_{t} + P_{t} + Y_{t}$;
\FOR {each $t$ in $m$}
\STATE $a_{0} \leftarrow \frac{1}{m} \sum\limits_{t=1}^{m}{P_{t}}$;
\FOR {each $i$ in $k$}
\STATE  get subsequence $\ell_{a} = \{k_{j}, ~..., ~k_{j + \mu}\}$ from $P_{t}$;
\STATE  set the length of $\ell_{b}$ $\left\lfloor(1 - \varphi) n_{a}\right\rfloor \leq n_{b} \leq \left\lceil(1 + \varphi) n_{a}\right\rceil$;
\STATE  get comparison subsequence $\ell_{b} = \{k_{j + \mu + 1}, ~..., ~k_{j + \mu + n_{b}}\}$ from $P_{t}$;
\STATE  calculate morphological similarity $S_{a,b} \leftarrow  f(AS_{a,b}, ~TLS_{a,b}, ~MaxS_{a,b}, ~MinS_{a,b}, ~VIS_{a,b})$;
\STATE $a_{i(a,b)} \leftarrow  \frac{2}{m} \sum\limits_{t=1}^{m}{P_{t}\cos{\left(\frac{2\pi i}{|\ell_{a}+\ell_{b}|}S_{a,b}t + i \mu \right)}}$;
\STATE $b_{i(a,b)} \leftarrow  \frac{2}{m} \sum\limits_{t=1}^{m}{P_{t}\sin{\left(\frac{2\pi i}{|\ell_{a}+\ell_{b}|}S_{a,b}t + i \mu\right)}}$;
\ENDFOR
\STATE  calculate the estimate value $P_{t}' \leftarrow a_{0} + \sum_{i=1}^{k}{\left[a_{i}\cos{(i\omega_{0}t)} + b_{i}\sin(i\omega_{0}t)\right]}$;
\STATE calculate the overall variance $S_{P}^{2} \leftarrow \frac{1}{m}\sum_{t=1}^{m}{(P_{t}^{2}-\overline{P_{t}})^{2}}$;
\STATE calculate the spectrum composition $C_{i}^{2} \leftarrow  (a_{i}^{2} + b_{b}^{2})$;
\STATE calculate $\vartheta_{i} \leftarrow  \frac{C_{i}^{2}}{S_{P}^{2}-\frac{1}{m}C_{i}^{}2}$;
\ENDFOR
\STATE find the maximum $i \leftarrow arg ~max(\vartheta_{i})$;
\STATE obtain the period length $L \leftarrow \frac{m}{i}$;
\FOR {$j$ in $\left\lfloor\frac{m}{K}\right\rfloor$}
\STATE obtain period model $P_{1j} \leftarrow X_{L}[j,j*L]$;
\STATE  append period model $P_{L1} \leftarrow P_{1j}$;
\ENDFOR
\RETURN $P_{L1}$.
\end{algorithmic}
\end{algorithm}

In Algorithm \ref{alg2}, $k=\left\lfloor \frac{m}{2} \right\rfloor$ is the highest item of the Fourier spectrums and $m$ is the length of $X_{K}$.
The length of comparison subsequence pairs is increased by the step size of $\mu$.
Assuming that the time complexity of each morphological similarity measurement process is $\lambda$ and the time complexity of the first-layer period recognition is $O(\left\lfloor\frac{m}{K}\right\rfloor)$.
Hence, the computational complexity of Algorithm \ref{alg2} is $O\left(\frac{mk}{2\mu}\lambda + \left\lfloor\frac{m}{K}\right\rfloor\right)$.

\subsubsection{Multi-layer Periodic Pattern Recognition}
\label{section3.2.3}
Considering that there exists potential multi-layer periodicity in given time-series datasets, we propose a multi-layer periodic pattern model to adaptively recognize the multi-layer time periods.
After obtaining the first-layer periodic pattern, the time-series dataset is recognized into multiple periods.
The contents of each period in the first-layer periodic pattern are further abstracted and represented by the Gaussian Blur function.
Let $X_{1j} = \{(x_{1},t_{1}),~...,~(x_{nj},t_{nj})\}$ be the dataset of the $j$-th period $P_{1j}$ in the first-layer periodic model $P_{L1}$, where $j$ is the period length of $P_{1j}$.
We calculate the weight of each data point $x_{i}$ in $X_{1j}$ using the Gaussian Blur function, as defined in Equation (\ref{eq18}):

\begin{equation}
\label{eq18}
\begin{aligned}
w_{i} &= G(x_{i},t_{i})\\
       &= \frac{1}{2\pi\sigma^{2}}e^{-\frac{x_{i}^{2}+t_{i}^{2}}{2\sigma^{2}}},
\end{aligned}
\end{equation}
where $\sigma$ is the variance of all data points in $X_{1j}$.
Based on $w_{i}$, we obtain the new value $(x_{i}' = x_{i}\times w_{i}, ~t_{i}' = t_{i} \times w_{i})$.
In this way, the dataset $X_{1j}$ is updated as $X_{1j}' = \{(x_{1}',t_{1}'),~...,~(x_{nj}',t_{nj}')\}$.

\begin{figure}[!ht]
 \setlength{\abovecaptionskip}{0pt}
 \setlength{\belowcaptionskip}{0pt}
\centering
\includegraphics[width=3.4in]{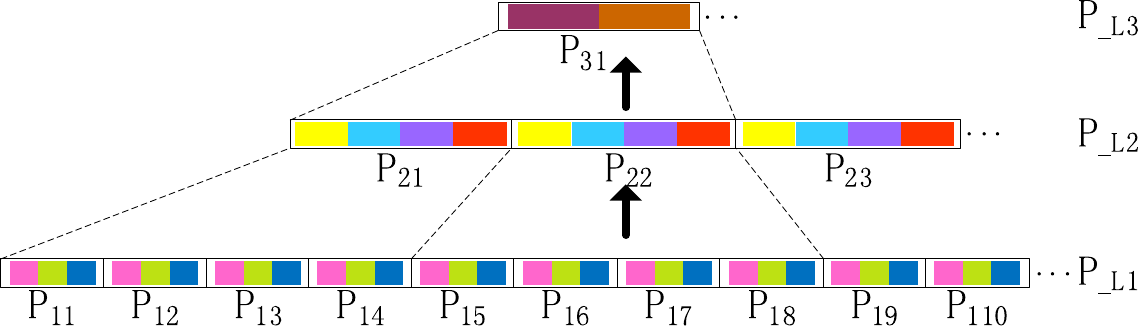}
\caption{Multi-layer periodic model of time-series dataset.}
\label{fig08}
\end{figure}

For the updated dataset $X_{1j}'$, we apply the big data compression and abstraction method on $X_{1j}$ to further reduce the volume of each period and extract the key information.
Then, the FSA-based periodic pattern recognition algorithm is used on the compressed first-layer dataset to obtain the second-layer periodic patterns.
Repeat these steps until there is no significant periodic pattern can be recognized.
Thus, the multi-layer periodic model of the time-series dataset is built, as defined as:
\begin{center}
$P = \{P_{L1}, P_{L2}, ..., P_{Lq}\}$,
\end{center}
where $q$ is the number of period layers for the time-series dataset.
An example of the multi-layer periodic model of a given time-series dataset is shown in Figure \ref{fig08}.

\subsection{Periodicity-based Time Series Prediction}
\label{section3.3}
\begin{figure}[!ht]
 \setlength{\abovecaptionskip}{0pt}
 \setlength{\belowcaptionskip}{0pt}
\centering
\includegraphics[height=2.3in,width=3.5in]{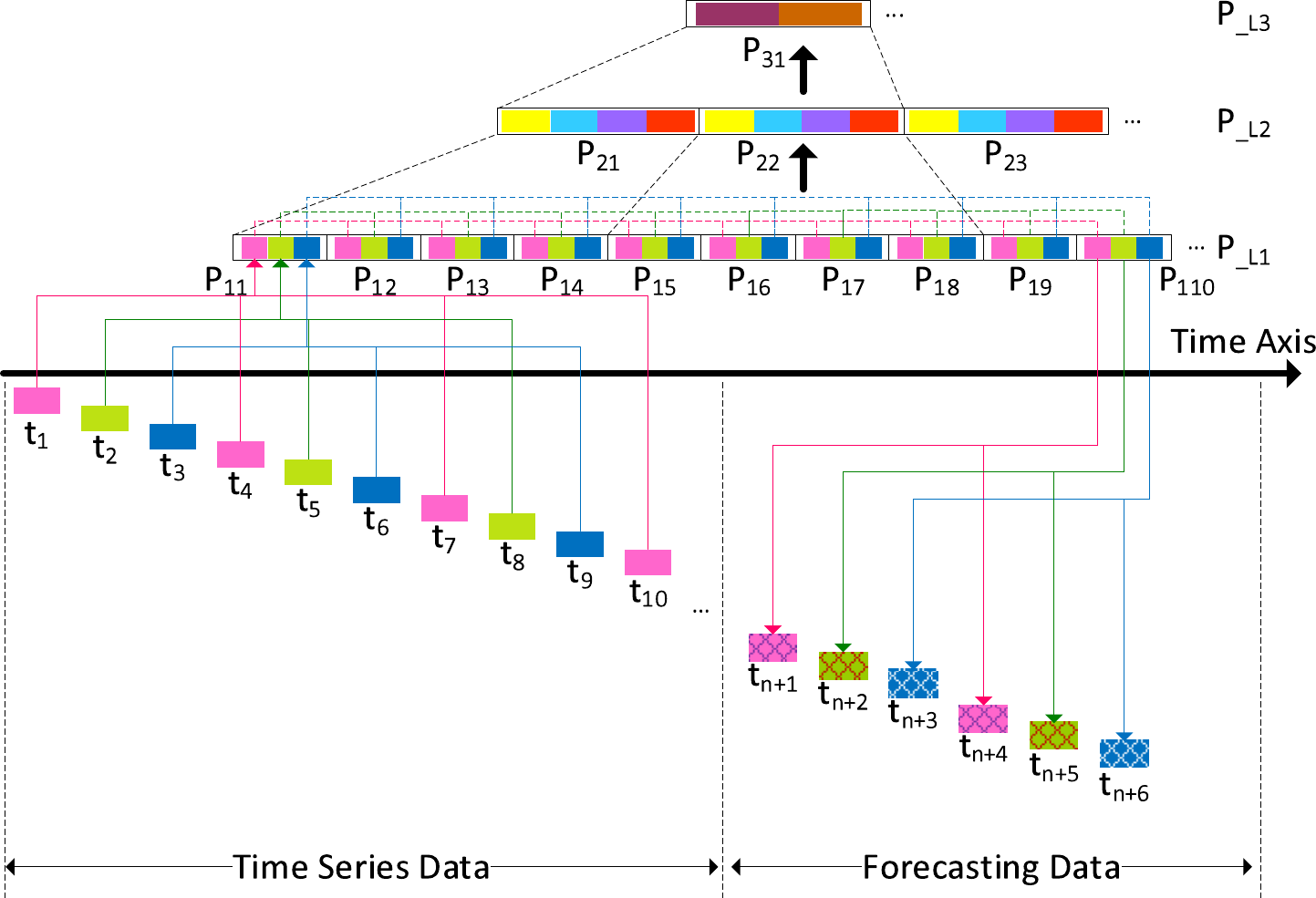}
\caption{Periodicity-based time series prediction method.}
\label{fig09}
\end{figure}

Based on the multi-layer periodic model described in Section \ref{section3.2}, we propose a Periodicity-based Time Series Prediction (PTSP) algorithm in this section.
Different from the traditional time series prediction methods, in PTSP, the forecasting unit of upcoming data is one complete period rather than one timestamp.
According to the identified periodic models, the forecasting object of each prediction behavior is the contents of the next complete period, instead of the data point in the next timestamp.
The previous periodic models in different layers involve different contributions to the contents of the coming period.
The periodicity-based time series prediction method is shown in Figure \ref{fig09}.

(1) Prediction based on periodic model.

For each previous periodic model, its impact on the contents of the coming period is measured by a weight value, which is calculated using the time attenuation factor.
Given a multi-layer periodic model $P = \{P_{L1}, P_{L2}, ..., P_{Lq}\}$ for the time-series dataset, there are multiple period models $P_{La} = \{P_{a1}, P_{a2}, ..., P_{ak_{a}}\}$ in each layer, where $q$ is the number of period layers and $k_{a}$ is the number of periods in the $a$-th layer.
Assuming that $P_{1t}$ is the current time period and $P_{1t+1}$ is the next time period that will be predicted.
The contents of $P_{1t+1}$ are predicted based on all of the periodic models in each layer in the identified multi-layer periodic model.
To evaluate the impact of each previous model on the contents of $P_{1t+1}$, a time attenuation factor is introduced to calculate the weight value of each periodic model in each layer, respectively.
For example, for each period $P_{1i}$ in the first layer, the weight of $P_{1i}$ for $P_{1t+1}$ is defined in Equation (\ref{eq19}):

\begin{equation}
\label{eq19}
w_{1i} = \frac{e^{\frac{t-i}{t}}}{\sum_{j=1}^{i}{e^{\frac{t-i}{t}}}}.
\end{equation}
Based on the weights of periodic models in the first layer $P_{L1}$, we can calculate the $P_{1t+1}$ prediction component $P_{1t+1}^{(1)}$ from $P_{L1}$, as defined in Equation (\ref{eq20}):

\begin{equation}
\label{eq20}
P_{1t+1}^{(1)} = \sum_{i=1}^{t}{(w_{i} \times P_{1i})}.
\end{equation}

\begin{figure}[!ht]
 \setlength{\abovecaptionskip}{0pt}
 \setlength{\belowcaptionskip}{0pt}
\centering
\includegraphics[width=4.5in]{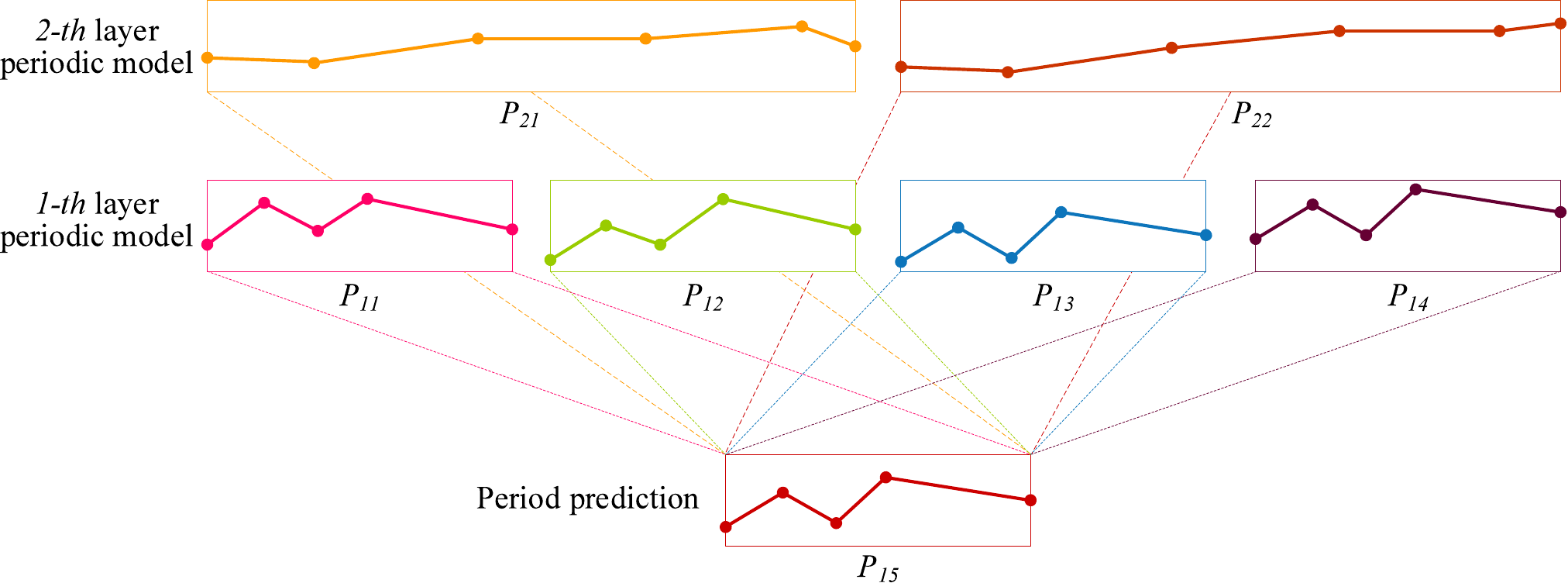}
\caption{Example of the periodicity-based time series prediction process.}
\label{fig10}
\end{figure}

We continue to calculate the weight of periods in each layer $P_{La}$ for $P_{1t+1}$.
Assuming that there are $|a|$ periods in the $a$-th layer model $P_{La}$, namely, the current period in the $a$-th layer is $P_{a|a|}$, which corresponds to the current period $P_{1t+1}$.
For each period $P_{ai}$ in the $a$-th layer, we use the time attenuation factor to measure the weight of $P_{ai}$ for $P_{1t+1}$, as calculated in Equation (\ref{eq21}):

\begin{equation}
\label{eq21}
w_{ai} = \frac{e^{\frac{|a|-i}{|a|}}}{\sum_{j=1}^{i}{e^{\frac{|a|-i}{|a|}}}}.
\end{equation}
Based on each prediction component $P_{1t+1}^{(a)}$ calculated from each layer $P_{La}$, we get the predicted contents of $P_{1t+1}$, as defined in Equation (\ref{eq22}):

\begin{equation}
\label{eq22}
P_{1t+1} = \sum_{a=1}^{q}{P_{1t+1}^{(a)}} = \sum_{a=1}^{q}{\sum_{i=1}^{|a|}{(w_{ai} \times P_{ai})}}.
\end{equation}
An example of the periodicity-based time series prediction process is illustrated in Figure \ref{fig10}.

(2) Calculation of the inflection points.

Due to the big data compression and abstraction, each periodic model in the identified multi-layer periodic model is built based on the inflection points rather than the raw time-series datasets.
In this way, the predicted contents of the next time period are inflection points with the corresponding time points, rather than the data values at all time points.
Therefore, we should calculate the values of all inflection points in $P_{1t+1}$ and further fit the data values at all time points.

Considering that different periods in each layer $P_{La}$ contain different numbers of inflection points located at different time points, we need to map them to the corresponding positions on the time axis of the predicting period to form new predicting inflection points.
The set of predicting inflection points in $P_{1t+1}$ is defined as $X_{K(1t+1)}$, there are multiple prediction components from all periods to form the values of $X_{K(1t+1)}$.
Assuming that there is a set of inflection points $X_{K(ai)} = \{(k_{1}, t_{k1}), ~..., ~(k_{m}, t_{km})\}$ in the period $P_{ai}$ in $P_{La}$, we calculate the prediction component of $X_{K(ai)}$ for $X_{K(1t+1)}$, as defined in Equation (\ref{eq23}):
\begin{equation}
\label{eq23}
\begin{aligned}
X_{K(1t+1)}^{(ai)} &= w_{ai} \times X_{K(ai)}\\
\left[
\begin{array}{c}
(k_{1}',t_{k1})\\
(k_{2}',t_{k2})\\
...\\
(k_{m}',t_{km})\\
\end{array}
\right]
&= w_{ai}  \times
\left[
\begin{array}{c}
(k_{1},t_{k1})\\
(k_{2},t_{k2})\\
...\\
(k_{m},t_{km})\\
\end{array}
\right].
\end{aligned}
\end{equation}

According to Equation (\ref{eq23}), the set of predicting inflection points $X_{K(1t+1)}$ in $P_{1t+1}$ is integrated based on the prediction components $X_{K(1t+1)}^{(ai)}$ from all previous periodic models.

(3) Fit data values at all time points in the predicting period $P_{1t+1}$.

\begin{figure}[htbp]
 \setlength{\abovecaptionskip}{0pt}
 \setlength{\belowcaptionskip}{0pt}
\centerline{\includegraphics[width=1.8in]{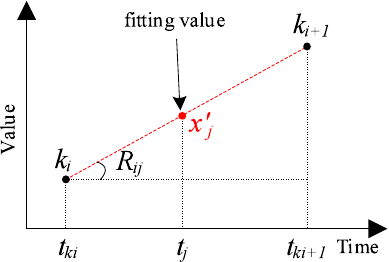}}
\caption{Calculation of fitting data points between inflection points.}
\label{fig11}
\end{figure}

Based on the predicted inflection points $X_{K(1t+1)}$, the data values at all time points among these inflection points are fitted.
For each two adjacent inflection points $(k_{i},t_{ki})$ and $(k_{i+1},t_{ki+1})$ in $X_{K(1t+1)}$, the fitting data value $x_{j}'$ at each time point $t_{j}$ in the range of $(t_{ki}, t_{ki+1})$ is calculated in Equation (\ref{eq24}):

\begin{equation}
\label{eq24}
x_{j}' = \frac{k_{i+1} - k_{i}}{t_{ki+1} - t_{ki}} \times (t_{j} - t_{ki}) + k_{i}.
\end{equation}
In this way, we obtain the predicted values of each time point in the coming period $P_{1t+1}$.
An example of the calculation of fitted data points between two adjacent inflection points is shown in Figure \ref{fig11}.
The detailed steps of the periodicity-based time series prediction algorithm are presented in Algorithm \ref{alg3}.

\begin{algorithm}[!ht]
\caption{Periodicity-based time series prediction (PTSP) algorithm}
\label{alg3}
\begin{algorithmic}[1]
\REQUIRE ~\\
   $P$: the trained multi-layer periodic model $P = \{P_{L1}, P_{L2}, ..., P_{Lq}\}$;\\
\ENSURE ~\\
    $P_{1t+1}$: the prediction of the coming period.
\FOR {each layer $P_{La}$ in $P$}
\FOR {each period model $P_{ai}$ in $P_{La}$}
\STATE  calculate the time attenuation factor $w_{ai} \leftarrow \frac{e^{\frac{|a|-i}{|a|}}}{\sum_{j=1}^{i}{e^{\frac{|a|-i}{|a|}}}}$;
\STATE  predict $P_{1t+1}  \leftarrow \sum_{a=1}^{q}{\sum_{i=1}^{|a|}{(w_{ai} \times P_{ai})}}$, $X_{K(1t+1)}^{(ai)}  \leftarrow w_{ai} \times X_{K(ai)}$;
\ENDFOR
\ENDFOR
\FOR{each inflection point $k_{i}$ in $X_{K(1t+1)}$}
\FOR{each unit time $j$ in $(t_{ki}, t_{ki+1})$}
\STATE  calculate the fitting data $x_{j}' \leftarrow \frac{k_{i+1} - k_{i}}{t_{ki+1} - t_{ki}} \times (t_{j} - t_{ki}) + k_{i}$;
\STATE append $x_{j}'$ to $P_{1t+1}$;
\ENDFOR
\ENDFOR
\RETURN $P_{1t+1}$.
\end{algorithmic}
\end{algorithm}

In Algorithm \ref{alg3}, let $q$ be the number of layers of the multi-layer periodic model, $\overline{|a|}$ be the average number of periods in each layer $P_{La}$, and $|P_{1t+1}|$  be the length of the prediction period $P_{1t+1}$.
Assuming that the number of inflection points in $P_{1t+1}$ is $m$ and the length of the unit time is $u$, there are $\left\lfloor\frac{|P_{1t+1}|}{u}\right\rfloor$ data points need to fit.
Hence, the computation complexity of Algorithm \ref{alg3} is $O\left(q\overline{|a|} + \left\lfloor\frac{|P_{1t+1}|}{u}\right\rfloor\right)$.

\section{Parallel Implementation of the Proposed Algorithms}
\label{section4}
To efficiently handle the large-scale time-series datasets and improve the performance of time series periodic pattern recognition and prediction algorithms, we propose parallel solutions for the proposed TSDCA, MTSPPR, and PTSP algorithms on the Apache Spark cloud computing platform.
The parallel execution process of the proposed algorithms based on the Apache Spark platform is introduced in Section \ref{section4.1}.
The parallel solutions of the proposed TSDCA, MTSPPR, and PTSP algorithms are provided in Sections \ref{section4.2}, \ref{section4.3}, and \ref{section4.4}, respectively.

\subsection{PPTSP Architecture on Apache Spark}
\label{section4.1}
\begin{figure}[!ht]
 \setlength{\abovecaptionskip}{0pt}
 \setlength{\belowcaptionskip}{0pt}
\centering
\includegraphics[height =3.3in,width=3.5in]{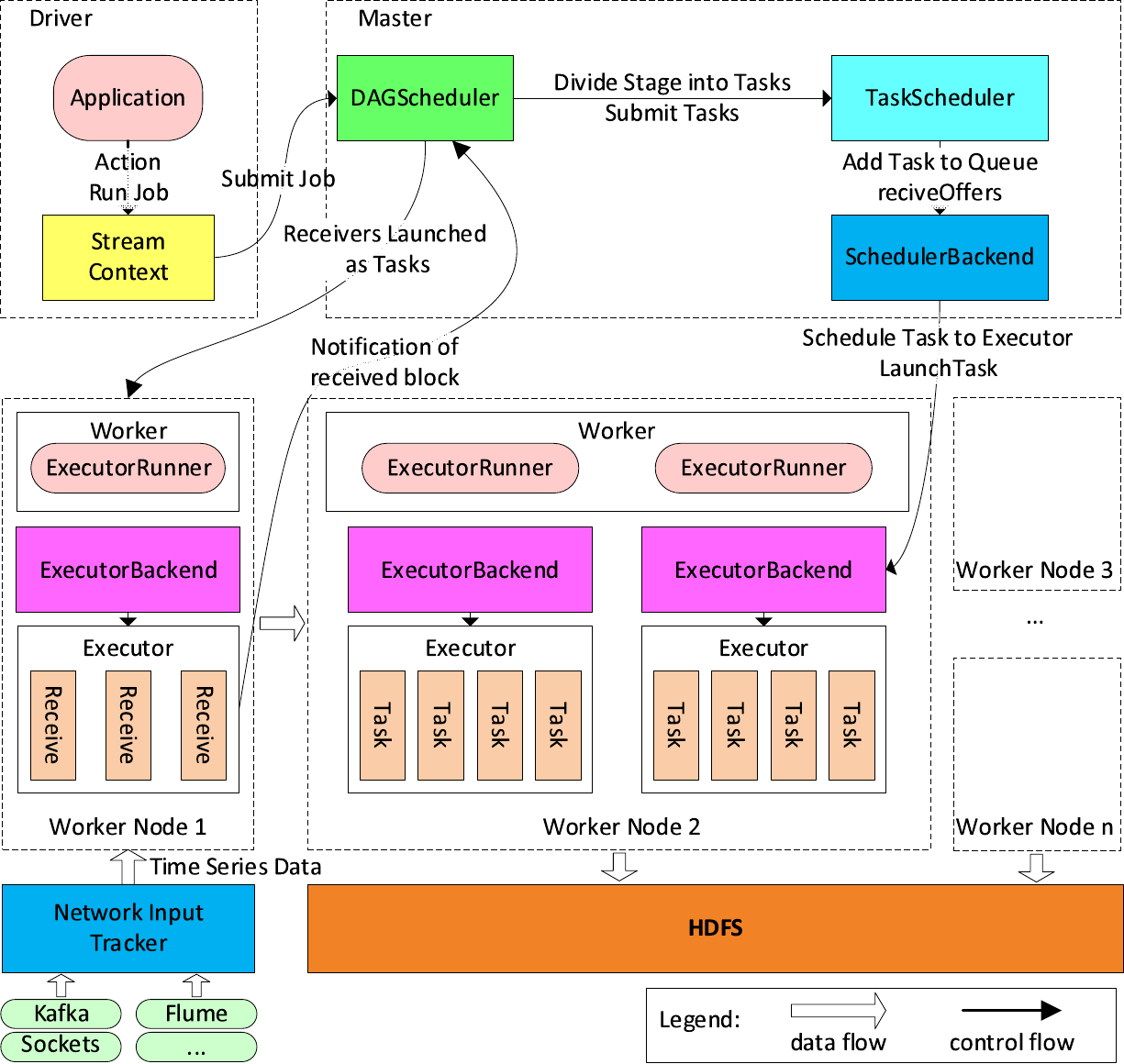}
\caption{PPTSP architecture on Spark Streaming platform.}
\label{fig12}
\end{figure}

The Periodicity-based Parallel Time Series Prediction (PPTSP) algorithm is implemented on the Apache Spark, using the Streaming and RDD modules.
The Spark Streaming module is a real-time computing framework, which extends the storage capacity to handle large-scale streaming and time-series datasets.
The Spark Streaming module is suitable for various applications, including data analysis and mining with the combination of historical and real-time datasets.
In our work, an Apache Spark cluster is equipped with a driver computer, an application master node and multiple worker nodes.
Then, a Spark Streaming module is deployed on the Spark cluster.
The architecture of the proposed PPTSP algorithm on Spark Streaming platform is shown in Figure \ref{fig12}.

The main workflow of the PPTSP algorithm is described as follows.

(1) Monitor and receive time-series datasets.

We collect time-series datasets from practical applications at a fixed time frequency and send them to the system in a streaming way.
After starting the Apache Spark cluster, we submit the PPTSP program to the master node and create a stream listener to monitor the incoming time-series data stream.
Stream listeners directly support various data sources, such as Kafka, Flume, Twitter, ZeroMQ, and sockets TCP.

(2) Receive input data streams and generate data blocks.

Once the input data streams are received, they are divided into a series of data blocks at predetermined time intervals.
Then, each data block is stored as an RDD object for further calculations on the Spark Streaming framework.

(3) Calculate logical dependencies and data dependencies.

There are three main processes in the PPTSP program:
(1) the data compression and abstraction process,
(2) the multi-layer periodic pattern recognition process,
and (3) the periodicity-based time series prediction process.
Each process has the different calculation flow and logic dependencies, as shown in algorithms \ref{alg1}, \ref{alg2}, and \ref{alg3}, respectively.
Therefore, in each process, the logical and data dependencies of each data block in the form of RDD are analyzed as the corresponding RDD dependencies.

(4) Create jobs and tasks according to RDD dependencies.

Three Spark jobs are created for the three processes in the PPTSP program, respectively.
In each job, an RDD dependency graph is built based on RDD dependencies to generate the corresponding stages and tasks, using different types of RDD computing operations.
RDD objects support two kinds of computing operations: transformation and action operations.
Transformation operations include a series of functions, such as $map()$, $filter()$, $flatMap()$, $mapPartitions()$, $union()$, and $join()$.
In each transformation function, calculations are performed on the input RDD object and a new RDD object is created from the original one.
In addition, action operations include a series of functions, such as $reduce()$, $collect()$, $count()$, $saveAsHadoopFile()$, and $countBykey()$.
In each action function, calculations are executed on the input RDD object to get a result and callback to the driver program or save it to an external storage system.

(5) Schedule tasks for parallel execution.

A task Directed Acyclic Graph (DAG) is established for each job based on the RDD dependencies of the PTSPP program.
In each task DAG, multiple job stages are detected according to the transformation and action operations.
For each RDD action operation, a job stage is detected to generate several computing tasks.
In this way, each job may be split into multiple stages, and tasks in the same stage without RDD dependency can be executed in parallel.
These tasks are then submitted to the task scheduler and scheduled to the different work executors for parallel execution.

\subsection{Parallelization of the TSDCA Process}
\label{section4.2}
In this section, the time-series data compression and abstraction process of the PTSPP program is parallelized in the Spark Streaming framework.
To be able to batch process the received data stream, the input time-series data stream is split into a series of datasets using a sliding time-window method.
Datasets in each sliding time-window are saved as a Discretized Stream (DStream) and then converted to the corresponding RDD object.
In this way, operations on the DStream objects are conducted using the transformation or action functions of RDD objects. The input time-series data stream reception process based on the Spark Streaming framework is presented in Figure 13.

\begin{figure}[!ht]
 \setlength{\abovecaptionskip}{0pt}
 \setlength{\belowcaptionskip}{0pt}
\centering
\includegraphics[width=3.0in]{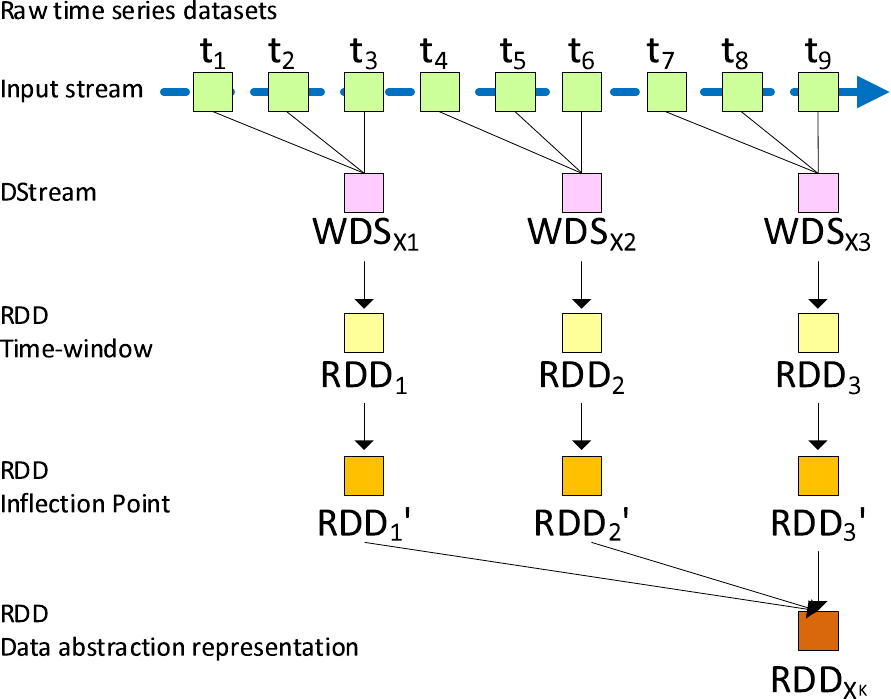}
\caption{Time-series data stream reception process based on the Spark Streaming freamwork.}
\label{fig13}
\end{figure}

\begin{figure}[!ht]
 \setlength{\abovecaptionskip}{0pt}
 \setlength{\belowcaptionskip}{0pt}
\centering
\includegraphics[width=4.0in]{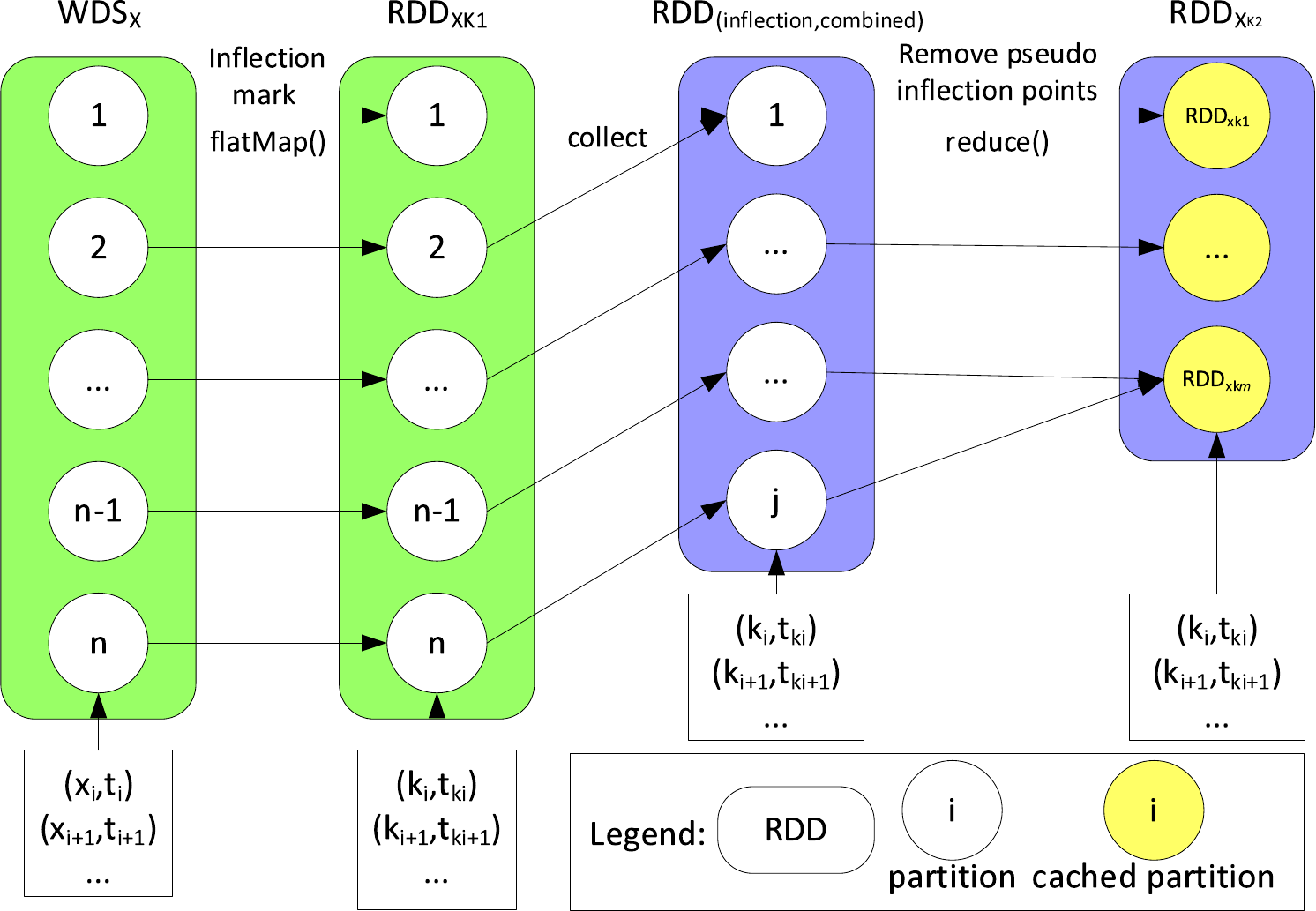}
\caption{RDD dependency graph of time-series data compression and abstraction.}
\label{fig14}
\end{figure}

As shown in Figure \ref{fig13}, based on the sliding time-window method, the received raw time-series data stream is buffered and split into a series of DStream objects at time intervals, e.g., $\{WDS_{X1}, ~WDS_{X2}, ~WDS_{X3}, ~...\}$.
Then, each DStream object $WDS_{Xi}$ is converted to an RDD object $RDD_{i}$ and written to the block manager for subsequent processing.
In addition, we create an RDD dependency graph for these RDD objects, and compute the logic and data dependencies between different RDD objects according to the workflow of the time-series data compression and abstraction process.
The RDD dependency graph of the time-series data compression and abstraction process is shown in Figure \ref{fig14}.

In Figure \ref{fig14}, suppose that there are $n$ data partitions in a DStream object $WDS_{X}$, and each partition $P_{(WDS_{X},i)}$ contains multiple time-series records $\{(x_{i}, t_{i}), ~(x_{i+1}, t_{i+1}), ~...\}$ with adjacent time stamps.
There are two kinds of dependencies between RDD objects: the narrow dependency and wide dependency.
A narrow dependency means that each partition of a parent RDD is used only by one partition of the child RDD objects.
Each partition of the child RDD object usually depends on a constant number of partitions of its parent RDD object.
In contrast, a wide dependency means that each partition of a parent RDD may be used by multiple partitions of the corresponding child RDD objects.
And each partition of the child RDD object usually depends on all the partitions of its parent RDD object.
Records in each partition $P_{(WDS_{X},i)}$ in $WDS_{X}$ are calculated and the inflection points are detected to generate the corresponding new partition of $RDD_{X_{K1}}$.
The records in each partition are independent of other partitions.
In such a case, a narrow dependency occurs between each partition of RDD objects $DS_{X}$ and $RDD_{X_{K1}}$.
Then, the records in each partition of $DS_{X}$ are calculated in parallel, without the synchronization wait constraints as well as causing data communication overhead.

After obtaining each partition of $RDD_{X_{K1}}$, the inflection points in $RDD_{X_{K1}}$ are combined to generate a new RDD object $RDD_{(inflection,combined)}$.
Based on $RDD_{(inflection,combined)}$, the pseudo-inflection points in $RDD_{(inflection,combined)}$ are further detected and removed.
In this process, for partitions in the RDD object $RDD_{(inflection,combined)}$, records in the edge regions of each partition may depend on the records in neighboring partitions to identify the potential pseudo-inflection points.
Namely, for each new partition in $RDD_{X_{K2}}$, it depends on one or more partitions in $RDD_{(inflection,combined)}$.
Therefore, a narrow dependency occurs between partitions of RDD objects $RDD_{(inflection,combined)}$ and $RDD_{X_{K2}}$.
The RDD object $RDD_{X_{K}}$ cached into memory system for the subsequent processing, while the raw RDD object $WDS_{X}$ is deserted.
The detailed steps of the parallel implementation of time-series data compression and abstraction process are presented in Algorithm \ref{alg41}.

\begin{algorithm}[!ht]
\caption{Parallel implementation of time-series data compression and abstraction algorithm (P-TSDCA)}
\label{alg41}
\begin{algorithmic}[1]
\REQUIRE ~\\
    $port$: the TCP port for the input streaming datasets monitoring and receiving;\\
    $L_{receive}$: the interval of the input streaming datasets monitoring and receiving;\\
    $L_{sliding}$: the sliding interval	of sliding window operation;\\
    $L_{window}$: the length of the time-window;\\
\ENSURE ~\\
    $RDD_{X_{K}}$: the data abstraction of $X$.
\STATE $conf \leftarrow$  create SparkConf().setMaster(``$master$'').setAppName(``$P-TSDCA$'');
\STATE $ssc \leftarrow$ create StreamingContext($conf$, Seconds(1));	
\STATE $DS_{X} \leftarrow$ $ssc$.socketTextStream($ssc$, $port$);
\STATE $WDS_{X}	\leftarrow$	$DS_{X}$.window($L_{window}$, $L_{sliding}$).countByValue();
\STATE $RDD_{X_{K}}$ $\leftarrow$ $WDS_{X}$.\textbf{flatMap}
\STATE \quad  {each data point ${x_{i}}$ in $WDS_{X}$}:
\STATE  \quad calculate inclination rates $r_{i,i+1} \leftarrow \frac{x_{i+1} - x_{i}}{t_{i+1} - t_{i}}$, $r_{i+1,i+2} \leftarrow \frac{x_{i+2} - x_{i+1}}{t_{i+2} - t_{i+1}}$;
\STATE \quad \textbf{if} {($r_{i,i+1} \times r_{i+1,i+2} \leq 0$)} \textbf{then} mark inflection point $RDD_{X_{K}} \leftarrow (k_{j} \leftarrow x_{i+1}, ~t_{k_{j}} \leftarrow t_{i+1})$;
\STATE \textbf{endMap}.collect();
\STATE $RDD_{X_{K}} \leftarrow$ $sc$.parallelize(1 to $m$, $RDD_{X_{K}}$).\textbf{map}
\STATE \quad each inflection point $k_{j}$ in $RDD_{X_{K}}$:
\STATE \quad calculate inclination rates $R_{j,j+1} \leftarrow \frac{k_{j+1} - k_{i}}{t_{k_{j+1}} - t_{k_{j}}}$, $R_{j+1,j+2} \leftarrow \frac{k_{j+2} - k_{i+1}}{t_{k_{j+2}} - t_{k_{j+1}}}$;
\STATE \quad  calculate inclination relationship of value $S_{R}(i,i+2) \leftarrow \frac{R_{i,i+2}}{max(R_{i,i+1}, ~R_{i+1,i+2})}$;
\STATE \quad calculate inclination relationship of time $S_{T}(i,i+2) \leftarrow \frac{t_{k_{i+2}} - t_{k_{i}}}{max(t_{k_{i+1}}-t_{k_{i}},~t_{k_{i+2}} - t_{k_{i+1}})}$;
\STATE \quad \textbf{if} {($S_{R}(i,i+2) \geq \delta$ and $S_{T}(i,i+2)  \geq \varepsilon$)} \textbf{then} remove pseudo-inflection point $k_{j+1}$ from $X_{K}$;
\STATE \textbf{endmap}.groupBykey().reduce();
\STATE $ssc$.start();
\STATE $ssc$.awaitTermination();
\RETURN $RDD_{X_{K}}$.
\end{algorithmic}
\end{algorithm}

In Algorithm \ref{alg41}, we initially import the core function libraries of the Apache Spark Streaming and RDD modules, such as \emph{SparkContext} and \emph{StreamingContext}.
Based on the imported function libraries, a \emph{StreamingContext} object named \emph{ssc} is created with a batch interval of 1.0 seconds.
Using the context \emph{ssc}, the program monitors and receives the input streaming datasets through a TCP port (lines 2-3).
A DStream object $DS_{X}$ is created to store the input streaming datasets.
The sliding window operation is used to split the input streams $DS_{X}$ into a plurality of DStream objects $WDS_{X}$ for batch execution.
We set the length of the sliding windows and the corresponding sliding interval using the parameters $L_{sliding}$ and $L_{window}$ (lines 3-4).
For the DStream object $WDS_{X}$ in each time window, parallel execution is performed by a $flatMap()$ function, which is an one-to-many DStream operation.
In the $flatMap()$ function, a new DStream object is created by generating multiple new partitions from each partition in the source DStream object.
Then, inclination rates of the data points in each partition are calculated in parallel (lines 5-9).
After obtaining the inclination rates, an RDD object $RDD_{X_{K}}$ is generated for the time-series data abstraction.
In addition, we use a $map()$ parallel function to detect and remove pseudo-inflection points from $RDD_{X_{K}}$, where the inclination relationships of the value and time of each data point are calculated, respectively.
In the $groupBykey()$ and $reduce()$ functions, the results of multiple RDD objects are combined to generate the final RDD object $RDD_{X_{K}}$ for downstream analysis (lines 10-16).

\subsection{Parallelization of the MTSPPR Process}
\label{section4.3}
\begin{figure}[!ht]
 \setlength{\abovecaptionskip}{0pt}
 \setlength{\belowcaptionskip}{0pt}
\centering
\includegraphics[width=3.5in]{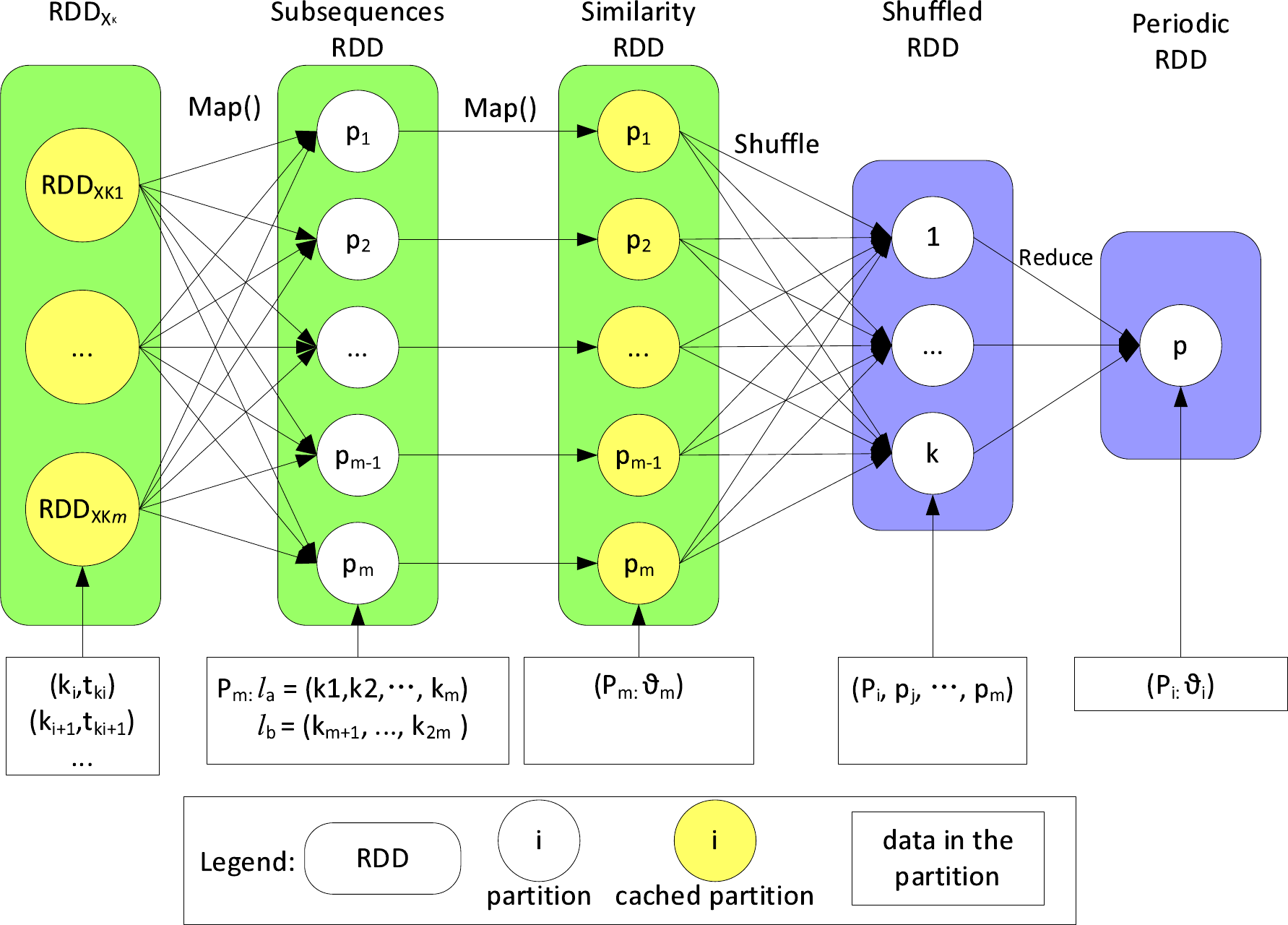}
\caption{RDD dependency graph of the multi-layer time series periodic pattern recognition.}
\label{fig15}
\end{figure}

Based on the results of time-series data compression and abstraction, the process of multi-layer time series periodic pattern recognition (MTSPPR) is performed in parallel.
The cached RDD object $RDD_{X_{K}}$ is reloaded into the Spark platform and each partition of $RDD_{X_{K}}$ is mapped into multiple time subsequences for parallel execution.
Similar to the P-TSDCA process, an RDD dependency graph is constructed for the operation of RDD objects in the P-MTSPPR process.
The logical and data dependencies among different RDD objects are considered and computed for the downstream parallel task decomposition and scheduling.
The RDD dependency graph of the multi-layer time series periodic pattern recognition (P-MTSPPR) process is illustrated in Figure \ref{fig15}.

According to the RDD dependency graph, we use a \emph{TaskSchedule} module of the Spark cloud computing platform to execute the P-MTSPPR process.
The detailed steps of the parallel process of multi-layer time series periodic pattern recognition are presented in Algorithm \ref{alg42}.

\begin{algorithm}[!ht]
\caption{Parallel implementation of multi-layer time series periodic pattern recognition process (P-MTSPPR)}
\label{alg42}
\begin{algorithmic}[1]
\REQUIRE ~\\
    $RDD_{X_{K}}$: the RDD object of the abstraction dataset;\\
    $\mu$: the growth step of the comparison subsequences;\\
    $\varphi$: the scaling ratio factor of the comparison subsequence length;\\
\ENSURE ~\\
    $RDD_{P_{L1}}$: the RDD object of the first-layer periodic model.
\STATE $conf \leftarrow$  create SparkConf().setMaster(``$master$'').setAppName(``$P-MTSPPR$'');
\STATE $sc \leftarrow$ create SparkContext($conf$);	
\STATE calculate the non-stationary data model $RDD_{X_{K}} \rightarrow RDD_{H_{t}} + RDD_{P_{t}} + RDD_{Y_{t}}$;
\STATE $RDD_{\mu} \leftarrow$ $sc$.parallelize(1 to $m$, $RDD_{X_{K}}$).\textbf{map}
\STATE  \quad $a_{0} \leftarrow \frac{1}{m} \sum\limits_{t=1}^{m}{P_{t}}$;
\STATE  \quad $RDD_{similarity} \leftarrow$ $sc$.parallelize($i$ in $k$).\textbf{map}
\STATE  \qquad get subsequence $\ell_{a} = \{k_{j}, ~..., ~k_{j + \mu}\}$ from $P_{t}$;
\STATE  \qquad set the length of $\ell_{b}$ $\left\lfloor(1 - \varphi) n_{a}\right\rfloor \leq n_{b} \leq \left\lceil(1 + \varphi) n_{a}\right\rceil$;
\STATE  \qquad get comparison subsequence $\ell_{b} = \{k_{j + \mu + 1}, ~..., ~k_{j + \mu + n_{b}}\}$ from $P_{t}$;
\STATE  \qquad calculate morphological similarity $S_{a,b} \leftarrow  f(AS_{a,b}, ~TLS_{a,b}, ~MaxS_{a,b}, ~MinS_{a,b}, ~VIS_{a,b})$;
\STATE \qquad $a_{i(a,b)} \leftarrow  \frac{2}{m} \sum\limits_{t=1}^{m}{P_{t}\cos{\left(\frac{2\pi i}{|\ell_{a}+\ell_{b}|}S_{a,b}t + i \mu\right)}}$, $b_{i(a,b)} \leftarrow  \frac{2}{m} \sum\limits_{t=1}^{m}{P_{t}\sin{\left(\frac{2\pi i}{|\ell_{a}+\ell_{b}|}S_{a,b}t + i \mu\right)}}$;
\STATE \quad \textbf{endmap}.reduce();
\STATE \quad calculate the estimate value $P_{t}' \leftarrow a_{0} + \sum_{i=1}^{k}{\left[a_{i}\cos{(i\omega_{0}t)} + b_{i}\sin(i\omega_{0}t)\right]}$;
\STATE \quad calculate the overall variance $S_{P}^{2} \leftarrow \frac{1}{m}\sum_{t=1}^{m}{(P_{t}^{2}-\overline{P_{t}})^{2}}$;
\STATE \quad calculate the spectrum composition $C_{i}^{2} \leftarrow  (a_{i}^{2} + b_{b}^{2})$;
\STATE \quad  calculate $\vartheta_{i} \leftarrow  \frac{C_{i}^{2}}{S_{P}^{2}-\frac{1}{m}C_{i}^{}2}$;
\STATE \textbf{endmap}.groupBykey();
\STATE find the maximum $i \leftarrow arg ~max(\vartheta_{i})$;
\STATE obtain the period length $L \leftarrow \frac{m}{i}$;
\STATE $RDD_{P_{L1}} \leftarrow$ $sc$.parallelize{$j$ in $\left\lfloor\frac{m}{K}\right\rfloor$}.\textbf{flatMap}
\STATE \quad obtain period model $P_{1j} \leftarrow X_{L}[j,j*L]$;
\STATE \quad append period model $P_{L1} \leftarrow P_{1j}$;
\STATE \textbf{endmap}.collect();
\RETURN $RDD_{P_{L1}}$.
\end{algorithmic}
\end{algorithm}

In Algorithm \ref{alg42}, after loading the cached RDD object $RDD_{X_{K}}$ from the main memory system, each partition of $RDD_{lr}$ is shuffled into a series of partitions using a shuffle operation.
Each partition in the subsequent RDD object $RDD_{similarity}$ contains a pair of time subsequences for morphological similarity measurement.
A wide dependency occurs between partitions of RDD objects $RDD_{X_{K}}$ and $RDD_{similarity}$.
For each partition $P_{i}$ in $RDD_{similarity}$, let $\ell_{a} = \{k_{1}, k_{2}, ..., k_{m}\}$ be the first subsequence and $\ell_{b}=\{k_{m+1}, ~k_{m+2}, ~..., ~k_{2m}\}$ be the comparison subsequence, and then we measure the morphological similarity of $\ell_{a}$ and $\ell_{b}$.
Benefiting from the independence of different partitions of the subsequent RDD, each partition is calculated in parallel to generate new partitions of $RDD_{similarity}$.
Based on morphological similarity results in $RDD_{similarity}$, the overall variance of each periodic item is calculated in parallel using the Fourier spectrum analysis.
Then, an RDD object $RDD_{\vartheta}$ is created and the maximum value of $\vartheta$ is detected.
According to the maximum of $\vartheta$, the first-layer periodic model is recognized as $RDD_{P_{L1}}$.

\subsection{Parallelization of the PTSP Process}
\label{section4.4}
\begin{figure}[!ht]
 \setlength{\abovecaptionskip}{0pt}
 \setlength{\belowcaptionskip}{0pt}
\centering
\includegraphics[width=3.5in]{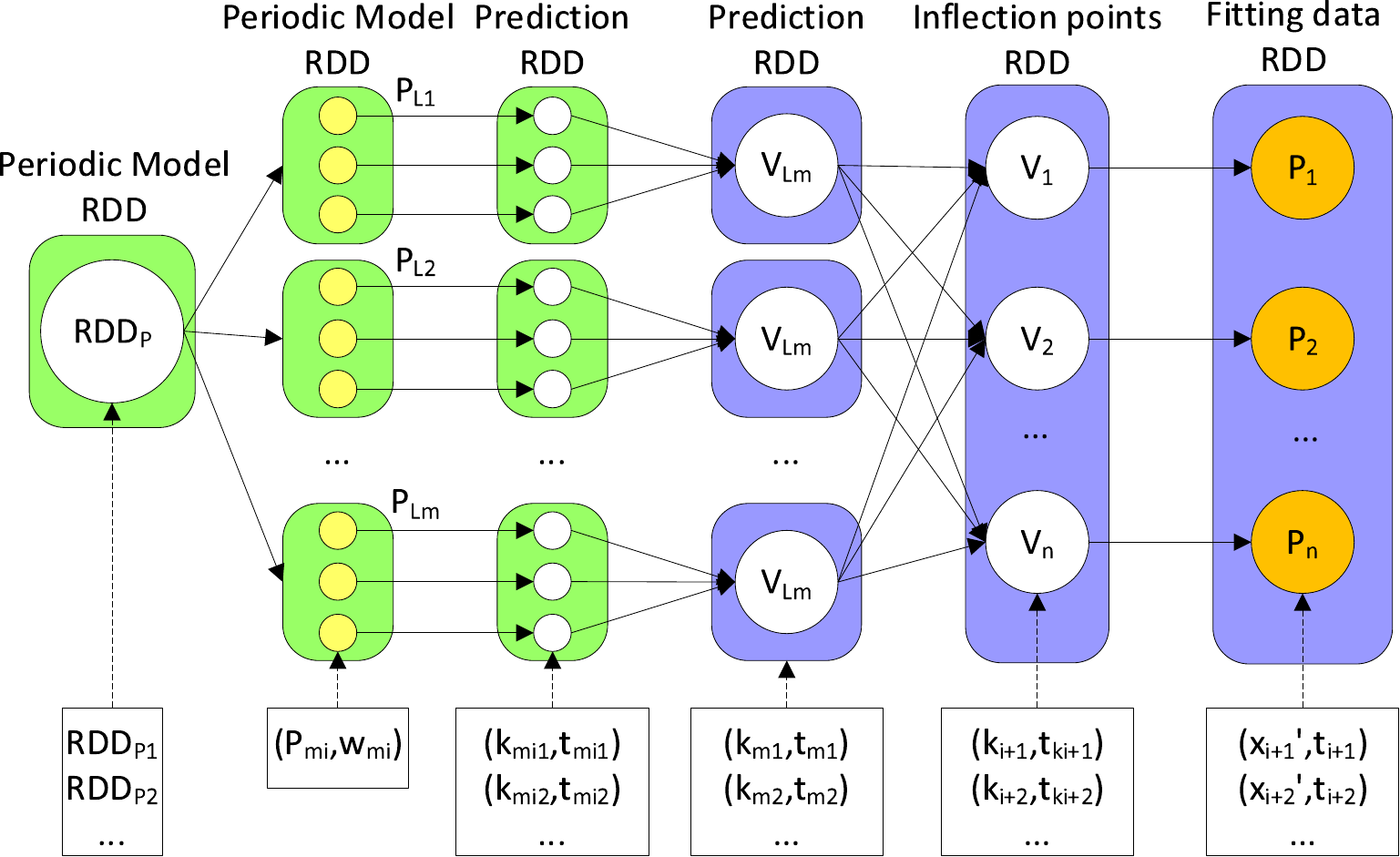}
\caption{RDD dependency graph of the parallel periodicity-based time series prediction.}
\label{fig16}
\end{figure}

Based on the detected periodic models, we perform a parallel solution for the periodicity-based time series prediction (P-PTSP) process on Spark platform.
In the P-PTSP process, the results of the next period are predicted based on the RDD object $RDD_{P}$ of the periodic model.
RDD object $RDD_{P_{Li}}$ of the periodic model in each layer is used in the prediction process at the same time.
Similar to the P-TSDCA and T-MTSPPR processes, the logical and data dependencies among different RDD objects in the $P-PTSP$ process are considered and computed.
We construct the RDD dependency graph of the P-PTSP process, as shown in Figure \ref{fig16}.

In Figure \ref{fig16}, the RDD object $RDD_{P}$ of the multi-layer periodic model is divided into separate partitions, where each partition $RDD_{P_{La}}$ represents a collection of periodic models in the $a$-th layer.
These partitions are independent of each other, so we can easily parallelize the subsequent processing.
According to the periodic models in $RDD_{P_{La}}$, the value of the next period is predicted and saved to the corresponding RDD object $RDD_{V_{La}}$.
Since each new partition in $RDD_{V_{La}}$ only depends on the corresponding periodic models $RDD_{P_{La}}$ of the $a$-th layer, there is a narrow dependency between partitions in RDD objects $RDD_{V_{La}}$ and $RDD_{P_{La}}$.
Then, we use a $reduce()$ function to combine all prediction values and generate an RDD object $RDD_{P_{1t+1}}$, where a set of predicted inflection points is created (e.g. $\{(k_{i+1}, t_{ki+1}), (k_{i+2}, t_{ki+2}), ...\}$).
Finally, based on the predicted inflection points, the fitting data values at all time stamps among these inflection points.
Note that the fitting data values between every two adjacent inflection points are calculated in parallel.
Algorithm \ref{alg43} gives the detailed steps of the P-PTSP process.

\begin{algorithm}[!ht]
\caption{Parallel implementation of the periodicity-based time series prediction (P-PTSP) process}
\label{alg43}
\begin{algorithmic}[1]
\REQUIRE ~\\
   $RDD_{P}$: the RDD object of the trained multi-layer periodic model $RDD_{P} = \{RDD_{P_{L1}}, RDD_{P_{L2}}, ~..., RDD_{P_{Lq}}\}$;\\
\ENSURE ~\\
    $RDD_{P_{1t+1}}$: the RDD object of the prediction of the next period.
\STATE $conf \leftarrow$  create SparkConf().setMaster(``$master$'').setAppName(``$P-PTSP$'');
\STATE $sc \leftarrow$ create SparkContext($conf$);	
\STATE $RDD_{P_{1t+1}} \leftarrow$ $RDD_{P}$.\textbf{foreach}
\STATE \quad each layer $RDD_{P_{La}}$ in $RDD_{P}$:
\STATE \quad $RDD_{p}' \leftarrow$ $RDD_{P_{La}}$.\textbf{map}
\STATE \qquad each period model $P_{ai}$ in $RDD_{P_{La}}$:
\STATE  \qquad calculate the time attenuation factor $\alpha_{ai} \leftarrow \frac{e^{\frac{|a|-i}{|a|}}}{\sum_{j=1}^{i}{e^{\frac{|a|-i}{|a|}}}}$;
\STATE  \qquad predict $P_{p}'  \leftarrow \sum_{a=1}^{q}{\sum_{i=1}^{|a|}{(\alpha_{ai} \times P_{ai})}}$, $X_{K(1t+1)}^{(ai)}  \leftarrow \alpha_{ai} \times X_{K(ai)}$;
\STATE \quad \textbf{endmap}.collect();
\STATE \quad predict $RDD_{P_{1t+1}}$.append($RDD_{p}'$);
\STATE \textbf{endfor}.reduce();
\STATE $P_{1t+1} \leftarrow$ $P_{1t+1}$.parallize().\textbf{map}
\STATE \quad each inflection point $k_{i}$ in $P_{1t+1}$:
\STATE \quad $sc$.parallelize($j$ in $(t_{ki}, t_{ki+1})$).\textbf{flatMap}
\STATE  \qquad calculate the fitting data $x_{j}' \leftarrow \frac{k_{i+1} - k_{i}}{t_{ki+1} - t_{ki}} \times (t_{j} - t_{ki}) + k_{i}$;
\STATE \qquad append $x_{j}'$ to $P_{1t+1}$;
\STATE \quad \textbf{endmap}.groupBykey().reduce();
\STATE \textbf{endmap}.collect();
\RETURN $RDD_{P_{1t+1}}$.
\end{algorithmic}
\end{algorithm}

\begin{figure}[!ht]
 \setlength{\abovecaptionskip}{0pt}
 \setlength{\belowcaptionskip}{0pt}
\centering
\includegraphics[width=4.5in]{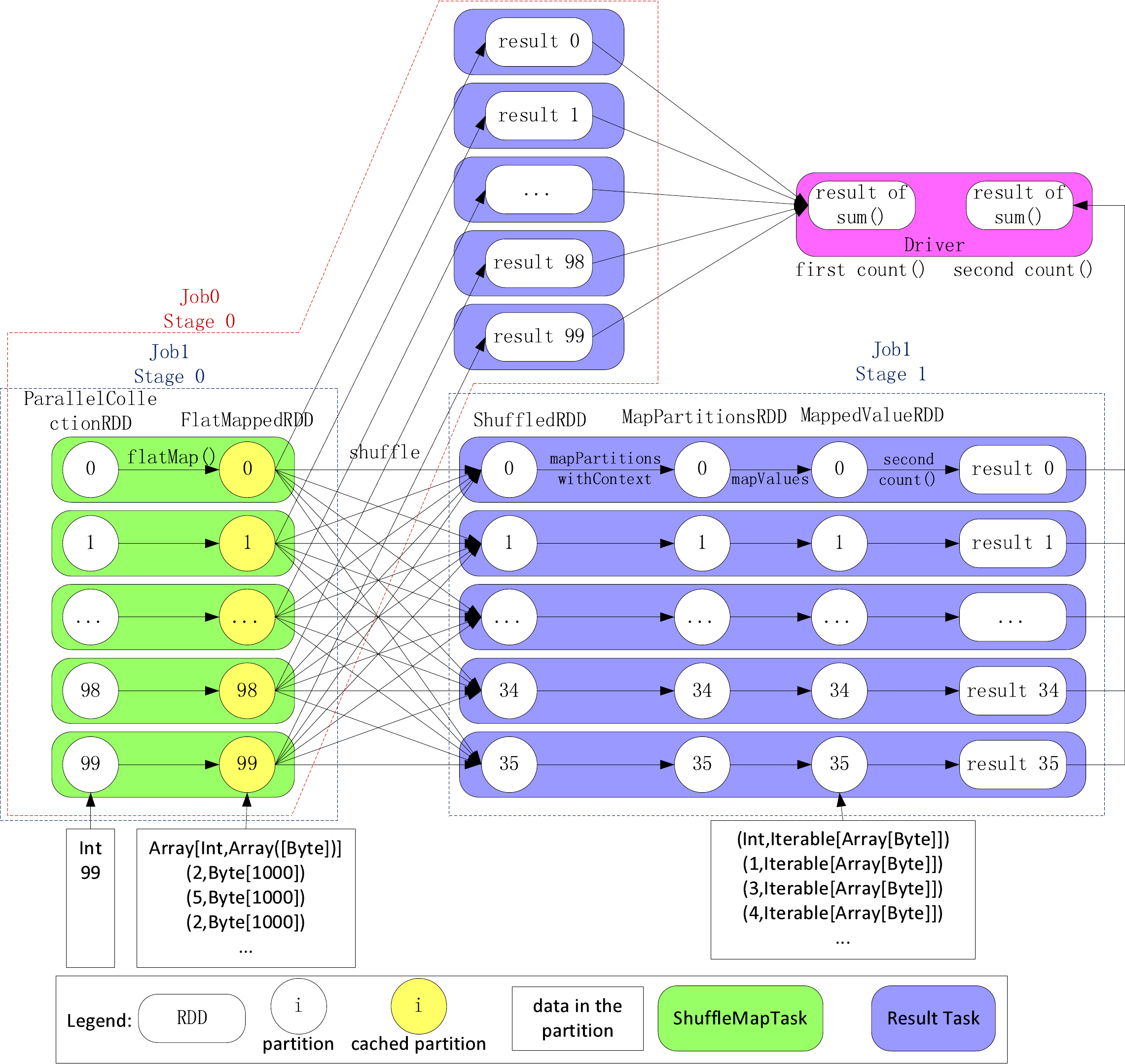}
\caption{Parallel job physical graph of the P-PTSP process.}
\label{fig17}
\end{figure}

According to the RDD dependency graphs of P-TSDCA, P-MTSPPR, and P-PTSP processes, we construct corresponding task DAGs and submit the tasks in these DAGs to the Apache Spark task scheduler.
In each task DAG, one or more jobs are generated based on the program's logical dependencies.
In addition, in each job, for each wide dependency between RDD objects, a separate job stage is created to generate the corresponding tasks.
In Spark, the task scheduler listener module monitors the submitted jobs and splits each job into different stages and tasks.
The task scheduler module then receives the tasks and allocates them to the appropriate executors for parallel execution.
Taking the P-PTSP process as an example, the Parallel job physical graph of P-PTSP is shown in Figure \ref{fig17}.

\section{Experiments}
\label{section5}
In this section, experiments are conducted to evaluate the prediction accuracy and performance of the proposed PPTSP algorithm.
Section \ref{section5.1} describes the experimental setup.
Section \ref{section5.2} evaluates the effectiveness of the big data compression and abstraction, as well as the accuracy of periodic pattern recognition and time series prediction.
Section \ref{section5.3} presents the performance comparison of the proposed algorithm in terms of execution time, speedup, data communication, and workload balancing.

\subsection{Experiment Setup}
\label{section5.1}
Our experiments are performed on a workstation at the National Supercomputing Center in Changsha, China \cite{url00}.
The workstation is equipped with 25 high-performance computing nodes, and each of them is equipped with an Intel Xeon Nehalem EX CPU and 48 GB main memory, respectively.
Each Nehalem-EX processor features up to 8 cores inside a single chip supporting 16 threads and 24 MB of cache.
All computing nodes are connected by a high-speed Gigabit network.
The operating system running on each computing node is Ubuntu 12.04.4.
An Apache Spark cloud computing platform is configured on the workstation with the version of 2.1.2, while equipped with Spark Streaming, SQL and DataFrames, and MLlib (machine learning) libraries.
The proposed PPTSP algorithm is implemented in Python 3.5.4.

Four groups of actual time-series datasets in different fields are used in our experiments, including meteorology, finance, ocean engineering, and transportation.
Specifically, in the financial sector, stock price datasets from more than 150 companies are collected.
In the field of ocean engineering, more than 950,612 sea-surface temperature datasets are collected from 101 sites in the Pacific, Atlantic, and Indian Oceans.
In the area of transportation, more than 15,255,905 traffic flow records are collected from 5971 sensors (detectors) in Minnesota, USA.
The detailed description of the meteorological datasets is provided in Table \ref{table51}.

\begin{table}[!ht]
\centering
\renewcommand{\arraystretch}{1.3}
\setlength{\abovecaptionskip}{0pt}
\setlength{\belowcaptionskip}{0pt}
\caption{Time-series datasets used in the experiments}
\label{table51}
\tabcolsep1pt
\begin{tabular}{C{1.0cm} L{6.5cm} C{3.0cm} C{2.0cm}}
\hline
\textbf{No.} & \textbf{Dataset name} & \textbf{Year/Month}  & \textbf{Samples} \\
\hline
1   & Meteorological (MET) dataset \cite{dataurl01}          & 2001/01 - 2018/04 & 6,091,680 \\
2   & Sea-Surface temperature (SST) dataset \cite{dataurl03} & 1978/02 - 2018/04 & 1,250,612 \\
3   & Traffic flow (TF) dataset \cite{dataurl04}  & 2011/01 - 2018/04 & 15,255,905 \\
4   & Stock price (SP) dataset \cite{dataurl02}             & 1991/01 - 2018/04 & 1,032,750 \\
\hline
\end{tabular}
\end{table}

The MET dataset is downloaded from the China Meteorological Data Service Center (CMDC) \cite{dataurl01}.
We collect the historical meteorological datasets for four cities/provinces in China, including Beijing, Shanghai, Hunan, and Fujian.
There are various meteorological factors in each sample, such as temperature, air pressure, wind, humidity, cloud, and precipitation.
In our experiments, the temperature feature is considered.

The SST dataset is collected from 101 sites of Global Tropical Moored Buoy Array (GTMBA) in the Pacific, Atlantic, and Indian Oceans \cite{dataurl03}.
The sea surface temperature measured at 1 meter below the sea surface.
Specifically, these SST datasets are collected from 71 sites in the Pacific Ocean, 21 sites in the Atlantic Ocean, and 29 sites in the Indian Ocean.

The TF dataset is continuously collected by the Regional Transportation Management Center (RTMC) \cite{dataurl04}, at a 30-second interval from over 5,971 loop sensors (detectors) located around the Twin Cities Metro freeways, seven days a week and all year round.
The raw traffic flow records are stored in the format of Binary Unified Transportation Sensor Data Format (UTSDF).
There are 5760 records generated by each sensor in one day.
Before the experiments, we converted these binary datasets into numeric time-series datasets.

The SP dataset is collected from the stock channel of Sohu.com Inc. \cite{dataurl02}.
We collect the stock price data for more than 150 companies from 1991/01 to 2018/04.
There are 240 - 255 Stock trading records (samples) every year, and each sample includes opening price, high price, low price, and closing price.

\subsection{Effectiveness Evaluation of Data Compression and Representation}
In this section, we conduct experiments on actual time-series datasets to compare the proposed TSDCA algorithm and the related time-series data compression and representation algorithms.

\label{section5.2}
\subsubsection{Evaluation Metric}
We introduce the common metrics of Data Compression Ratio ($R_{DC}$) and Data Extraction Accuracy ($Acc_{DE}$) in the experiments to evaluate the effectiveness of the comparison algorithms.
The metrics of $R_{DC}$ and $Acc_{DE}$ are also used in the related data compression algorithms in \cite{ea03, ea09}.
$R_{DC}$ refers to the ratio of the data size of the abstracted dataset $X_{K}$ to that of the raw dataset $X_{T}$, as defined in Equation (\ref{eq25}):

\begin{equation}
\label{eq25}
\begin{aligned}
R_{DC} = \frac{|X_{K}|}{|X_{T}|},
\end{aligned}
\end{equation}
where $|X_{T}|$ is the number of data samples in $X_{T}$ and $|X_{K}|$ is the number of data samples in $X_{K}$.
Given a set of extracted inflection points $X_{K} = \{(k_{1}, t_{1}), ~...,~(k_{m}, t_{k_{m}})\}$, we can use Equation (\ref{eq23}) to calculate the fitted value $x_{j}'$ at each time stamp between every two adjacent inflection points $k_{i}$ and $k_{i+1}$ .
The value of $Acc_{DE}$ is measured by comparing the difference between the $M$ fitting values $x_{j}'$ and the corresponding real values $x_{j}$, as defined in Equation (\ref{eq26}):

\begin{equation}
\label{eq26}
Acc_{DE} = \frac{1}{\sum_{j=1}^{M}{\frac{|x_{j}-x_{j}'|}{\max{(x_{j},~x_{j}')}}}} = \sum_{j=1}^{M}{\frac{\max{(x_{j},~x_{j}')}}{|x_{j}-x_{j}'|}}.
\end{equation}

\subsubsection{Effectiveness Evaluation of Comparison Algorithms}
\begin{figure}[!ht]
 \setlength{\abovecaptionskip}{0pt}
 \setlength{\belowcaptionskip}{0pt}
 \centering
 \subfigure[Sea-Surface temperature (SST) dataset]{
 \label{chart01:a}
 \includegraphics[width=3.1in]{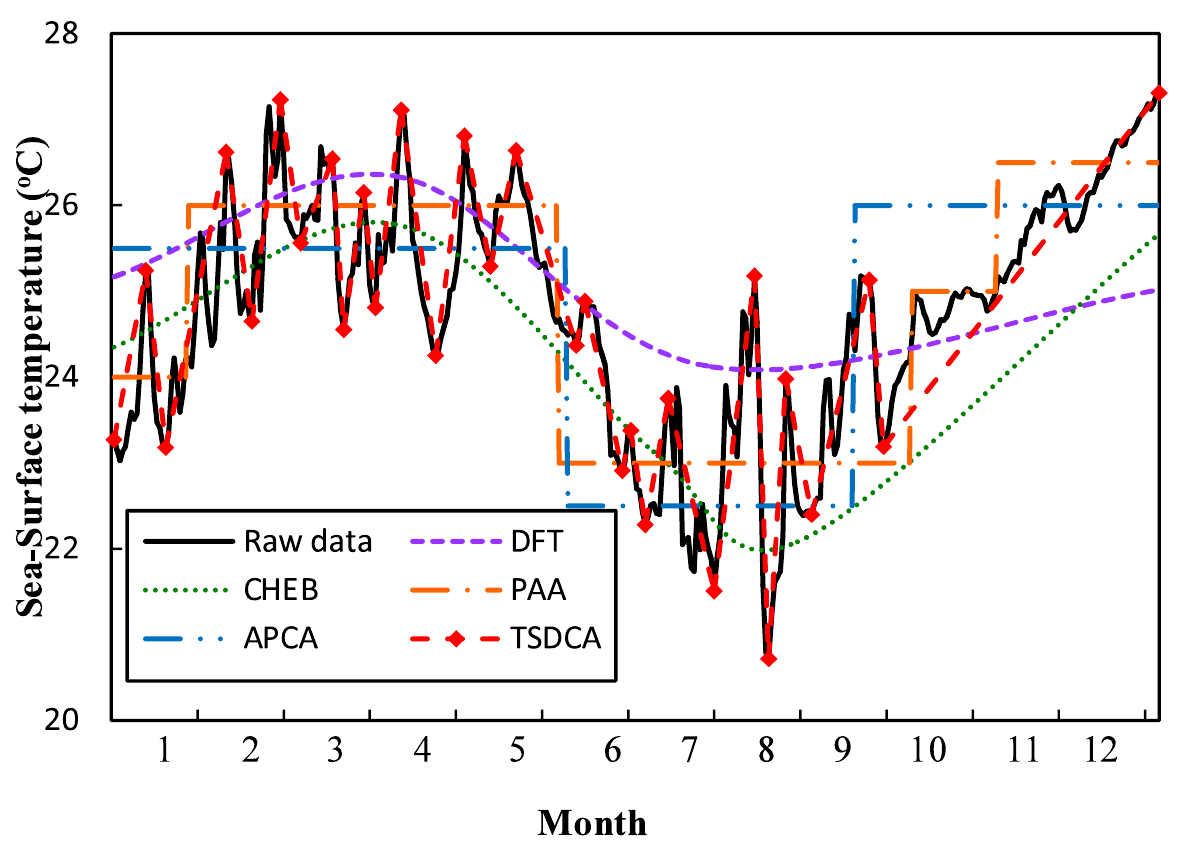}}
 \subfigure[Stock price (SP) dataset]{
 \label{chart01:b}
 \includegraphics[width=3.1in]{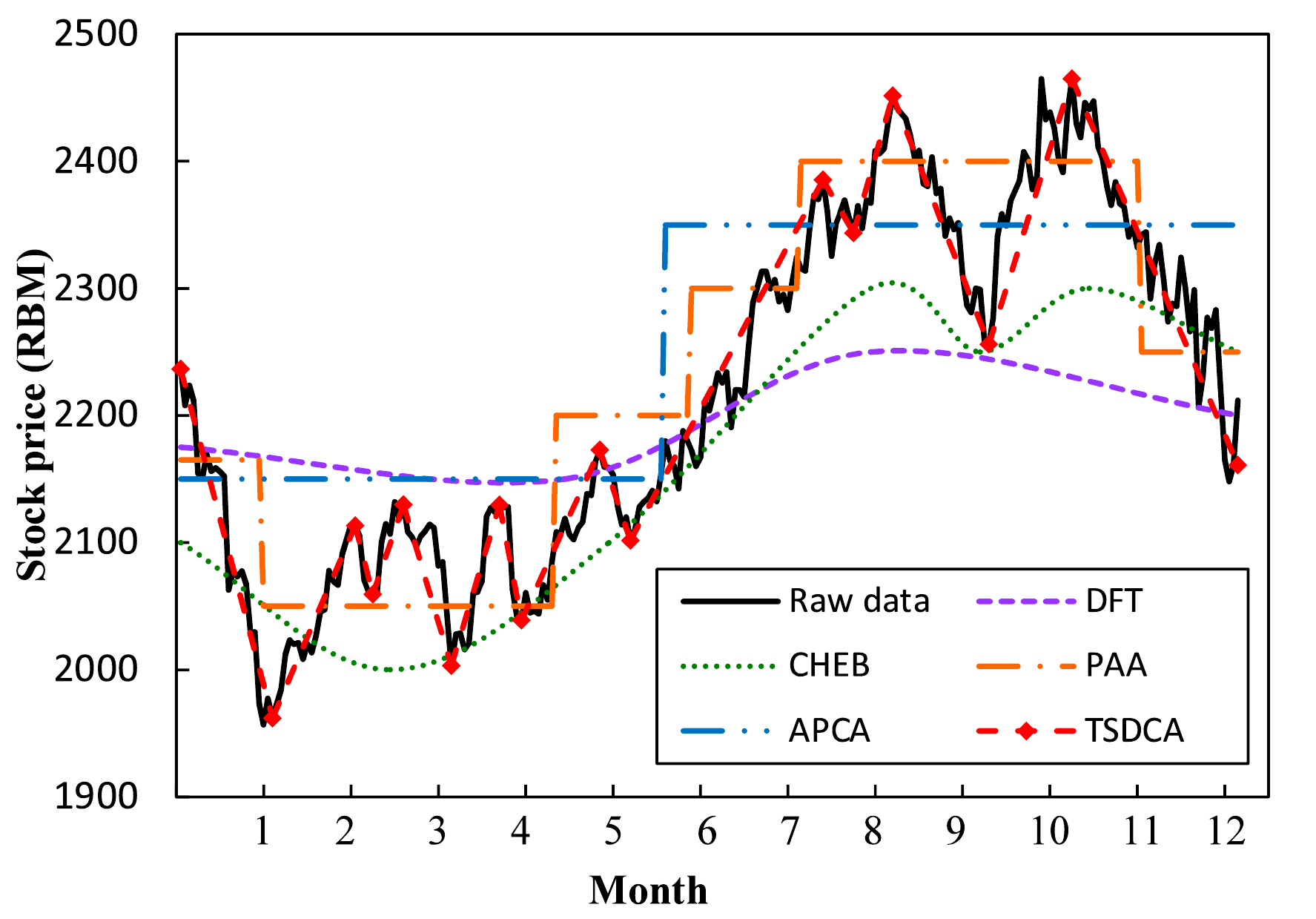}}
 \subfigure[Travel flow (TF) dataset]{
 \label{chart01:c}
 \includegraphics[width=3.1in]{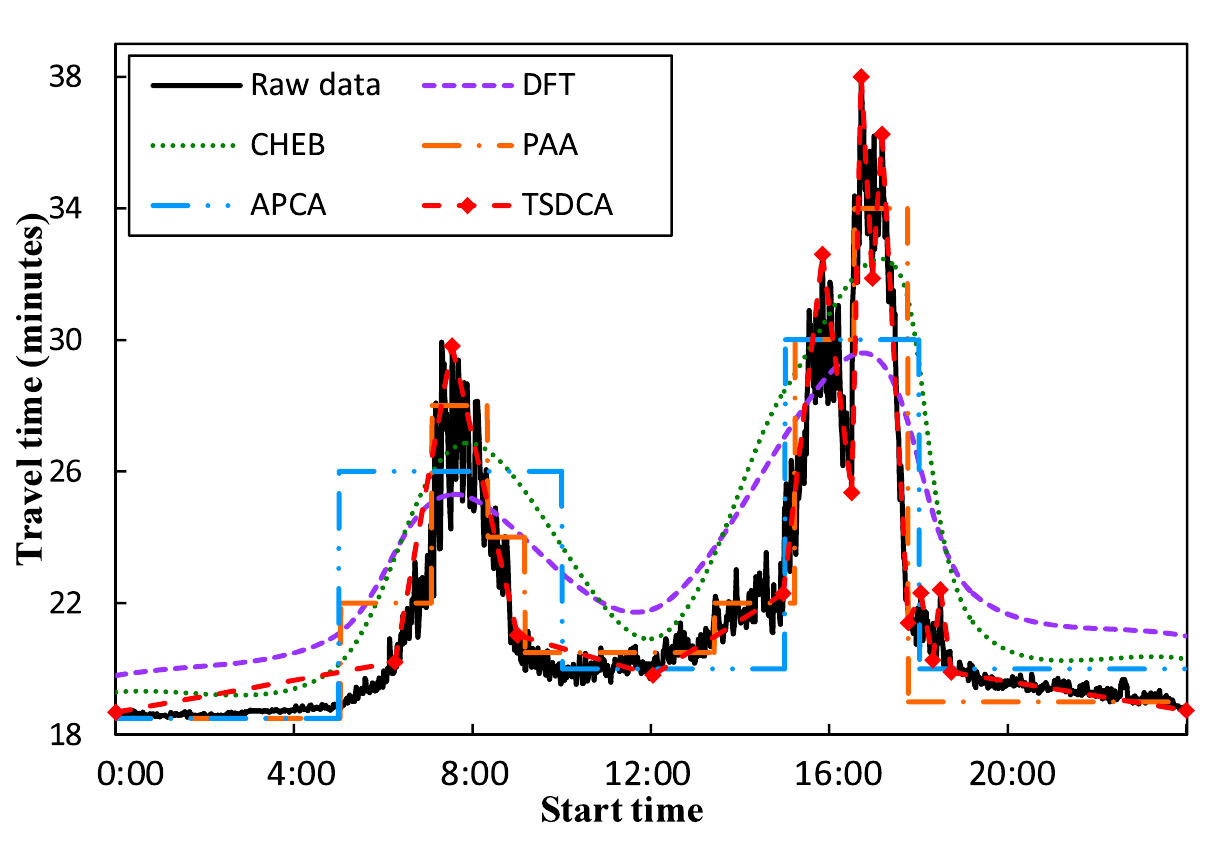}}
 \subfigure[Meteorological (MET) dataset]{
 \label{chart01:d}
 \includegraphics[width=3.1in]{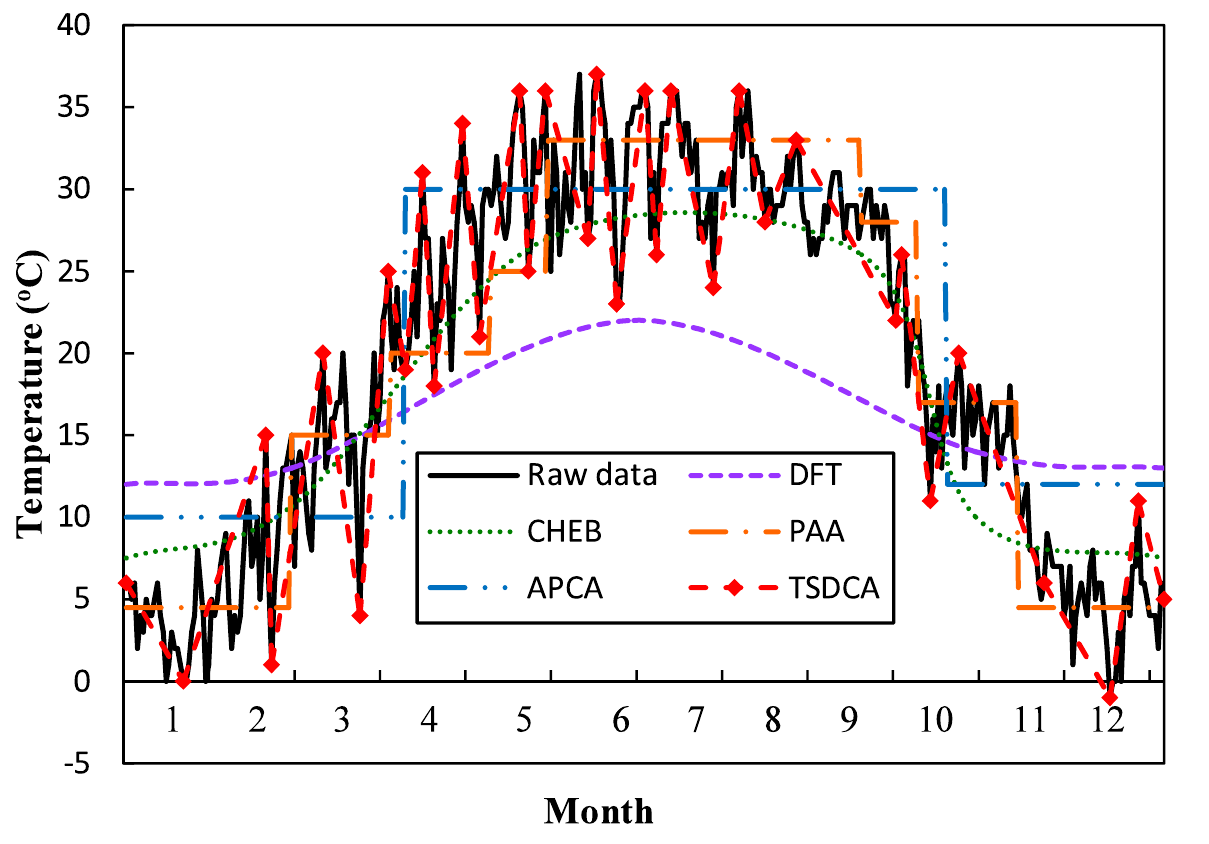}}
 \caption{Effectiveness comparison of data compression and representation algorithms.}
\label{chart01}
\end{figure}

To evaluate the effectiveness of the proposed TSDCA algorithm, experiments are performed using the TSDCA algorithm and the DFT \cite{ea04}, PAA \cite{ea03}, APCA \cite{ea09}, and CHEB \cite{ea02} algorithms on the four groups of actual time-series datasets described in Table \ref{table51}.
In each case, to the TSDCA algorithm, the parameter thresholds of inflection rate $\delta$ and time length $\varepsilon$ are both set to $0.85$.
And the parameters of the comparison algorithms take the recommended values in their paper.
Examples of the experimental results of time-series data compression are shown in Figure \ref{chart01}.

\begin{figure}[!ht]
 \setlength{\abovecaptionskip}{0pt}
 \setlength{\belowcaptionskip}{0pt}
 \centering
 \subfigure[Data compression ratio]{
 \label{chart02:a}
 \includegraphics[width=3in]{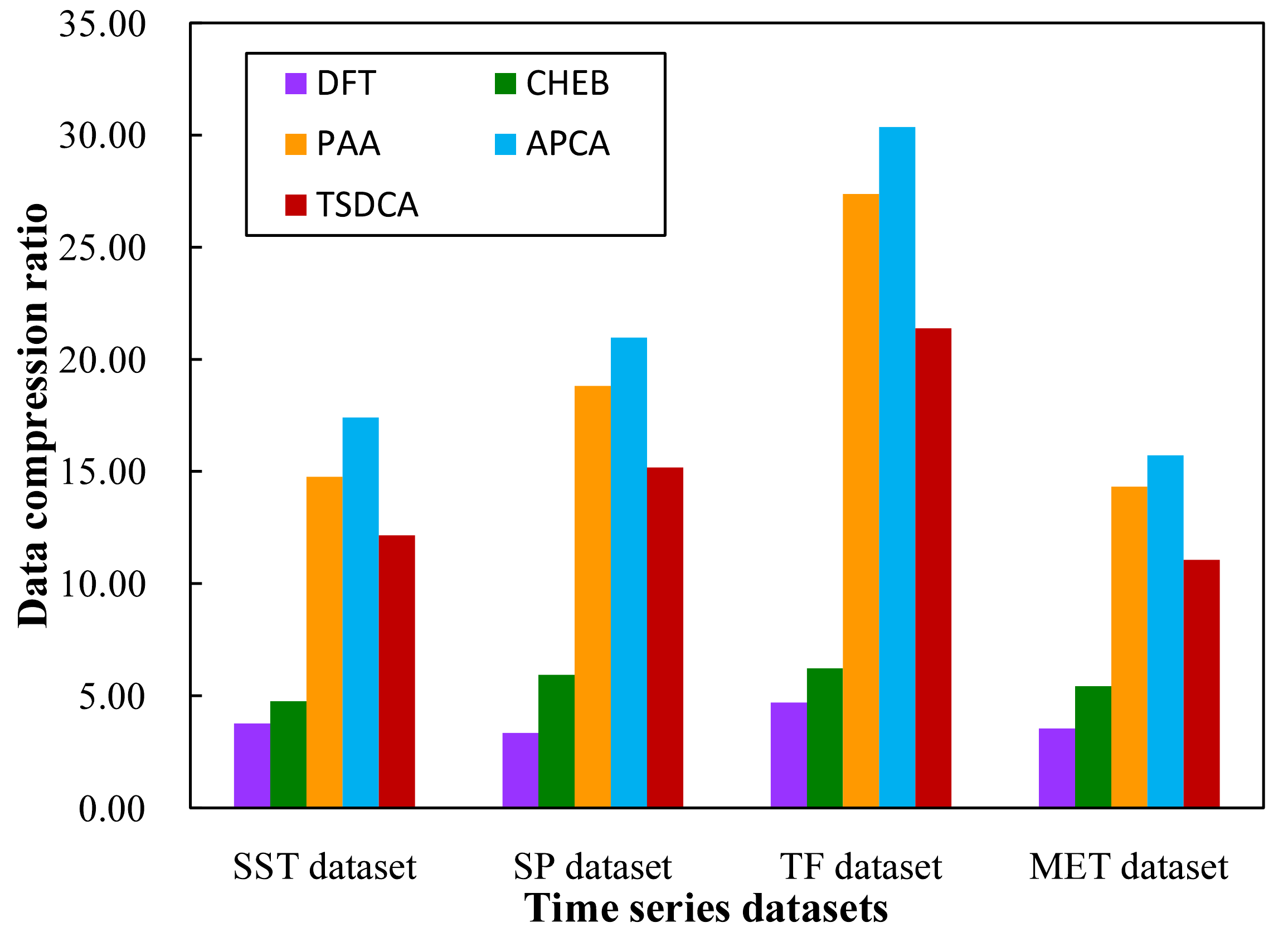}}
 \subfigure[Data extraction accuracy]{
 \label{chart02:b}
 \includegraphics[width=3in]{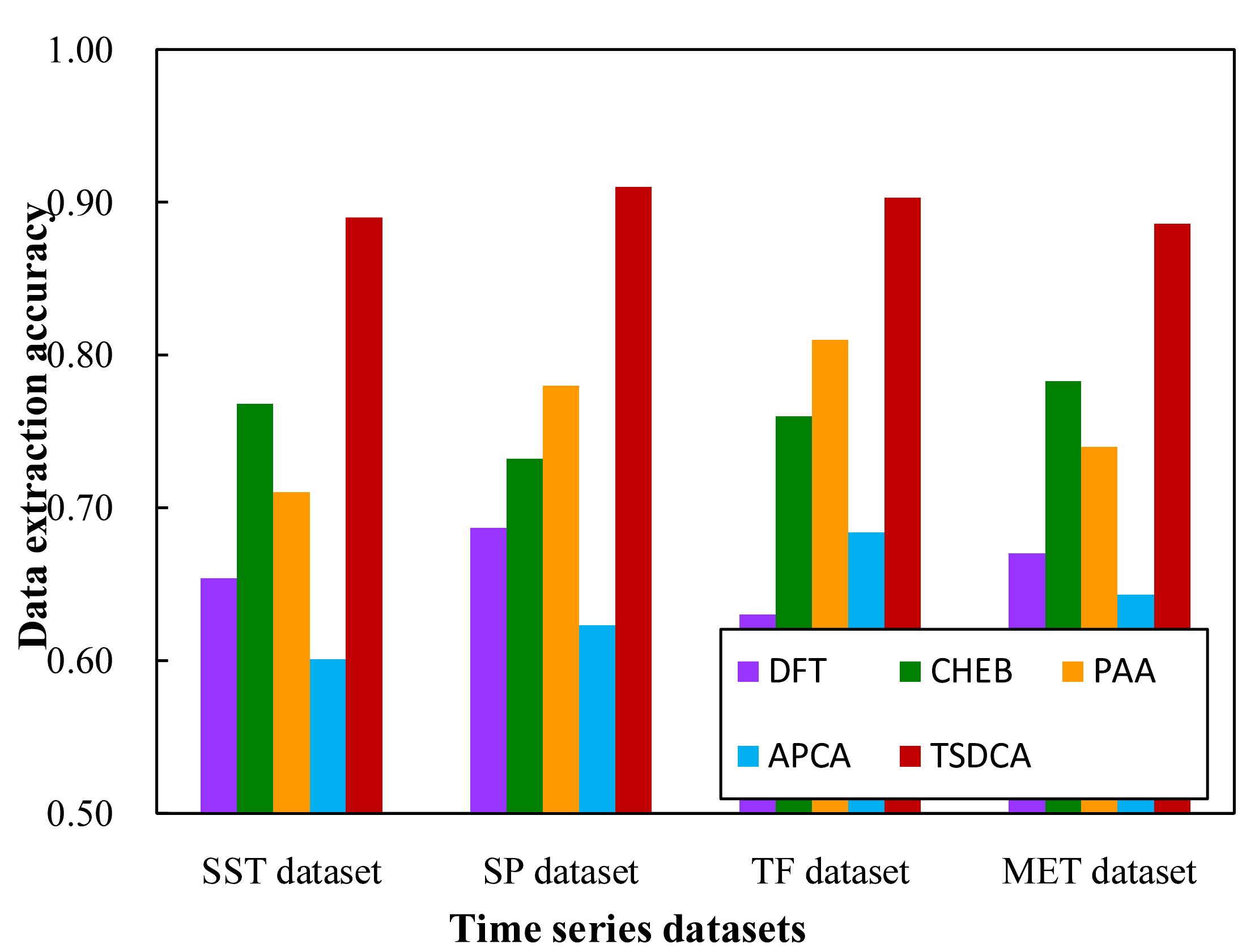}}
 \caption{Data compression ratio and data extraction accuracy of the comparison algorithms.}
\label{chart02}
\end{figure}

Figures \ref{chart01} (a) -(b) show the time series of different datasets over a small period and the corresponding data compression and representations of the comparison algorithms.
Figure \ref{chart01} (a) shows the SST time series of an observation site (located at 0N 95W) in the Pacific Ocean in 1982 (365 days).
Figure \ref{chart01} (b) shows the time series of the opening stock price of a company in 2012 (243 working days).
Figure \ref{chart01} (c) shows the predicted travel time of a freeway (U.S.169, NB) from the S1142 station to S757 station on a weekday.
Figure \ref{chart01} (d) shows the temperature in Beijing in 2017 (365 days).
From the four cases in Figure \ref{chart01}, it can be observed that the proposed TSDCA achieves high data extraction accuracy and effective compression for the original dataset.
Compared with the approximate algorithms of CHEB, PAA, and APCA, the proposed TSDCA algorithm can accurately extract the skyline of the time series.
In addition, in comparison of the dimensionality reduction algorithm of DTF, TSDCA can extract the critical characteristics in each dimension to form a data abstraction.
To clearly compare the compression effectiveness of the comparison algorithms, the comparison results are shown in Figure \ref{chart02} in terms of data compression ratio and data extraction accuracy.

As shown in Figure \ref{chart02} (a) and (b), TSDCA obtains the highest $Acc_{DE}$ (in the range of 0.88 - 0.91) in all cases with competitive $R_{DC}$ compression ratios.
For example, in the SST case, the average $R_{DC}$ of TSDCA is 12.16, which is lower than that of APCA (17.41) and PAA (14.76), but much higher than that of DFT (3.77) and CHEB (4.75).
In the same case, TSDCA achieves the highest $Acc_{DE}$ value at 0.89, while that of CHEB, PAA, DFT, and APCA is 0.77, 0.71, 0.65, and 0.60, successively.
In the TF case, benefiting from the smooth feature values and few data inflection points, the average $R_{DC}$ of TSDCA is as high as 21.38 and its $Acc_{DE}$ is equal to 0.90.
Although the APCA algorithm achieves the highest compression ratios in each case, it has the lowest data extraction accuracy.
In contrast, the DFT algorithm is significantly inferior to the comparison algorithms in both $Acc_{DE}$ and $R_{DC}$.
Therefore, the average data extraction accuracy of the proposed TSDCA algorithm is significantly higher than that of SCHEB, PAA, APCA, and DTF, while achieving higher $R_{DC}$ values than CHEB and DFT.

\subsubsection{Impact of Thresholds of Pseudo Inflection Point Parameters}
\begin{figure}[!ht]
 \setlength{\abovecaptionskip}{0pt}
 \setlength{\belowcaptionskip}{0pt}
 \centering
 \subfigure[Data compression ratio]{
 \label{chart03:a}
 \includegraphics[width=3.0in]{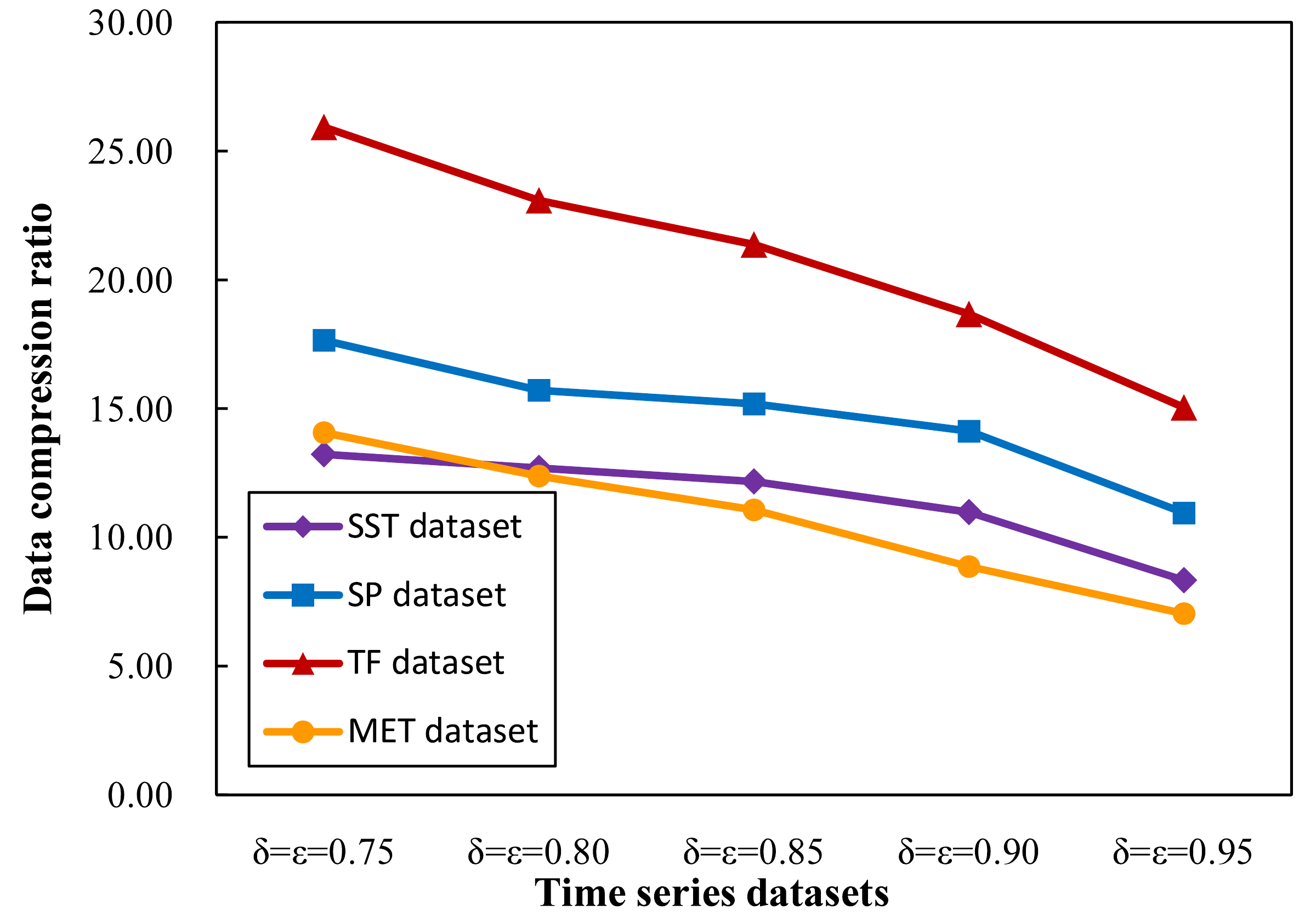}}
 \subfigure[Data extraction accuracy]{
 \label{chart03:b}
 \includegraphics[width=3.0in]{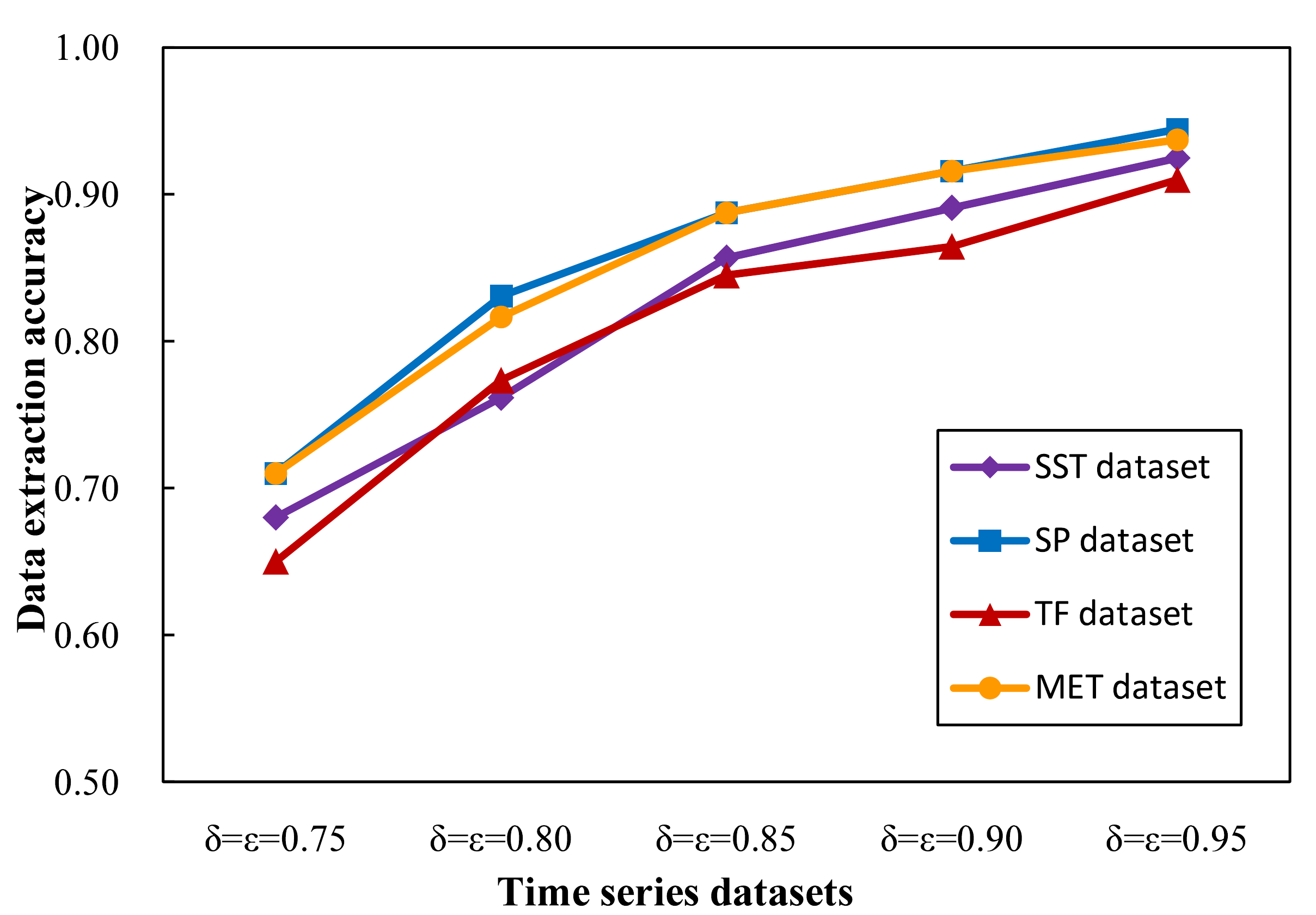}}
 \caption{Impact of pseudo inflection point parameter thresholds on data compression.}
\label{chart03}
\end{figure}

We conduct experiments to evaluate the impact of the thresholds of pseudo inflection point parameters on the results of time-series data compression.
Four groups of time-series datasets in Table \ref{table51} are used for TDSCA with different pseudo inflection thresholds.
From observing the different $R_{DC}$ values at different thresholds of parameters $\delta$ and $\varepsilon$ in each case, we can evaluate the impact of the two parameters on the data compression ratio and extraction accuracy.
In each case, thresholds of inflection rate $\delta$ and time length $\varepsilon$ are both set to 0.75, 0.80, 0.85, 0.90, and 0.95, gradually.
The experimental results of the impact of inflection rate ¦Ä and time length ¦Å on the time-series data compression are shown in Figure \ref{chart03}.

As shown in Figure \ref{chart03} (a) and (b), as the thresholds of $\delta$ and $\varepsilon$ increase, TSDCA obtains a lower data compression ratio in each case and achieves higher data extraction accuracy at the same time.
In the SST case, when $\delta$ = $\varepsilon$ =0.75, the $R_{DC}$ of TSDCA is equal to 13.22.
When the values of $\delta$ and $\varepsilon$ increase, this means that there are more strict measurements when detecting pseudo inflection points and fewer data can be removed, resulting in a lower data compression ratio.
Hence, when $\delta$ = $\varepsilon$ =0.95, we can see that in the case of SST, the $R_{DC}$ decreases to 8.33.
As the thresholds of $\delta$ and $\varepsilon$ increase from 0.75 to 0.95, the $R_{DC}$ for the SP dataset decreases from 17.65 to 10.94, from 25.93 to 15.04 in the case of TF, and from 14.06 to 7.03 in the case of MET.

However, the increase of the thresholds of $\delta$ and $\varepsilon$ facilitates the accuracy of data extraction.
In contrast to $R_{DC}$, the $Acc_{DE}$ for each dataset under the high thresholds of parameters $\delta$ and $\varepsilon$ is higher than that under the low thresholds.
As the thresholds of $\delta$ and $\varepsilon$ increase from 0.75 to 0.95, the $Acc_{DE}$ increases from 0.68 to 0.92 in the case of SST, and from 0.71 to 0.93 in the case of MET.
Therefore, we can conclude that the TSDCA algorithm achieves high data extraction accuracy and maintains high data compression ratios for time-series datasets.
In addition, the $R_{DC}$ for each dataset under low thresholds of parameters $\delta$ and $\varepsilon$ is higher than that under high thresholds.
The higher the thresholds of parameters $\delta$ and $\varepsilon$, the lower the data compression ratios and the higher the data extraction accuracy.
Users can achieve a compromise between these two issues by setting appropriate parameter thresholds.

\subsection{Accuracy Evaluation of Periodicity Detection Algorithms}
In this section, we conduct comparison experiments to evaluate the accuracy of the proposed MTSPPR algorithm and the related periodic pattern detection algorithms, including the CONV \cite{eb01}, WARP \cite{eb02}, FPPM \cite{eb03}, and STNR \cite{eb06} algorithms.
In addition, we also present examples of the results of periodic pattern detection and prediction on the SST, SP, TF, and MET time series datasets.

\subsubsection{Evaluation Metric}
We introduce the metrics of confidence to evaluate the comparative periodicity detection algorithms, which is widely used in \cite{eb01, eb02, eb05, eb03}.
Given a time series dataset $X_{T}$ and a candidate periodic pattern $P$, the confidence of $P$ is the ratio of its actual periodicity $P_{A}$ to its expected perfect periodicity $P_{E}$, as defined in Equation (\ref{eq27}):

\begin{equation}
\label{eq27}
conf(P,st,|P|) = \frac{P_{A}(P,st,|P|)}{P_{E}(P,st,|P|)},
\end{equation}
where $|P|$ is the length of the candidate periodic, $st$ is the starting position of $P$, and $conf(P,st,|P|) \in (0, 1)$ and
\begin{equation}
\label{eq28}
P_{E}(P,st,|P|) =\left\lfloor \frac{|X_{T}|-st + 1}{|P|} \right\rfloor,
\end{equation}
and $P_{A}(P,st,|P|)$ is computed by counting the number of occurrences of $P$ in $X_{T}$,  starting at $st$ and repeatedly jumping by $|P|$ positions.

\subsubsection{Confidence Evaluation of Comparison Algorithms}
\begin{figure}[!ht]
 \setlength{\abovecaptionskip}{0pt}
 \setlength{\belowcaptionskip}{0pt}
 \centering
 \includegraphics[width=3in]{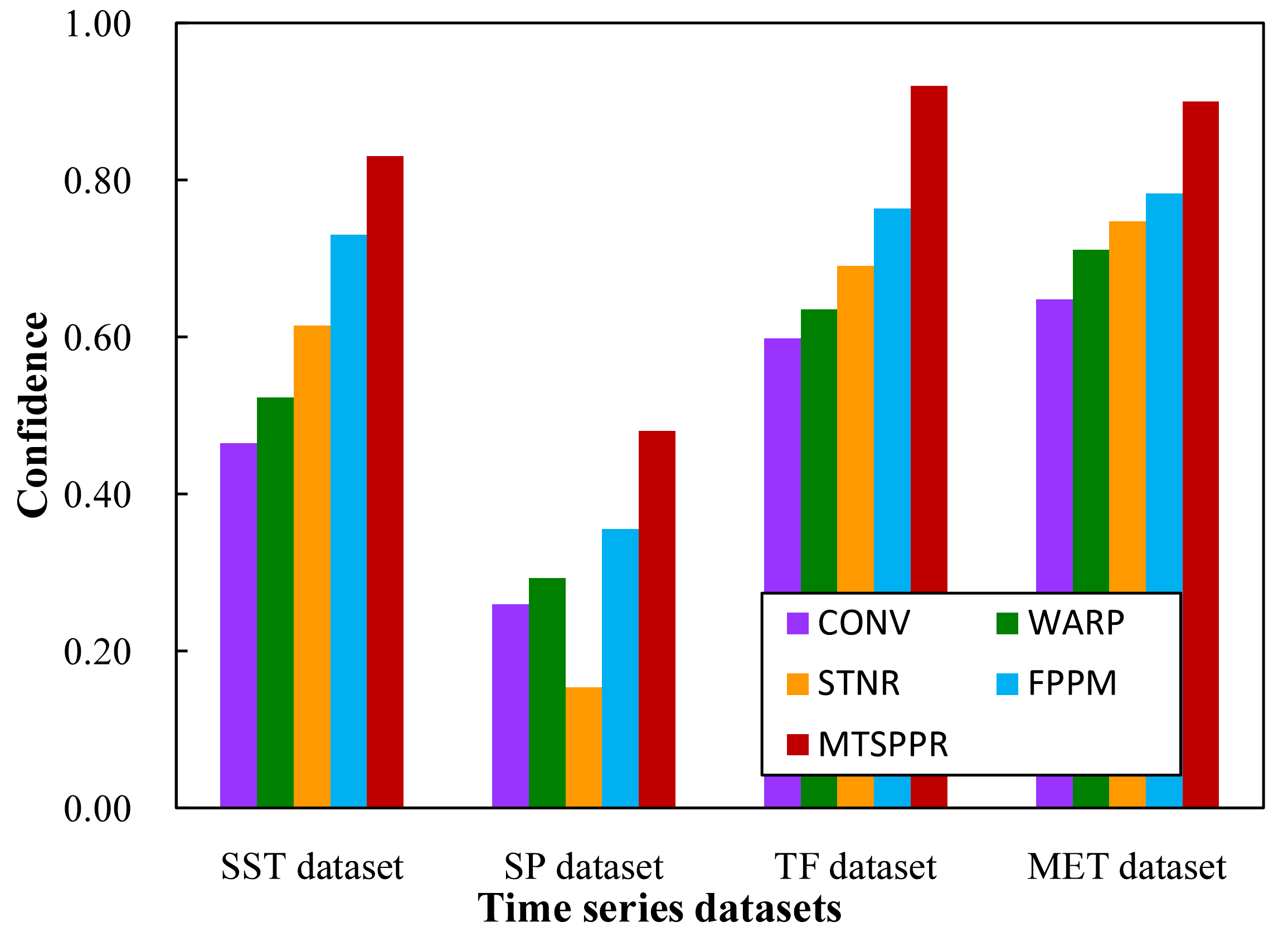}
 \caption{Confidence of the periodic pattern detection algorithms.}
\label{chart04}
\end{figure}

We evaluate the confidence of the proposed MTSPPR algorithm by comparing it with the CONV, WARP, FPPM, and STNR algorithms.
For each comparison algorithm, four groups of actual time-series datasets in Table \ref{table51} are tested.
The parameters of the comparison algorithms take the recommended values in their paper.
For the MTSPPR algorithm, the parameter thresholds of the growth set $\mu$ and the scaling ratio $\varphi$ are set as the optimized values described in Section 5.3.3.
Experimental results of the confidence evaluation of the comparative periodicity detection algorithms are shown in Figure \ref{chart04}.

As shown in Figure \ref{chart04}, the proposed MTSPPR algorithm obtains significantly higher confidence than the comparison algorithms in all cases.
For example, benefiting from the obvious periodicity of the TF dataset, MTSPPR obtains the highest confidence of 0.92 for TF, while achieving the value of 0.90 for MET and 0.83 for SST.
In contrast, although the CONV algorithm works well on synthetic datasets, it obtains the least effect on the actual time-series datasets in our experiments.
In the MET case, CONV yields a confidence of 0.65, and 0.60 for TF and 0.46 for SST.
Thanks to the noise resilient and the suffix tree structure, the STNR algorithm is superior to the CONV, WAPR and FPPM algorithms in the cases of SST, TF, and MET.
However, STNR is less efficient in variable-length flexible patterns and reaches the lowest confidence in the SP case.
Although all comparison algorithms work well on the datasets having obvious periodicity, i.e., the SST, TF, and MET datasets, these algorithms consistently yield poor confidence in the SP case because it is complex and almost without periodicity.
The proposed MTSPPR algorithm achieves a confidence value of 0.48 on SP, while FPPM obtains 0.36, WARP gets 0.29, CONV gets 0.26, and STNR has a minimum value of 0.15.
Therefore, the experimental results denote that the proposed MTSPPR algorithm achieves higher accuracy than the comparison algorithms on the periodic detection of time-series datasets.

\subsubsection{Impact of Thresholds of Parameters $\mu$ and $\varphi$}
In the MTSPPR algorithm, we propose a morphological similarity measurement method to calculate the distance between every two comparison subsequences, where two parameters $\mu$ and $\varphi$ are used to control the increase of the comparison subsequences.
$\mu$ is the growth step of the comparison subsequences, namely, there are $\mu$ inflection points increasingly  incorporated into the comparison subsequences each time.
In addition, the number of inflection points in the comparison subsequences that might exist periodic patterns may be slightly different due to the inflection points marking and the pseudo inflection points deletion operations.
Therefore, we introduce a scaling ratio factor $\varphi$ ($\varphi \in [0,1)$) to control the number of inflection points of the latter comparison subsequence $\ell_{b}$.
In this way, the comparison subsequences are optimized from fixed-length rigid sequences to variable-length flexible sequences.
The length of $\ell_{b}$ is within the range of the left and right extension of the length of the previous comparison subsequence $\ell_{a}$.

\begin{figure}[!ht]
 \setlength{\abovecaptionskip}{0pt}
 \setlength{\belowcaptionskip}{0pt}
 \centering
 \subfigure[Impact of the growth step $\mu$]{
 \label{chart05:a}
 \includegraphics[width=3.0in]{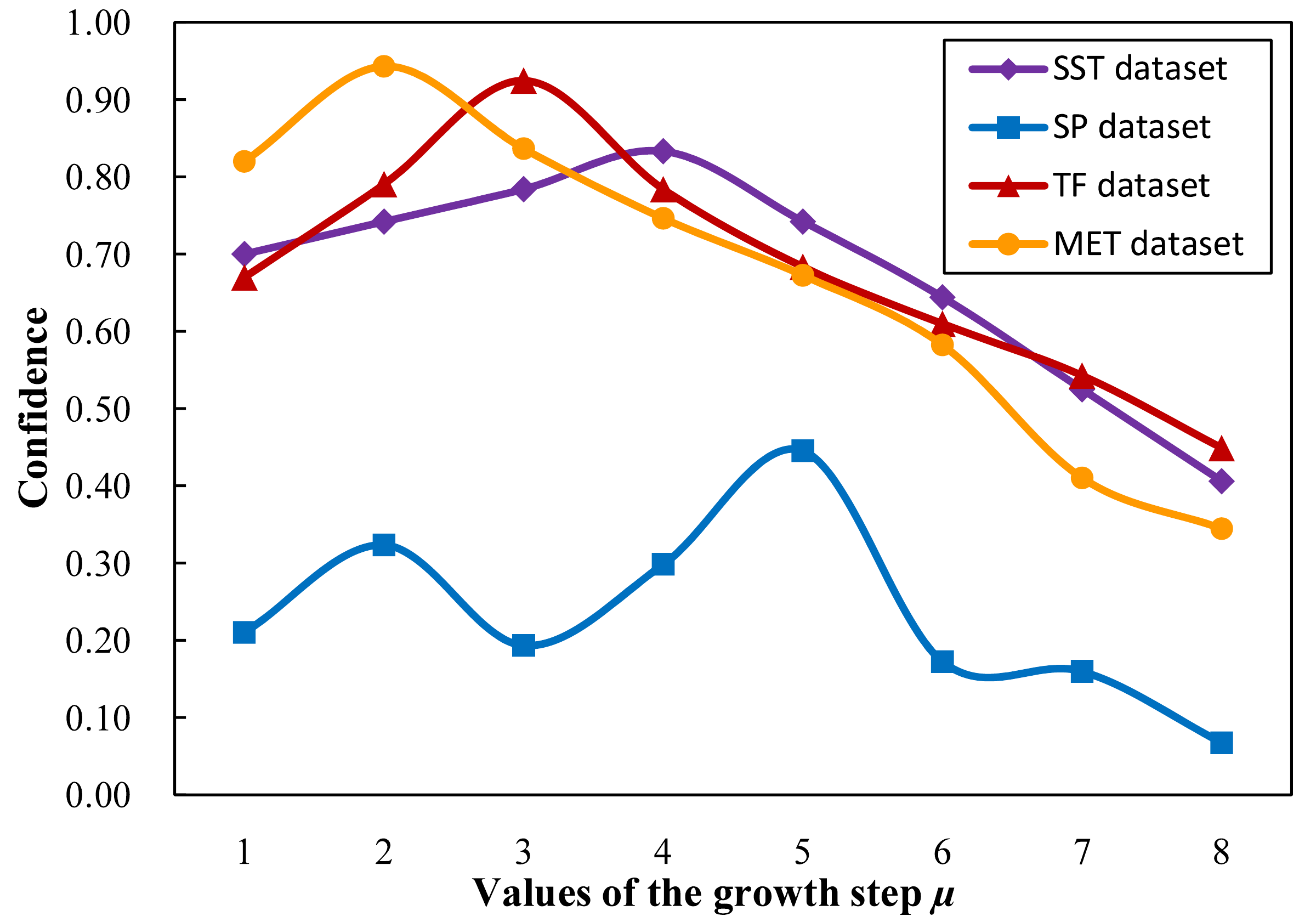}}
 \subfigure[Impact of the scaling ratio factor $\varphi$]{
 \label{chart05:b}
 \includegraphics[width=3.0in]{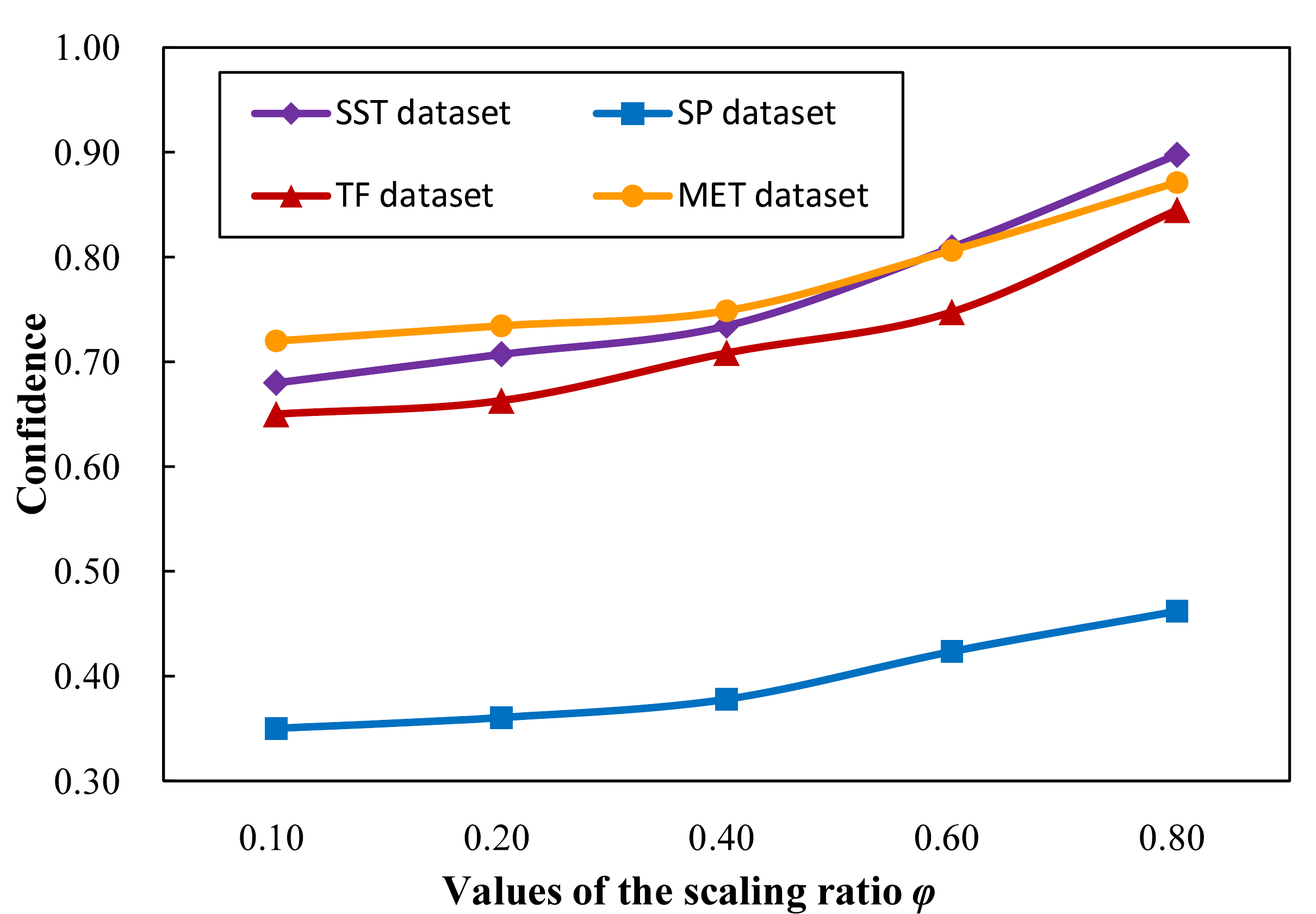}}
 \caption{Impact of parameter thresholds on periodic pattern detection.}
\label{chart05}
\end{figure}

In this section, we conduct experiments to measure the impact of the two parameters on the periodic detection results of the proposed MTSPPR algorithm.
By observing the different confidence of MTSPPR with the different thresholds of $\mu$ and $\varphi$, we can evaluate the impact of the parameters on the accuracy of the periodic patterns.
In each case, the threshold of the growth step $\mu$ is gradually set from 5 to 40.
And the scaling ratio $\varphi$ of the comparison subsequence length is set to 0.1, 0.2, 0.4, 0.6, and 0.8, respectively.
The experimental results of the impact of parameters $\mu$ and $\varphi$ on the MTSPPR algorithm are shown in Figure \ref{chart05}.

Figure \ref{chart05} (a) shows the impact of the growth step $\mu$ on the periodic pattern detection.
We can clearly see that the optimal growth steps of the comparison subsequences are different for different datasets.
For example, the best growth step for MET is 10, with the highest confidence at 0.94.
For the TF dataset, the optimal value of $\mu$ is 15 and the highest confidence is reached at 0.92.
When $\mu$ is set to a small value (i.e., 1 or 2), a candidate periodic sequence may be divided into different comparison subsequences and the best match opportunity is missed.
But in general, the threshold of the growth step $\mu$ should not be set to a too large value (i.e., $\mu \geq 40$).
Because in this case, there might be multiple periodic sequences contained in a single comparison subsequence, and  also missed the best detection opportunity.
As shown in Figure \ref{chart05} (a), when the threshold of $\mu$ exceeds the optimal value, the confidence of MTSPPR decreases in all cases.
For example, in the case of SST, when $\mu$ increases from 20 to 40, the confidence of MTSPPR decreases from 0.83 to 0.41.
The special case is SP.
Due to the presence of partial or weak periodic patterns, it is difficult to take an appropriate growth step threshold that is conducive to confidence.

Figure \ref{chart05} (b) shows the impact of the scaling ratio $\varphi$ of the comparison subsequence length on periodic pattern detection.
We can observe that the confidence of MTSPPR increases with $\varphi$ in all cases.
For example, when the threshold of $\varphi$ increases from 0.1 to 0.8, the confidence of MTSPPR gradually rises from 0.68 to 0.90 in the case of SST, from 0.72 to 0.87 for MET, from 0.65 to 0.85 for TF, and from 0.35 to 0.45 for SP.
According to Equation (\ref{eq16}), an increase in the threshold of $\varphi$ means more candidate comparison subsequences with different lengths are involved to a given subsequence in each operation.
Frankly speaking, while facilitating the accuracy of periodicity detection, the increase of the $\varphi$ threshold will also lead to an increase in computational overhead.
Fortunately, we propose the corresponding parallel algorithms in this work to address this performance problem.

\subsection{Examples of Periodic Detection and Prediction on Actual Time-series Datasets}
In this section, we shown some examples of the periodic pattern detection and the corresponding prediction of the proposed MTSPPR and PTSP algorithms on actual time-series datasets described in Table \ref{table51}.

\subsubsection{Periodic Recognition and Prediction on Meteorological (MET) Datasets}
For the MET datasets, the temperature time-series subset of Beijing city in China is used as an example to show the periodic detection and prediction results.
After training the large-scale historical time-series datasets from 2001/01/01 to 2018/01/31, we can detect a two-layer periodic pattern by the MTSPPR algorithm and predict the temperature in February 2018 by the PTSP algorithm.
But to clarify, the temperature prediction depends on all historical records from 2001/01/01 to 2018/01/31, not just a few days.

\begin{figure}[!ht]
 \setlength{\abovecaptionskip}{0pt}
 \setlength{\belowcaptionskip}{0pt}
\centering
 \subfigure[First-layer periodic pattern]{
 \label{chart06:a}
 \includegraphics[width=3.0in]{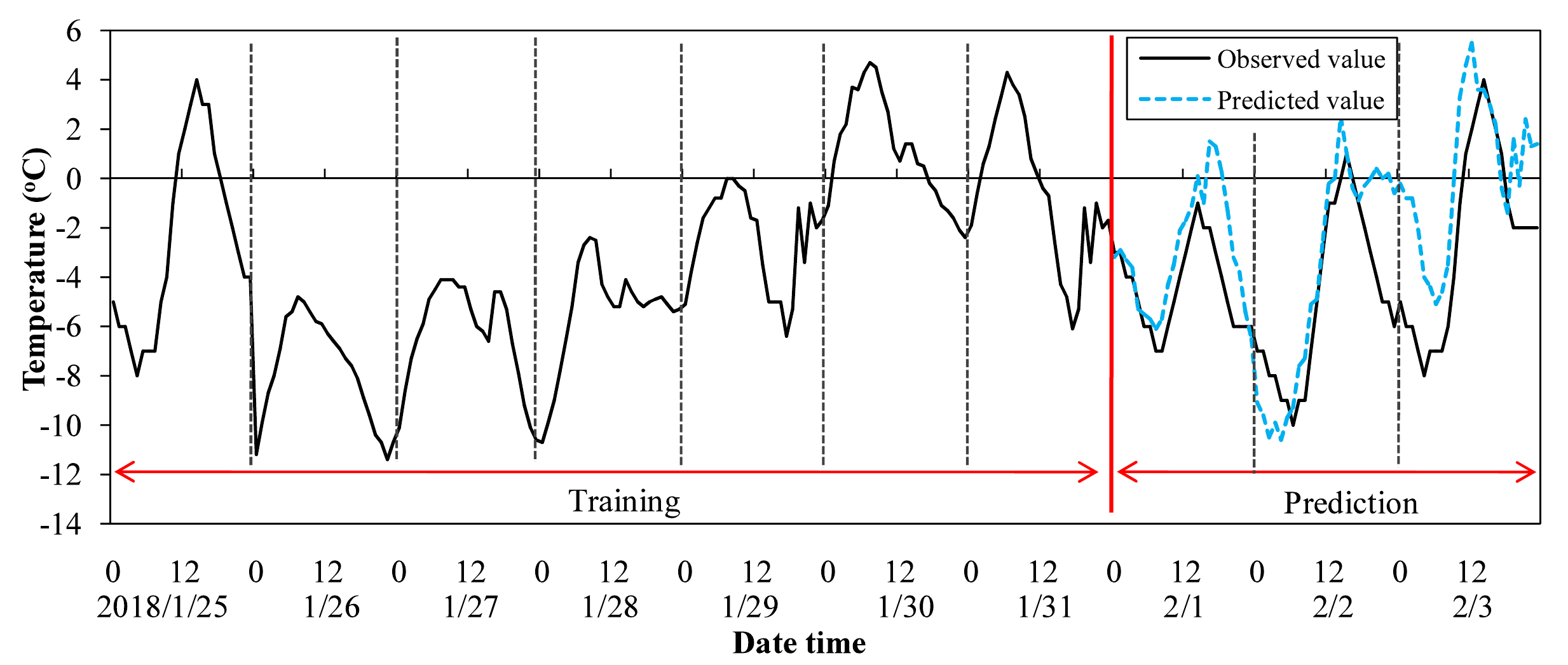}}
 \subfigure[Second-layer periodic pattern]{
 \label{chart06:b}
 \includegraphics[width=3.0in]{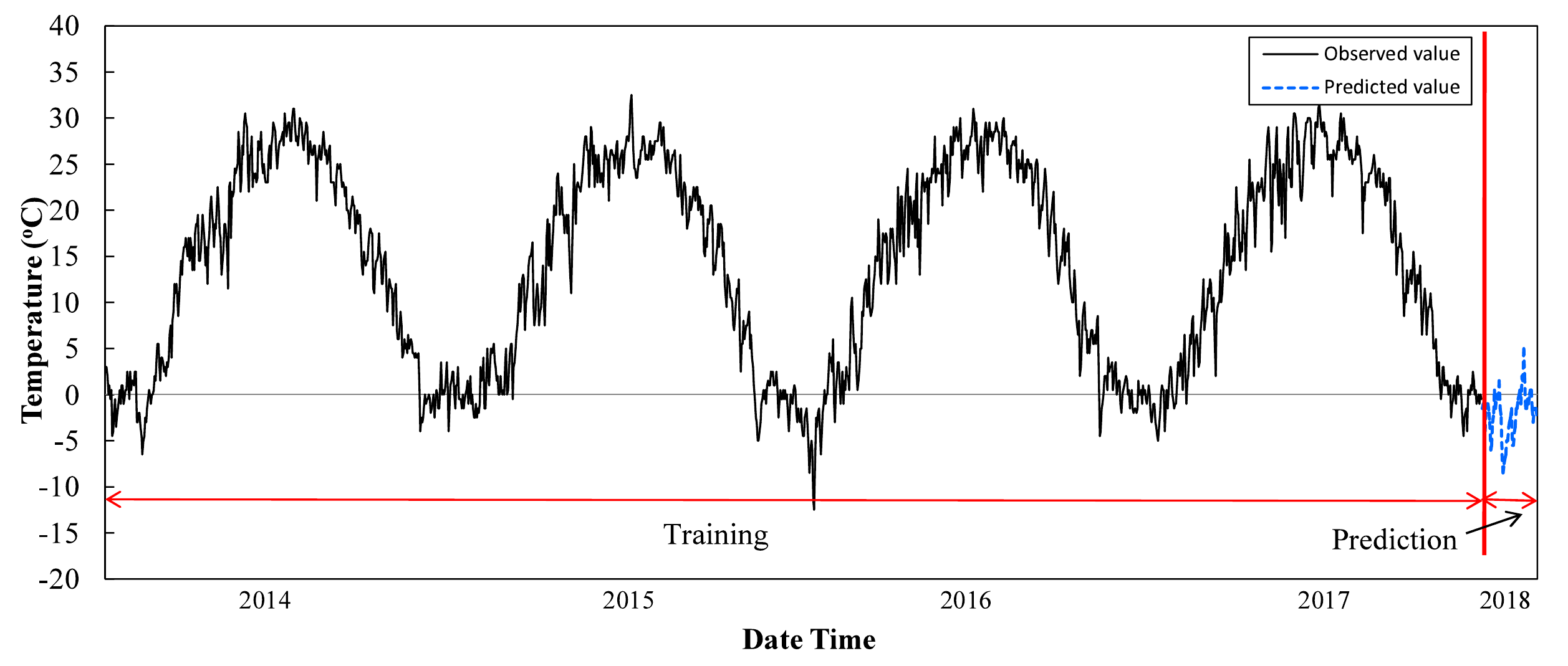}}
\caption{Periodic pattern and prediction on MET dataset.}
\label{chart06}
\end{figure}

After detecting periodic patterns by MTSPPR, it is easy to obtain a two-layer periodic pattern for the MET dataset.
The length of periodicity in the first layer is 24 hours (one day), and the length of periodicity in the second layer is 365 days (one year), as shown in Figure \ref{chart06} (a) and (b), respectively.
At the first layer, there is a peak temperature in each periodic, and the temperature starts with the lowest value at midnight, then continuously rises and reaches the peak value at noon.
Then, the temperature drops and returns the lowest value at midnight.
However, there are not strictly consistent periodic patterns in different days.
On the second layer, periodic patterns are easy to observed, which is a strictly consistent periodic pattern with the length of one year.
Based on the two-layer periodic model, we predict the hourly temperature from February 1st to February 3rd of 2018, as shown in Figures \ref{chart06} (a) and (b).
To clearly show the hourly temperature, Figure \ref{chart06} (a) only depicts the records between January 25th and January 31st of 2018.
We can see from Figure \ref{chart06} (a) that the predicted values are very close to the observed values.
Then, we evaluate the accuracy of the prediction results by comparing the observed values and the predicted values, achieving a value of RMSE at 2.652 by Equation (\ref{eq27}).
The example shows the proposed MTSPPR and PTSP algorithms achieve high confidence in periodic pattern detection and high accuracy in time-series prediction on the MET dataset, respectively.

\subsubsection{Periodic Recognition and Prediction on Sea-Surface temperature (SST) dataset}
\begin{figure}[!ht]
 \setlength{\abovecaptionskip}{0pt}
 \setlength{\belowcaptionskip}{0pt}
\centering
\includegraphics[width=5.4in]{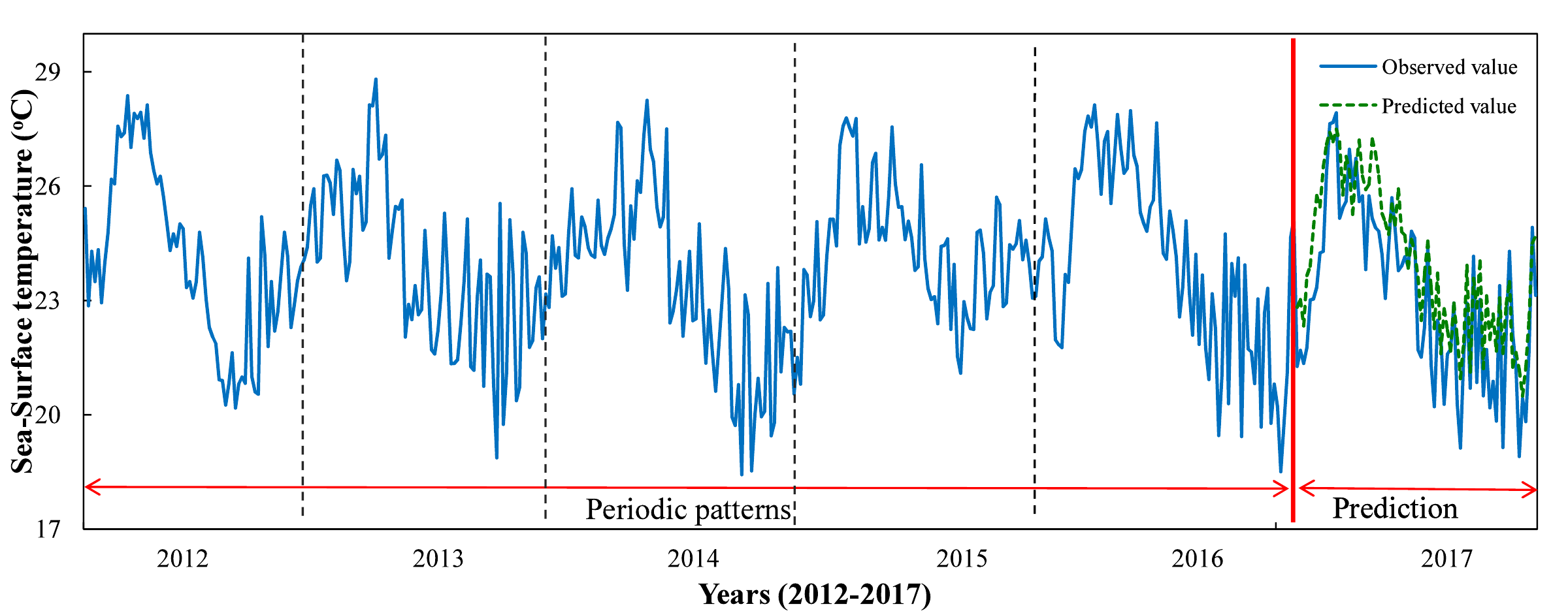}
\caption{Periodic pattern and prediction on SST dataset.}
\label{chart07}
\end{figure}

For the SST datasets, a time-series subset of an observation site (location: 0N, 95W) in the Pacific Oceans is used as an example to show the periodic detection and prediction results.
The temperature at 1 meter below the sea surface is collected at 12 o'clock every day.
We use the average of the same position in all years to replace the missing values.
After training the large-scale historical time-series datasets from 1978/01/01 to 2016/12/31, we can detect the periodic pattern by the MTSPPR algorithm and predict the temperature in 2017 by the PTSP algorithm.
To clearly show the periodic pattern and the prediction results, we only show the records between 2012 and 2017 in Figure \ref{chart07}.

After detecting periodic patterns by MTSPPR, it is not difficult to obtain a single-layer periodic pattern for the SST dataset.
There is an obvious periodicity in the SST dataset with a length of 365 days (one year).
Although the sea-surface temperature at each time point fluctuates, the overall tendency of the temperature for each periodic is fixed.
The first half of the periodic shows an upward trend, while the second half of the periodic shows a downward trend and ends with an increase.
In addition, the range of values for each peak and valley is similar.
In each period, the temperature starts from about 23 $^\circ$C in January, then rises to the peak values of 27.3 $^\circ$C and 28.5 $^\circ$C during the summer.
Then, the temperature drops to the minimum values of about 18 $^\circ$C between October and December.
Based on the detected periodic model from 1978 to 2016, we predict the temperature for the next periodic (the whole year of 2017), as shown on the right side of Figure \ref{chart07}.
We can see that the predicted values are very close to the observed values.
Then, we evaluate the accuracy of the prediction results by comparing the observed values and the predicted values, achieving a value of RMSE at 1.327.
The example shows the proposed MTSPPR and PTSP algorithms achieve high confidence in periodic pattern detection and high accuracy in time-series prediction on the SST dataset, respectively.

\subsubsection{Periodic Recognition and Prediction on Traffic flow (TF) dataset}

For the TF datasets,  traffic data are continuously collected from over 5,971 loop sensors located around the Twin Cities Metro freeways, seven days a week and all year round.
In this section, we choose a subset of a freeway (U.S.169, NB), where the start station is (S1142, mp=0.000mi, Old Shakopee Rd) and the destination is (S757, mp=20.138mi, 77th Ave).
After observing and preprocessing the raw traffic data in the format of binary UTSDF (Unified Transportation Sensor Data Format), we obtain the expected travel time between the start station and the destination station at different starting times.
After training the large-scale historical time-series datasets from 2011/01/01 to 2018/04/20, we can detect a two-layer periodic pattern through the MTSPPR algorithm and use the PTSP algorithm to predict the travel time in the last week of April 2018.

\begin{figure}[!ht]
 \setlength{\abovecaptionskip}{0pt}
 \setlength{\belowcaptionskip}{0pt}
\centering
\includegraphics[width=5.4in]{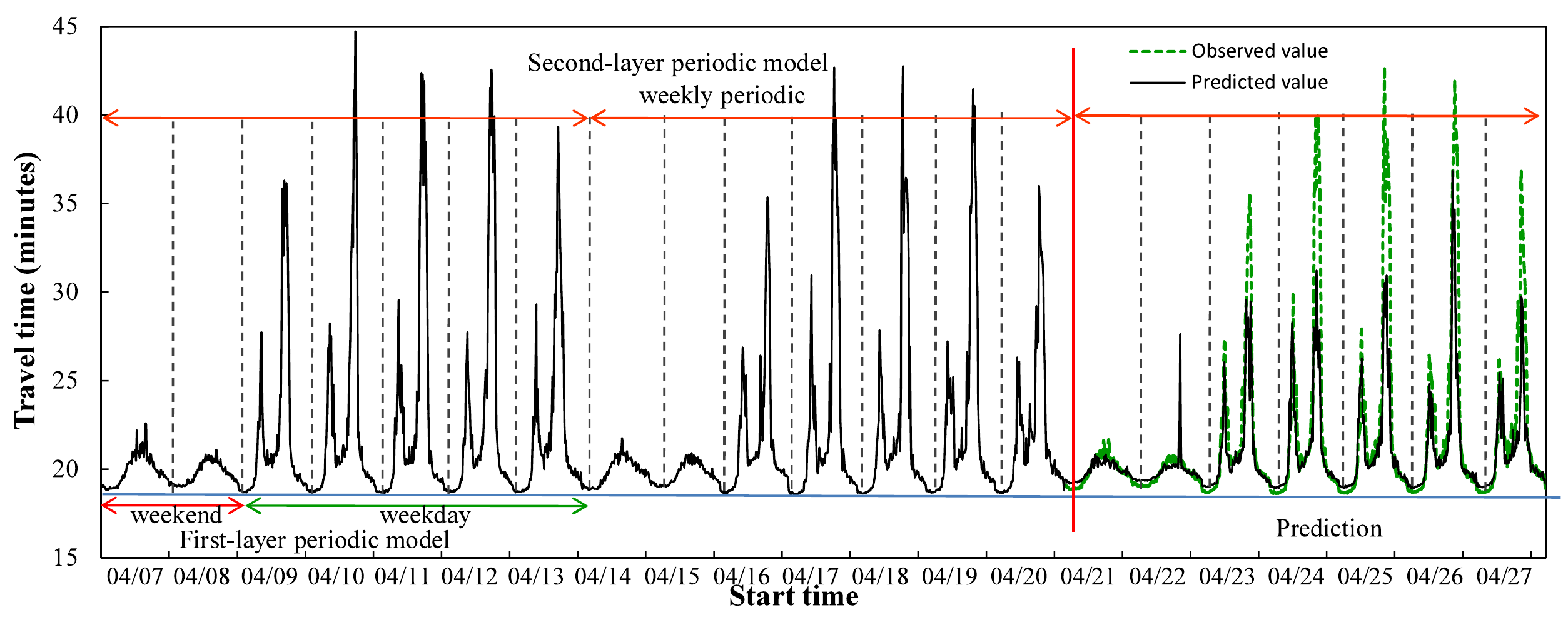}
\caption{Periodic pattern and prediction on TF dataset.}
\label{chart08}
\end{figure}

To clearly show the periodic pattern and the prediction results, we only show periodic detection results from 2018/04/07 to 2018/04/20, and the prediction results between 2018/04/21 and 2018/04/27, as shown in Figure \ref{chart08}.
After detecting periodic patterns by MTSPPR, it is easy to obtain a two-layer periodic pattern of the TF dataset.
There are two types of periodic patterns in the first layer: 2 weekend periods and 5 weekday periods, where each period has the same length of 24 hours (one day).
In each weekend period, there is only one peak travel time between 14:00 and 16:00, and the peak value is approximately 21 (minutes).
In each weekday period, there are two peaks of travel time, where the first one appears at 7:00 - 8:00 with values of approximately 28 (minutes), and the second one appears at 16:30 - 19:00 with the values of about 39 - 43 (minutes).
The periodic pattern in the second layer is a strictly consistent periodic pattern with the length of 7 days (one week), containing 2 weekend periods and 5 weekday periods.
Based on the detected periodic model between 2011/01/01 and 2018/04/20, we predict the temperature for the next periodic (the last week in April 2008: 2018/04/21 - 2018/04/27), as shown on the right side of Figure \ref{chart08}.
We can see that the predicted values are close to the observed values, although there is a temporary peak in the observed values of 04/22 may be caused by traffic accidents.
Then, we evaluate the accuracy of the prediction results by comparing the observed values and the predicted values, achieving a value of RMSE at 2.551.
The example shows the proposed MTSPPR and PTSP algorithms achieve high confidence in periodic pattern detection and high accuracy in time-series prediction on the TF dataset, respectively.

\subsubsection{Periodic Recognition and Prediction on Stock price (SP) dataset}
For the MET dataset, the opening stock price of a company (000001) between 1991/01 and 2004/12 is used as an example to show the periodic detection and prediction results.
After training the large-scale historical time-series datasets from 1991/01 and 2004/12, we can detect a partial periodic pattern through the MTSPPR algorithm and use the PTSP algorithm to predict the opening price in 2005.

\begin{figure}[!ht]
 \setlength{\abovecaptionskip}{0pt}
 \setlength{\belowcaptionskip}{0pt}
\centering
\includegraphics[width=5.4in]{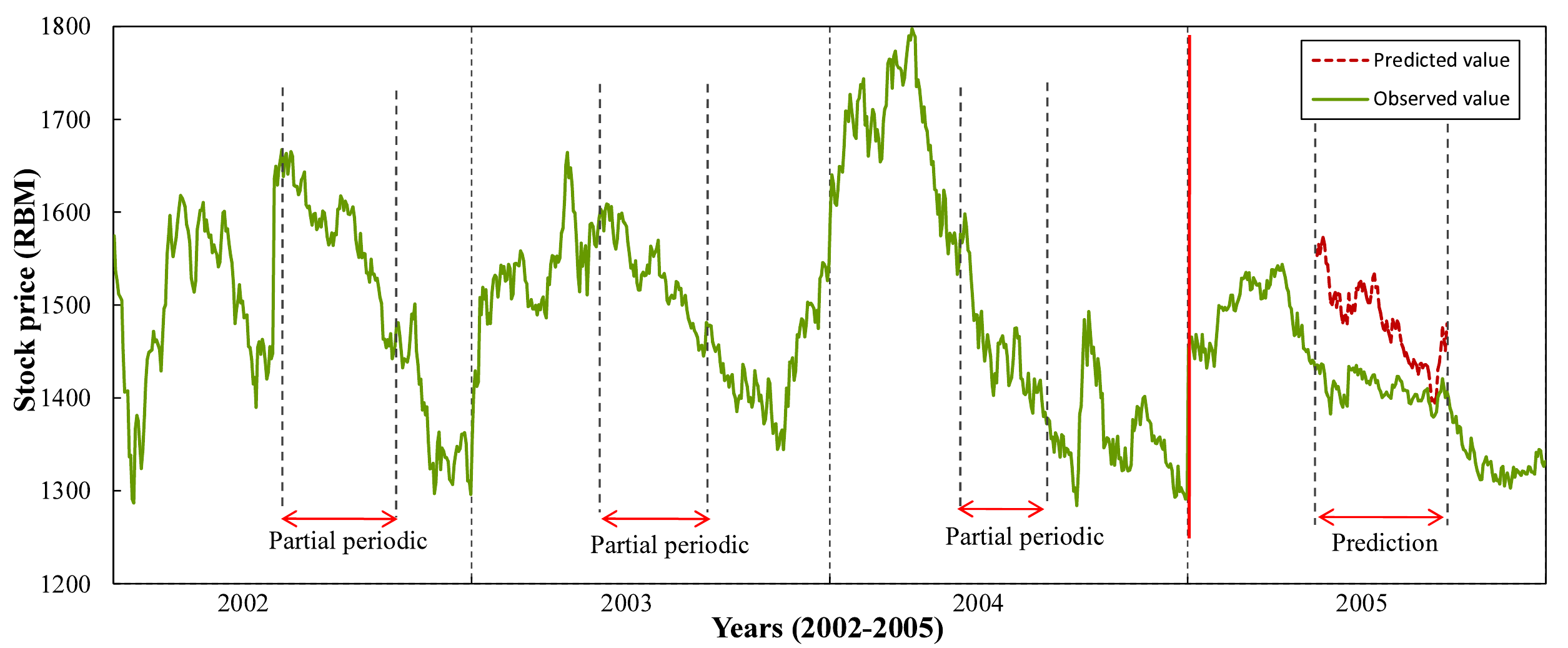}
\caption{Periodic pattern and prediction on TF dataset.}
\label{chart09}
\end{figure}
To clearly show the periodic pattern and the prediction results, we only show periodic detection results from 2002 to 2004 and the prediction results in 2005, as shown in Figure \ref{chart09}.
Different from the MET, TF, and SST datasets, the time-series of SP is complex and almost without periodicity.
After detecting periodic patterns by MTSPPR, a partial periodic pattern is found in each year between May and October, with the confidence value of 0.48.
Hence, depending on the detected partial periodic model, we can only predict the opening price for the same period in 2005, as shown on the right side of Figure \ref{chart09}.
We can see that the trend of the predicted values is similar to that of the observed values.
However, the difference between the predicted value and the observed value at each time point is large, achieving a value of RMSE at 78.981.
Therefore, the example shows the proposed MTSPPR and PTSP algorithms do not perform satisfactorily on the weak-periodic and complex time-series datasets such as TF.

\subsection{Accuracy Evaluation of Time Series Prediction}
In this section, we conduct comparison experiments to evaluate the accuracy of the proposed PTSP algorithm and the related time-series prediction algorithms, including the Holt-Winters \cite{ec02}, Support Vector Machine (SVM) \cite{ec03}, USM \cite{ec04}, TDAP\cite{ec12}, and OSAF+AKF \cite{ec08} algorithms.

\subsubsection{Evaluation Metric}
We introduce the Root Mean Square Error (RMSE) to evaluate the comparison algorithms, which is the square root of the observed values and the predicted values, as defined in Equation (\ref{eq29}):
\begin{equation}
\label{eq29}
RMSE = \sqrt{\frac{1}{n}\sum\limits_{i=1}^{n}{(x_{i} - x_{i}')^{2}}},
\end{equation}
where $n$ is the number of prediction time and $(x_{i} - x_{i}')^{2}$ is the deviation of a set of the predicted values from the observed values.
RMSE is very sensitive to a set of large or small error in a set of measurements.
Therefore, RMSE can reflect the accuracy of the measurements.

\subsubsection{Accuracy Evaluation of Comparison Algorithms}
\begin{figure}[!ht]
 \setlength{\abovecaptionskip}{0pt}
 \setlength{\belowcaptionskip}{0pt}
\centering
 \subfigure[MET dataset]{
 \label{chart10:a}
 \includegraphics[width=3.0in]{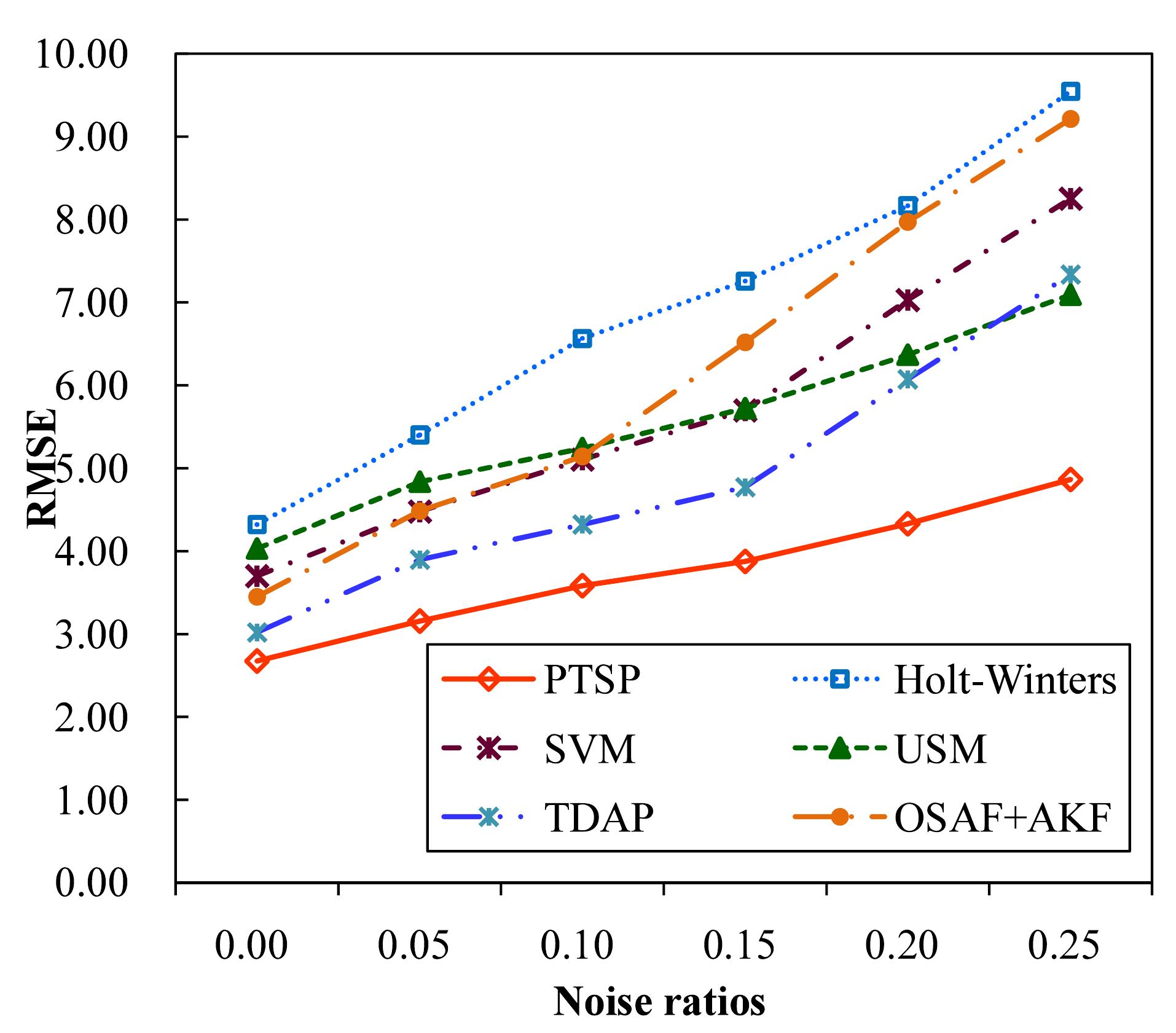}}
 \subfigure[SST dataset]{
 \label{chart10:b}
 \includegraphics[width=3.0in]{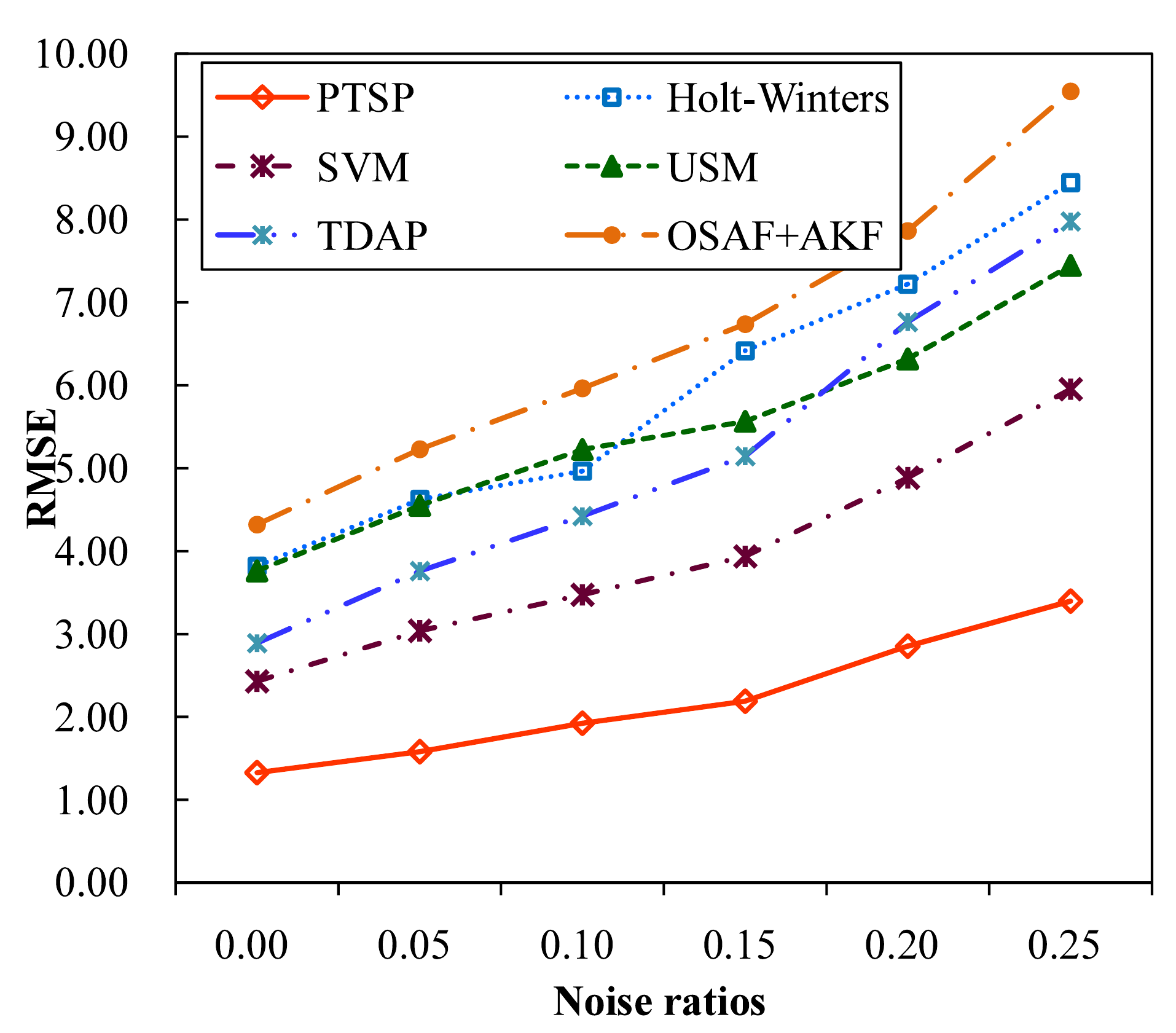}}
  \subfigure[TF dataset]{
 \label{chart10:c}
 \includegraphics[width=3.0in]{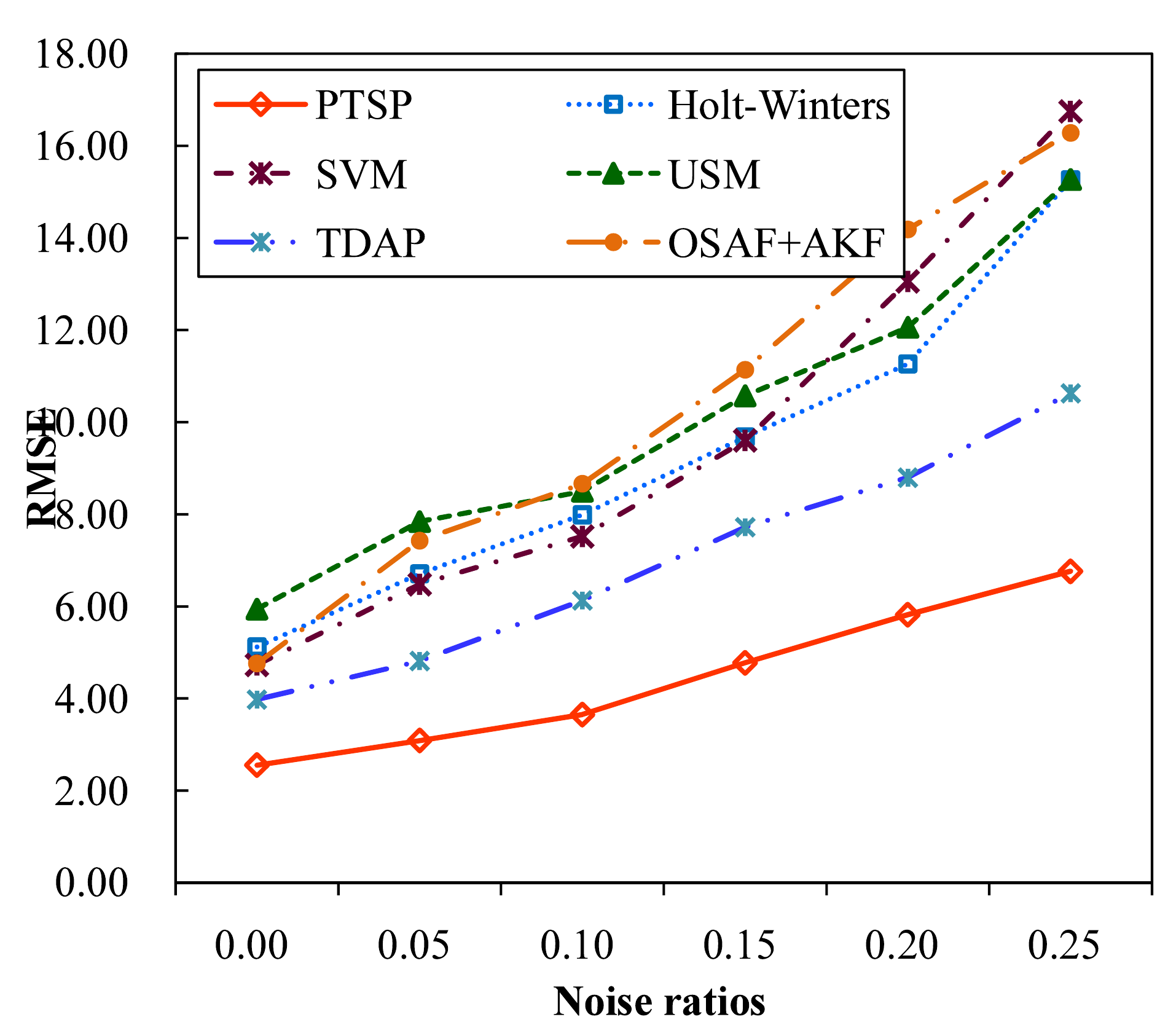}}
 \subfigure[SP dataset]{
 \label{chart10:d}
 \includegraphics[width=3.0in]{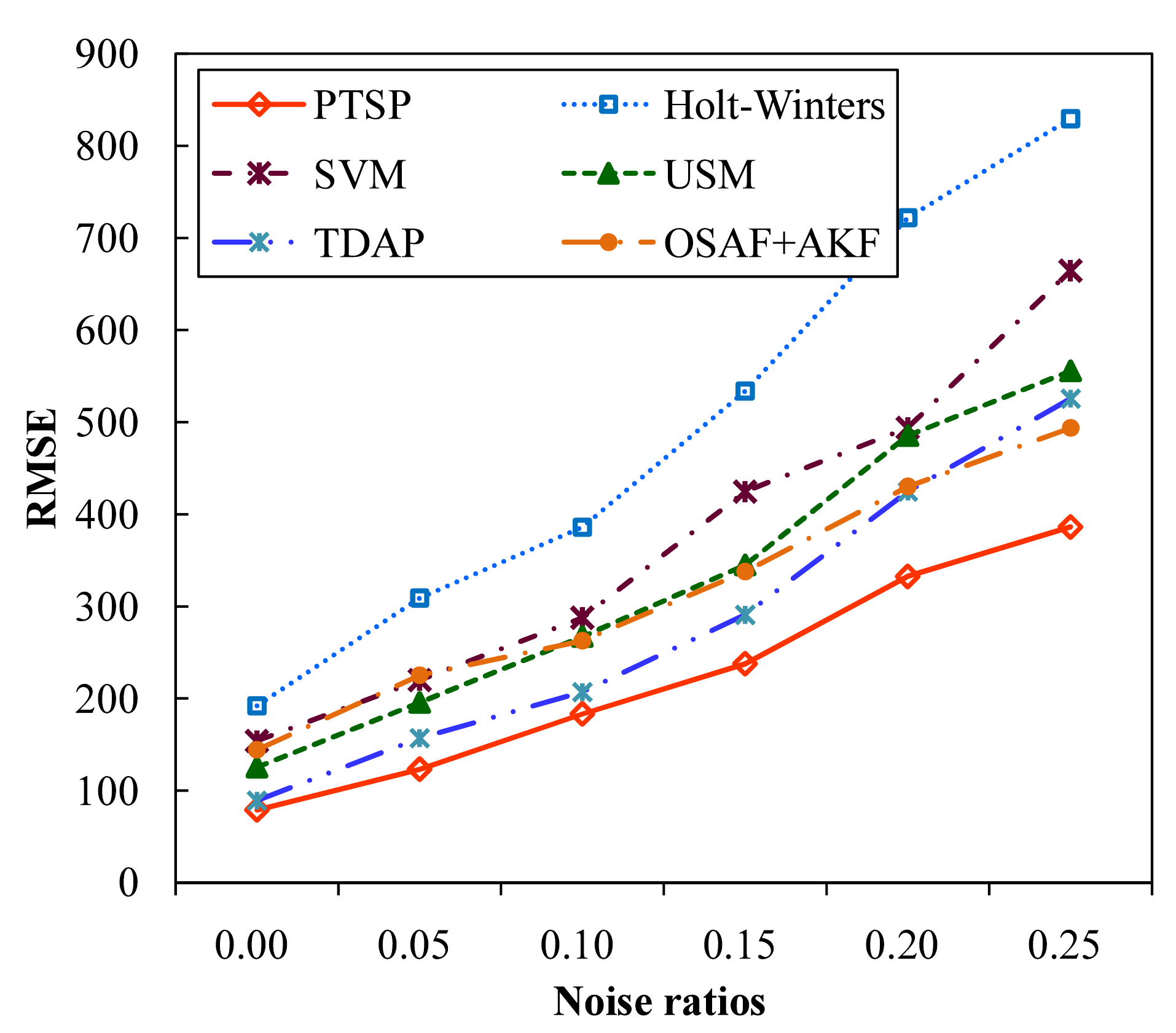}}
\caption{Prediction accuracy of comparison algorithms.}
\label{chart10}
\end{figure}

The actual time-series datasets described in Table \ref{table51} are used in the experiments to evaluate the prediction accuracy of the comparison algorithms.
Considering that some comparison algorithms are claimed to have good noise immunity and they can obtain high accuracy in the noisy data, we use the actual time-series datasets rather than synthesized datasets in our experiments.
If a time series prediction algorithm can achieve hight accuracy in actual and noisy data, then the algorithm is popular and feasible in practical applications.
Therefore, to clearly demonstrate the accuracy and robustness of each comparison algorithm under noisy data, we add different levels of noisy data to the original datasets.
We generate different amounts of random and non-repetitive data in the value space of the original dataset as noisy data.
The average prediction accuracy of each case is recorded and compared, as shown in Figure \ref{chart10}.

As can be observed from Figure \ref{chart10} (a) - (d), as the proportion of noisy data increases, the average RMSE of each algorithm increases in all cases.
Taking into account the different natural periodicity of each raw time-series dataset, comparison algorithms obtain different accuracy on different datasets.
The RMSE value of each algorithm on the MET dataset is in the range of 2.5 - 9.5, in the range of 1.3 - 9.8 on the SST dataset, and in the range of 2.5 - 17.5 on the TF dataset.
For the complex SP dataset, all comparison algorithms obtain high RMSE values, where the RMSE of PTSP is 78.981,
 that of TDAP is 89.21, that of USM is 125.43, and that of OSAF+AKF, SVM, and Holt-Winters is 144.32, 153.82, and 192.00, respectively.
In addition, obviously, our PTSP algorithm keeps the lowest RMSE in all cases against other comparison algorithms.
For example, on each original dataset (noise ratio = 0.0), PTSP achieves the value of RMSE at 2.67 on the MET dataset, 1.33 on SST, 2.55 on TF, and 78.981.
In cases of MET and SP, as the noise ration increase, the RMSE of Holt-Winters increases at the fastest speed, while that of TDAP decreases at least.
Taking the MET case as an example, when the noise increases from 0.00 to 0.25, the RMSE of Hot-Winters and increases from 4.32 to 9.55, and that of OSAF+AKF increases from 3.45 to 9.21, indicating that the robustness of Hot-Winters and OSAF+AKF is poor for noisy data.
In contrast, as the noise ration increase, the RMSE of PTSP increases only from 2.67 to 4.86 for MET,  from 1.33 to 3.40 for SST, and from 2.55 to 6.76 for TF.
Therefore, we can conclude that the proposed PTSP algorithm obtains higher prediction accuracy on time-series datasets and is more robust to noisy data than other comparison algorithms.

\subsection{Performance Evaluation}
\label{section5.3}
\subsubsection{Execution Time and Scalability}
\begin{figure}[!ht]
 \setlength{\abovecaptionskip}{0pt}
 \setlength{\belowcaptionskip}{0pt}
\centering
 \subfigure[Different data sizes]{
 \label{chart11:a}
 \includegraphics[width=3.0in]{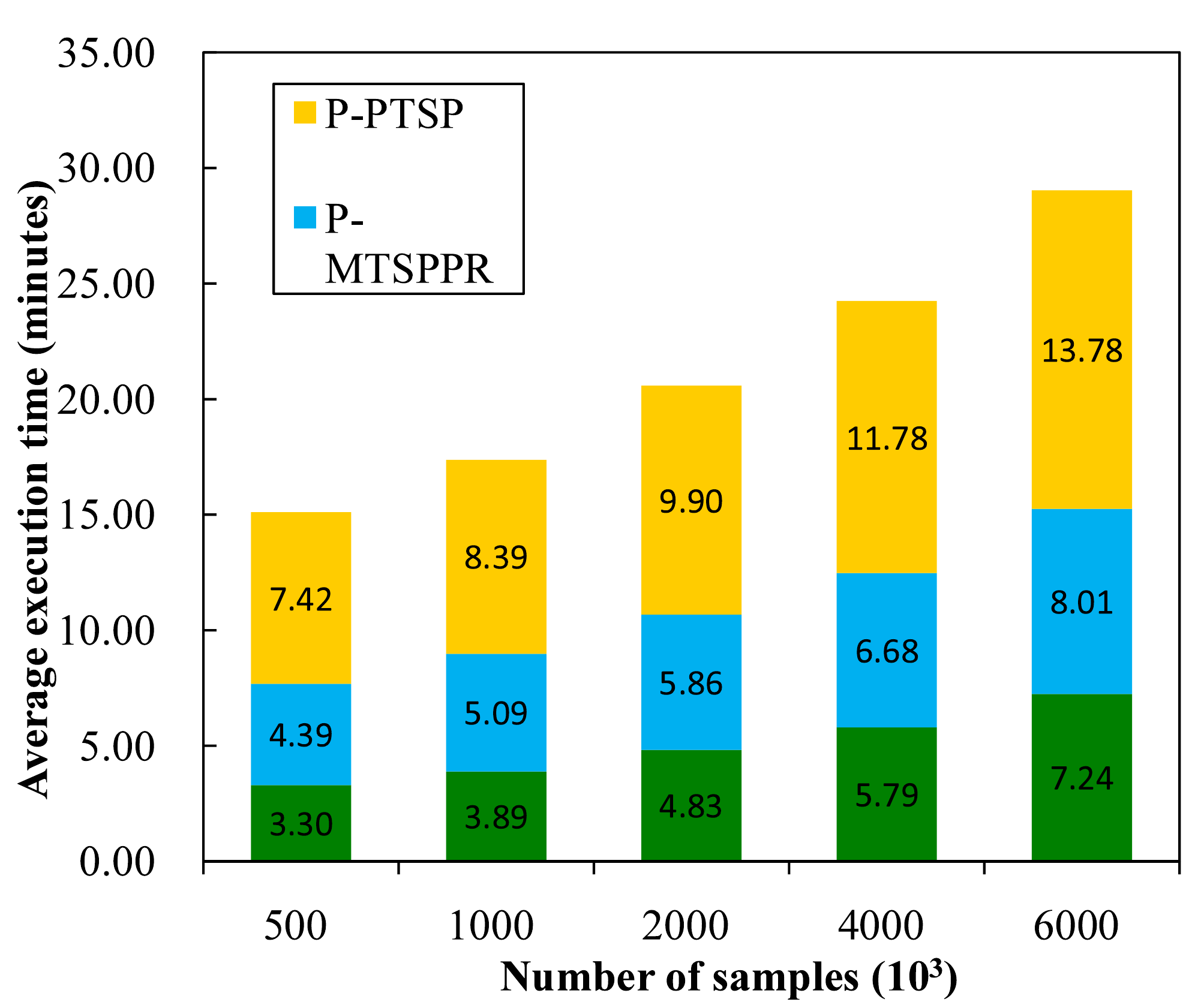}}
 \subfigure[Different computing nodes]{
 \label{chart11:b}
 \includegraphics[width=3.0in]{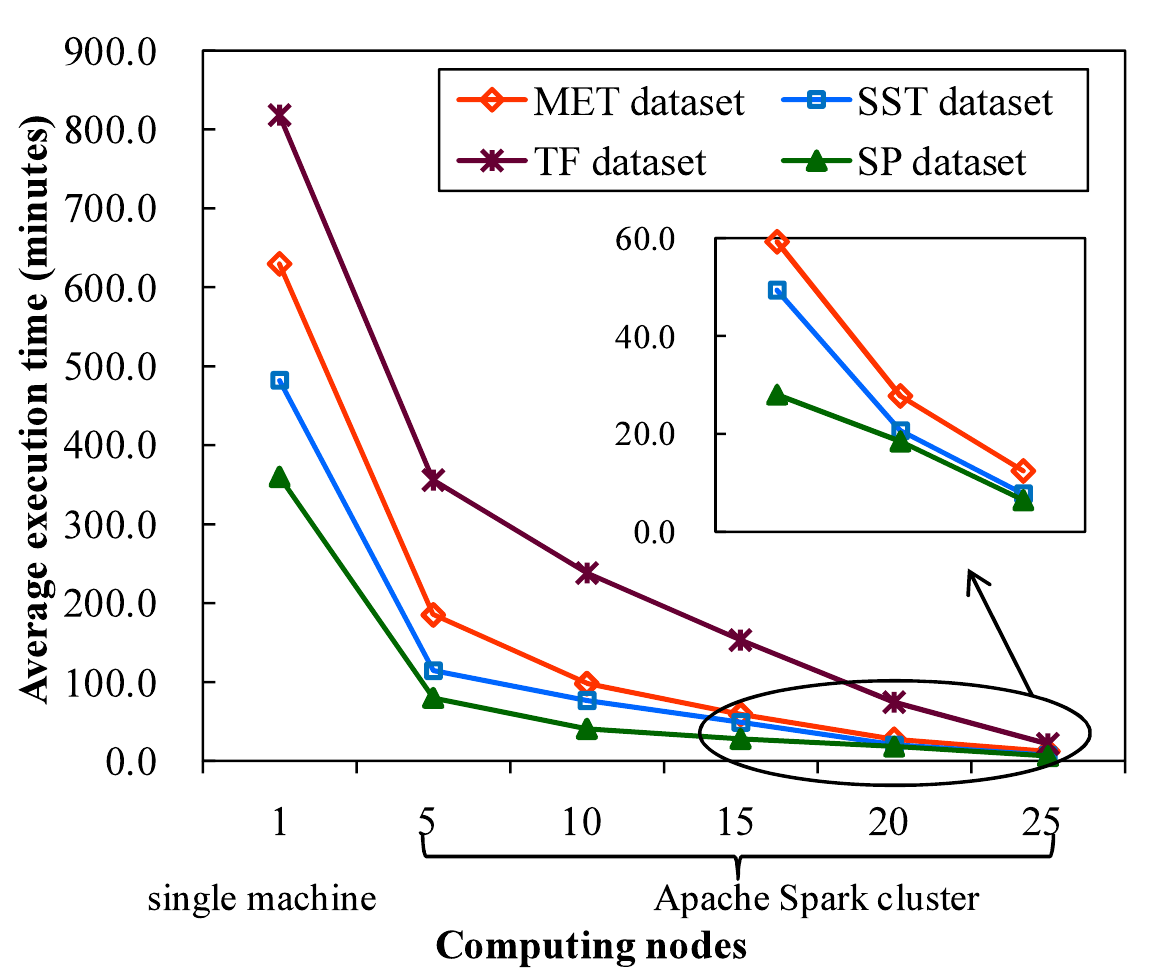}}
\caption{Performance evaluation of the PPTSP algorithm.}
\label{chart11}
\end{figure}

The performance of PPTSP is evaluated on the Apache Spark platform in terms of execution time and scalability.
The four time-series datasets described in Tabel \ref{table51} are used in the experiments.
Considering that the PPTSP algorithm consists of P-TSDCA, P-MTSPPR, and P-PTSP processes, the average execution time of these processes is recorded and compared.
The experiment results of execution time and scalability evaluation of PPTSP are presented in Figure \ref{chart11} (a) and (b).

In Figure \ref{chart11} (a), a Spark computing cluster equipped with 20 computing nodes is configured for the experiments.
Although the number of datasets doubled increases, benefiting from the high parallel computing power of the Apache Spark platform, the average execution time of each process slowly increases.
For example, when the sample size increases from 500 $\times 10^{3}$ to 1000 $\times 10^{3}$, the average execution time of P-TSDCA only increases by 3.94m (from 3.30m to 7.24s), that of P-MTSPPR increases by 3.62m, and that of P-PTSP raises 6.36m.
When the number of samples is 2000 $\times 10^{3}$, the overall execution time of the PPTSP algorithm is 20.58m, where the P-TSDCA process costs 4.83m, the T-MTSPPR process costs 5.86m, and the P-PTSP process costs 9.90m.

In addition, we compare the performance of the original version of the PTSP algorithm and the parallel version PPTSP, which includes the P-TSDCA, P-MTSPPR, and P-PTSP processes.
As shown in Figure \ref{chart05} (b), when the number of computing nodes is equal 1, meaning that we conduct the original version of the PTSP algorithm on a single machine.
The execution time of the original PTSP algorithm on the MET dataset is 629.6 minutes, as well as 482.0m for the SST dataset, 818.0m for the TF dataset, and 360.3m for the SP dataset.
After we conduct the parallel version PPTSP on the Spark computing cluster equipped with 5 nodes, the execution time rapidly decreases in all cases.
In the case of MET, the execution time of PPTSP decreases to 185.2m, achieving an acceleration of 3.4.
In the case of SP, the execution time of PPTSP decreases to 80.1m, achieving an acceleration of 4.5.
The scalability of the proposed PPTSP algorithm is evaluated under different scales of computing clusters.
Due to the massive volumes, MET and TF datasets spend more execution time than SST and SP.
When there are five computing nodes, MET and TF have obviously higher execution time than that of other datasets.
The execution time of TF is high to 355.6m and that of MET reaches 185.2m, while that of SP and SST is equal to 114.8m and 80.1m, respectively.
However, as the scale of the computing cluster increases, the gap becomes smaller.
When the number of computing nodes is equal to 25, the execution time of TF and MET is low at 21.7m and 12.4m, while that of SP and SST is equal to 7.6m and 6.4m, respectively.
Therefore, the experimental results indicate the proposed PPTSP algorithm obtains high performance and excellent scalability for large-scale time-series datasets.

\subsubsection{Data Communication and Workload Balance}
\begin{figure}[!ht]
 \setlength{\abovecaptionskip}{0pt}
 \setlength{\belowcaptionskip}{0pt}
\centering
 \subfigure[Different data sizes]{
 \label{chart12:a}
 \includegraphics[width=3.0in]{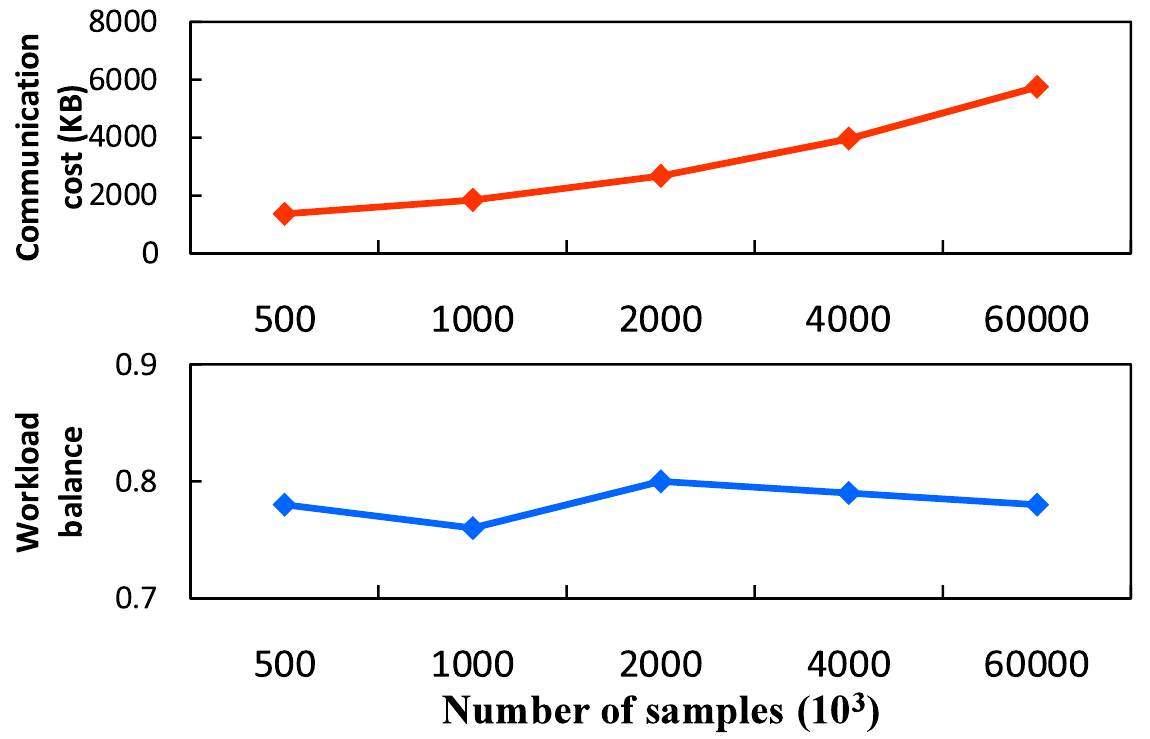}}
 \subfigure[Different computing nodes]{
 \label{chart12:b}
 \includegraphics[width=3.0in]{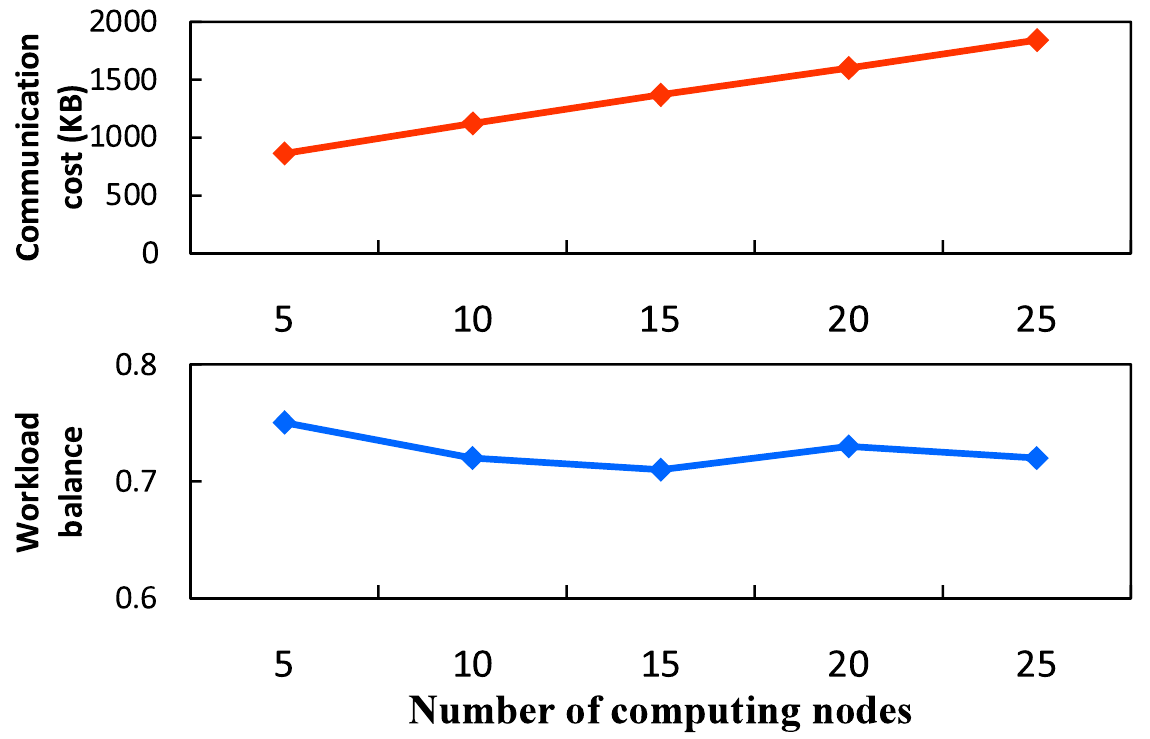}}
\caption{Data communication and workload balance of the PPTSP algorithm.}
\label{chart12}
\end{figure}

The experiments are conducted to evaluate the data communication and workload balance of the proposed PPTSP algorithm from the perspectives of the impact of data sizes and computing cluster scales, respectively.
The number of samples gradually increases from 100,000 to 6,000,000, while the number of computing nodes increases from 5 to 25 in each case.
The size of shuffle writing operation is recorded as the communication cost, and the variance of the CPU usage of all computing nodes is measured as the workload balance.
The experimental results of average data communication and workload balance of PPTSP are shown in Figure \ref{chart12}.

As can be clearly seen from Figure \ref{chart12} (a) and (b), in most cases, the PPTSP algorithm has a stable workload balancing and compromise data communication costs.
As the data size increases, more shuffle writing operations are required, resulting in the rise of data communication, from 136.32 KB to 5743.83 KB.
Satisfactory, the workload balance of the computing cluster remains within the range of (0.76 \~ 0.80), keeping well steady.
In addition, as the number of computing nodes increases, the data communication operations among these nodes are more than ever, which leads to more communication costs.
As shown in Figure \ref{chart12} (b), when the number of computing nodes increases from 5 to 10, the communication cost raises from 862.10 KB to 112.63 KB, with a growth rate of 1.3.
With the increase of computing nodes from 10 to 15, 20, and 25, the growth rate of communication costs is 1.22, 1.17, and 1.15, respectively.
Benefitting from cloud computing resource management and parallel task scheduling, the workload balance of the Spark computing cluster also keeps well steady in all cases.
Experimental results demonstrate that the increase in data volume and the expansion of computing clusters will not result in a substantial increase of data communication and a serious imbalance in workload.
The proposed PPTSP algorithm significantly keeps the workload balance of the distributed cloud computing cluster with acceptable communication costs.

\section{Conclusions}
\label{section6}
This paper proposed a periodicity-based parallel time series prediction (PPTSP) algorithm for large-scale time-series data on the Apache Spark cloud computing platform.
The PPTSP algorithm consists of three processes, including time-series data compression and abstraction (TSDCA), multi-layer time series periodic pattern recognition (MTSPPR), and periodicity-based time series prediction (PTSP).
The large-scale historical time-series datasets are effectively compressed using the TSDCA algorithm while the core data characteristics are accurately extracted.
Potential multi-layer periodic patterns were detected by the MTSPPR method from the compressed time-series dataset.
The parallel solutions of the TSDCA, MTSPPR, and PTSP algorithms were implemented on the Apache Spark cloud platform, where the data dependency and task scheduling problems are addressed.
Critical issues in terms of data communication, waiting for synchronization, and workload balancing, are considered by proposing the corresponding optimization methods.
Extensive experimental results demonstrate that the proposed PPTSP algorithm achieved significant advantages compared with other algorithms in terms of accuracy and performance of prediction.
The PPTSP algorithm can effectively improve the performance and keep high scalability and low data communication costs.

For future work, we will further concentrate on parallel and scalable time-series data mining algorithms in distributed computing clusters and high-performance computers.
In addition, the development of these time series mining algorithms in scientific applications is also an interesting topic, such as analysis of marine meteorological and biological time series data.

\section*{Acknowledgment}
This research was partially funded by the Key Program of the National Natural Science Foundation of China (Grant No. 61432005),
the National Outstanding Youth Science Program of National Natural Science Foundation of China (Grant No. 61625202),
the National Natural Science Foundation of China (Grant Nos. 61672221),
the China Scholarships Council (Grant Nos. 201706130080),
and the Hunan Provincial Innovation Foundation For Postgraduate (Grant No. CX2017B099).

\section*{References}

\bibliography{reference}

\end{document}